\documentclass{article}

\usepackage{arxiv}

\usepackage[utf8]{inputenc} 
\usepackage[T1]{fontenc}    
\usepackage{url}            
\usepackage{amsfonts}       
\usepackage{nicefrac}       
\usepackage{microtype}      
\usepackage{natbib}

\usepackage{doi}

\usepackage{times}
\usepackage{epsfig}
\usepackage{graphicx}
\usepackage{amsmath}
\usepackage{amssymb}
\usepackage{tikz}
\usepackage{comment}
\usepackage{color}
\usepackage{algorithm}
\usepackage{algpseudocode}
\usepackage{booktabs}
\usepackage{subcaption}
\usepackage{enumitem}

\usepackage{hyperref}

\renewcommand{\shorttitle}{Clustering-based Task Agnostic Continual Learning}

\begin{document}

\title{Clustering-based Domain-Incremental Learning}

\author{Christiaan Lamers {\renewcommand{\thefootnote}{1}\footnotemark}\\
NORCE\\
Grimstad, 4879, Norway\\
\texttt{chla@norceresearch.no}
\And 
Ren\'e Vidal\\
Institute for Data Engineering and Science \\
University of Pennsylvania \\ 
Philadelphia, USA\\
\texttt{vidalr@seas.upenn.edu}
\And 
Nabil Belbachir\\
NORCE\\
Grimstad, 4879, Norway\\
\texttt{nabe@norceresearch.no}
\And 
Niki van Stein\\
Leiden Institute of Advanced Computer Science\\
Leiden, 2333 CA, The Netherlands\\
\texttt{n.van.stein@liacs.leidenuniv.nl}
\And
Thomas B\"ack\\
Leiden Institute of Advanced Computer Science\\
Leiden, 2333 CA, The Netherlands\\
\texttt{t.h.w.baeck@liacs.leidenuniv.nl}
\And
Paris Giampouras\\
Johns Hopkins University\\
Baltimore, MD 21218, US\\
\texttt{parisg@jhu.edu}
}

\maketitle

{\renewcommand{\thefootnote}{1}
\footnotetext{Corresponding author.}}
\begin{abstract}
We consider the problem of learning multiple tasks in a continual learning setting in which data from different tasks is presented to the learner in a streaming fashion. A key challenge in this setting is the so-called ``catastrophic forgetting problem", in which the performance of the learner in an ``old task" decreases when subsequently trained on a ``new task". Existing continual learning methods, such as Averaged Gradient Episodic Memory (A-GEM) and Orthogonal Gradient Descent (OGD), address catastrophic forgetting by minimizing the loss for the current task without increasing the loss for previous tasks. However, these methods assume the learner knows when the task changes, which is unrealistic in practice. In this paper, we alleviate the need to provide the algorithm with information about task changes by using an online clustering-based approach on a dynamically updated finite pool of samples or gradients. We thereby successfully counteract catastrophic forgetting in one of the hardest settings, namely: domain-incremental learning, a setting for which the problem was previously unsolved. We showcase the benefits of our approach by applying these ideas to projection-based methods, such as A-GEM and OGD, which lead to task-agnostic versions of them. Experiments on real datasets demonstrate the effectiveness of the proposed strategy and its promising performance compared to state-of-the-art methods. 

\end{abstract}

This work is supported by the project ULEARN ``Unsupervised Lifelong Learning'' and co-funded under the grant number 316080 of the Research Council of Norway.

\section{Introduction}

\textit{Continual learning} can be described 
as the ability to continually learn over time by accommodating new knowledge while retaining previously learned experiences \cite{Thrun1998,Parisi2019}. We humans typically have no problem with retaining old experiences while at the same time being able to learn new tasks. For example: when a child learns to ride a bike, 
she does not forget the previous experience of learning how to walk.

In sharp contrast, standard machine learning algorithms typically assume that independent and identically distributed (i.i.d.) training examples of a task are given and use Empirical Risk Minimization (ERM) to learn a model for the task \cite{vapnik1999overview}. While this approach can be naturally extended to the setting in which samples arrive in an online fashion, when the task changes the conditional distribution of the data given the task also changes. As a consequence, the performance of the model on previously learned tasks significantly degrades when trained on new tasks, a phenomenon known as \textit{catastrophic forgetting}.

Existing methods that deal with catastrophic forgetting often assume that the moment the task changes and the identity of the task are known at training time. For instance, Averaged Gradient Episodic Memory (A-GEM) \cite{chaudhry2018efficient} and Orthogonal Gradient Descent (OGD) \cite{pmlr-v108-farajtabar20a} counteract catastrophic forgetting by solving a constrained optimization problem for each task change, which ensures that the loss function: a) decreases on the current task and b) does not increase on previous tasks. The constraints on previous tasks are enforced by storing either \emph{labeled data samples} (A-GEM) or \emph{model gradients} (OGD) from previous tasks as new tasks incrementally arrive. Thus, knowledge of a task change is needed to both solve the constrained optimization problem and update the pool of stored samples or gradients. Moreover, both A-GEM and OGD use pool size that grows with the number of tasks, making memory requirements prohibitive for a large number of tasks. While such memory requirements could be reduced by maintaining a constant and finite memory, this would inevitably lead to catastrophic forgetting as the number of tasks grows.

The aforementioned weaknesses raise two critical questions: 
\begin{enumerate}
\item \emph{Can we develop a memory and projection-based continual learning algorithm that does not require knowledge of task boundaries?} 
\item \emph{Can we address catastrophic forgetting more effectively for a large number of tasks while maintaining a constant and finite amount of memory?}
\end{enumerate}
%
\begin{figure*}
\centering
\includegraphics[width=0.92\textwidth]{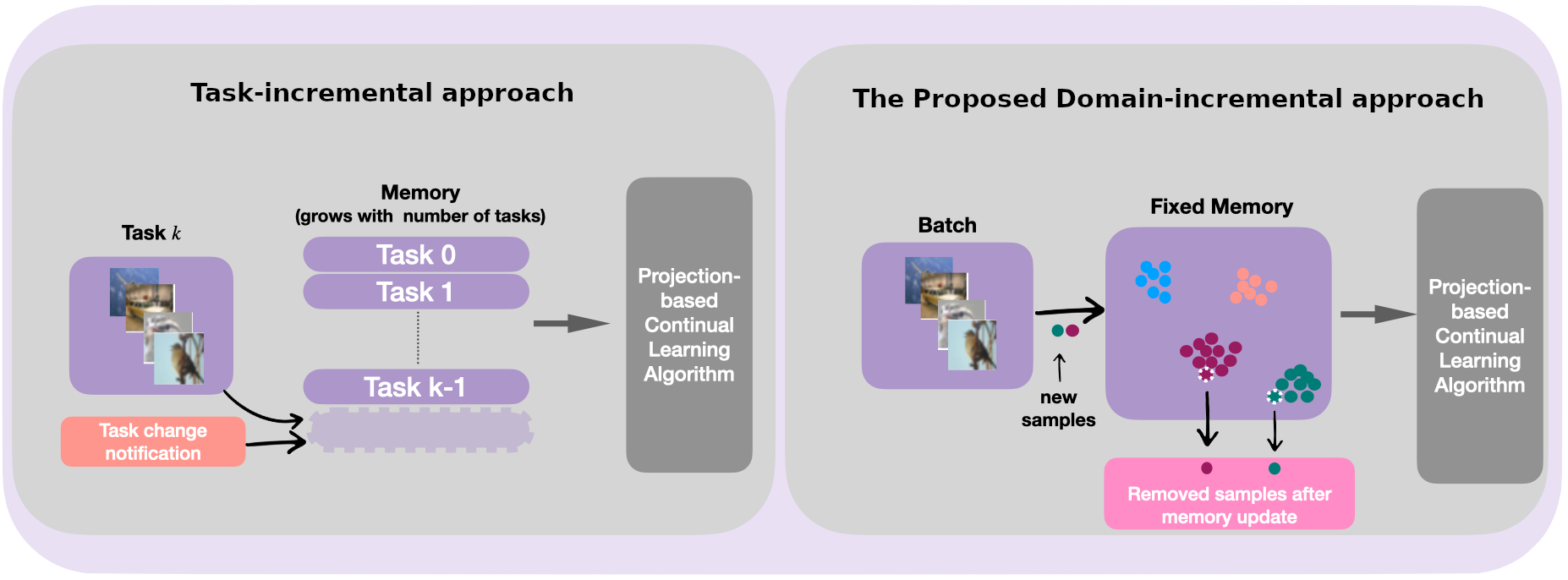}
\caption{After the task-incremental method is finished with the training on task $T_k$, the memory (containing either labeled data samples in the case of A-GEM or model gradients in the case of OGD) is updated. This method is made domain-incremental by using an online clustering-based approach for updating the memory while keeping its size fixed. 
}
\label{fig:OGDvsTAOGDhighlevel}
\end{figure*}
%
\paragraph{Paper contributions.}
In this work, we address these questions by proposing an online clustering-based approach that renders standard projection-based continual learning algorithms task-agnostic. This approach successfully counteracts forgetting in the setting of domain-incremental learning, a setting for which this problem was previously unsolved \cite{van2022three}.
The proposed approach is generic and can be applied to different projection-based algorithms. To showcase its merits, we focus on the A-GEM and OGD algorithms and propose two new task-agnostics versions called Task Agnostic Averaged Gradient Episodic Memory (TA-A-GEM) and Task Agnostic Orthogonal Gradient Descent (TA-OGD). These algorithms reduce the amount of forgetting when training on different tasks without the need to know any task boundaries and identities. 
This is achieved by dynamically updating the pool of \emph{labeled data samples} (A-GEM) or \emph{model gradients} (OGD) each time a new batch becomes available.
In addition, unlike A-GEM and OGD, which store a growing number of samples or gradients as the number of tasks increases, leading to prohibitive memory requirements in practical scenarios, the proposed TA-A-GEM and TA-OGD methods have constant and finite memory requirements by keeping a finite number of samples or gradients throughout the training process.
To achieve this, TA-A-GEM and TA-OGD leverage the structure of the training data, which are now grouped into clusters of samples or gradients. Specifically, for each new batch, we first uniformly draw samples or gradients from the current batch and use them to initialize a predefined number of clusters using the samples or gradients as the cluster centers. After initialization, new samples or gradients are assigned to the cluster center with minimum $\ell_2$ distance. To keep a constant memory, when the maximum cluster size is reached we remove less informative cluster members and update the cluster center with the average of the cluster members.

In short, this paper makes the following contributions:

\begin{itemize}[parsep=0pt,topsep=0pt]
\item We propose a generic clustering-based method for successfully extending projection-based continual learning algorithms to a task-agnostic context. 
We focus on two state-of-the-art projection-based algorithms i.e., A-GEM and OGD showing that the proposed strategy enjoys the merits of memory and projection-based methods \cite{pmlr-v108-farajtabar20a,lopez2017gradient,Doan2020} without requiring knowledge of the task identity or task changes.

\item By leveraging the structure of the data from previously seen tasks, we can retain the information needed to address catastrophic forgetting, such as training data (A-GEM) or model gradients (OGD), while keeping the memory-size finite via a simple and efficient clustering procedure.
We thus depart from the standard approach of OGD and A-GEM, which demand a growing amount of memory as new tasks sequentially arrive, which is impractical in real-world scenarios. 

\item We provide extensive experimental results for different continual learning settings on various datasets showing the promising performance of the proposed task-agnostic algorithms (TA-A-GEM and TA-OGD) compared to state-of-the-art methods.
\end{itemize}
%
%
\section{Related Work}
\label{sec:related}
 This section starts with an explanation of the three types of incremental learning. It then reviews the stability-plasticity dilemma, which continual learning methods have to face. Moreover, we present the main ideas of memory and projection-based continual learning approaches to which class the proposed TA-A-GEM and TA-OGD method belong and the main advances in task continual learning. Finally, we review the recent works leveraging representation learning for deriving efficient continual learning algorithms.

\subsection{Domain-incremental learning}
In continual learning, different tasks can arrive in sequence. The learner must therefore learn new tasks incrementally. This is referred to as \textit{incremental learning}. Three types of incremental learning can be specified: \textit{task-incremental learning}, \textit{domain-incremental learning} and \textit{class-incremental learning} \cite{van2022three}.
In task-incremental learning, the task identity is known to the learner during the training and testing phase. In domain-incremental learning, the task identity is not known to the learner at both training and testing time. In class-incremental learning, the learner must learn to identify a growing number of classes. 
Since we focus on a scenario where the number of classes is static and the task identity is not known during training and testing, we focus on the \textit{domain-incremental} setting. Alleviating catastrophic forgetting in such a scenario is an important unsolved challenge \cite{van2022three}.

\subsection{The Stability-Plasticity Dilemma}
The balancing act between being able to gain new knowledge while assuring old knowledge is not lost is referred to as the \textit{stability-plasticity dilemma} \cite{mermillod2013stability}. Continual learning approaches can be categorized in three major trends based on how the stability-plasticity dilemma is  handled \cite{de2021continual,Parisi2019}. The first trend is to use the concept of \textit{regularization} of synaptic plasticity, where the plasticity of important weights is constrained in order to retain old skills, like the Memory Aware Synapses used in a continual setting in \cite{aljundi2019task}. Elastic Weight Consolidation (EWC) is a seminal work of this class. When a new task arrives, EWC learns the optimal weights for this task, while penalizing changes of the weights towards values that are far from the optimal ones for the previous task \cite{kirkpatrick2017overcoming}. Several other variants of EWC have appeared in the literature and we refer the readers to \cite{de2021continual} for a detailed review. The second trend is \textit{expansion} \cite{rusu2016progressive,aljundi2017expert,mehta2021continual,douillard2022dytox}, where a neural network is expanded by allocating new neural resources in order to gain new skills, while leaving old neurons unchanged in order to retain old skills. Finally, according to the third trend, which is \textit{repetition}, old information is repeatedly fed to the network, along with new information. This can be implemented by applying a complementary learning system for integrating old and new skills and applying experience replay, or by simply mixing old and new data in the training step. In the literature, various approaches of the so-called replay-based methods which rely on the principle of repetition have come to the scene. These methods make use of memory resources and vary in the strategy they follow \cite{rebuffi2017icarl,lopez2017gradient,shin2017continual,chaudhry2019tiny,aljundi2019online,van2020brain, koh2021online, ye2022task}.

This paper uses the terms ``replay-based'' and ``memory-based'' interchangeably because they represent similar concepts. Still, we tend to favor ``replay-based'' when a method stores samples from the dataset and ``memory-based'' when it stores different information. The proposed TA-A-GEM builds on A-GEM \cite{chaudhry2018efficient}, which stores samples from the training set, and can thus be considered ``replay based''. The proposed TA-OGD builds on OGD \cite{pmlr-v108-farajtabar20a}, and thus, in principle, falls into the category of memory-based methods since it stores gradients. At the same time, the proposed TA-A-GEM and TA-OGD use a projected gradient step and, hence, are also a projection-based approach. Note that this projection step implicitly regularizes the weights; therefore, A-GEM and OGD bear similarities with the regularization-based methods. Next, we elaborate on the specific class of memory-based and projection-based continual learning algorithms. 

\subsection{Memory-based and Projection-based Continual Learning Methods}
Over the last few years, several memory-based and projection-based methods have been proposed in the literature, \cite{lopez2017gradient,pmlr-v108-farajtabar20a}. These make use of memory for storing information from the past, which helps to update the model towards non-forgetting directions. The goal is to address catastrophic forgetting by means of imposing certain constraints on the weight-updating process.
Many different approaches have appeared in the literature over the last few years. In \cite{lopez2017gradient}, the authors propose to update weights in directions that do not increase the loss function values on samples of previously seen tasks. The resulting algorithm, dubbed  Gradient Episodic Memory (GEM), thus stores a predefined number of gradients of the loss function corresponding to old tasks, \cite{chaudhry2018efficient,lopez2017gradient}. These are then used for updating the model by solving a constrained optimization problem. Orthogonal Gradient Descent (OGD) \cite{pmlr-v108-farajtabar20a} stores a growing number of gradients of the model corresponding to old tasks' samples. In the weight update step, it projects its loss gradient to a direction that is orthogonal to all stored gradients. Specifically, gradients of the loss are projected on the orthogonal basis spanned by the stored gradients. In doing so,  directions that increase forgetting of past tasks are excluded when the model learns a new task. This assumes however that the stored gradients remain relevant, even when the weights of the model move during the training process, thus arriving at a different point in the configuration space in which older tasks can have different gradients. Averaged Gradient Episodic Memory (A-GEM) \cite{chaudhry2018efficient} solves this problem by storing labeled data samples instead of gradients. It projects the loss gradient orthogonal to a reference gradient that is calculated at every training step from a subset of the stored labeled data. Though showing promising performance in addressing catastrophic forgetting, memory-based and projection-based methods suffer from two fundamental weaknesses: a) they require the moment of task change to be available in order to know when the memory should be updated, and b) memory cost should either scale with the number of tasks, e.g., in OGD \cite{pmlr-v108-farajtabar20a}, which is infeasible in real-world scenarios, or the stored data per task will decrease as in the case of GEM \cite{lopez2017gradient}, which also hinders the ability of the algorithm to address forgetting when it encounters a large number of tasks.



\subsection{Task Agnostic Continual Learning}
Task boundaries and identities are rarely available in practical continual learning applications. In light of this, various task-agnostic continual learning methods have been proposed in the literature. In \cite{Harrison2020}, the authors propose an auxiliary mechanism to detect tasks while counteracting forgetting. The resulting method operates in a task-agnostic environment showing promising empirical performance. Several other approaches have been proposed in the same spirit \cite{Caccia2020, He2019}.  Another line of work hinges on online learning ideas completely neglecting task identity or the need to know the moment of task change. In \cite{Zeno2018}, the authors propose Bayesian Gradient Descent (BGD), an online variational Bayes approach in which model parameters with low variance are considered more important for previous tasks and, thus, are less updated. The opposite holds for parameters with high variance (hence high uncertainty). A similar idea for task-free continual learning appeared in \cite{Aljundi}. Namely, the authors modified the so-called Memory Aware Synapses (MAS) algorithm in \cite{aljundi2018memory}, in order to operate in a task-agnostic online learning setup. For, they use an importance weight regularizer which penalizes changes to model parameters which negatively affect model performance on prior tasks. Finally, in  \cite{Jin2020} the authors propose an online task-agnostic memory-based method. The main idea is to edit the stored-in-memory gradients used for addressing forgetting by solving an optimization problem in an online fashion. Recently, the idea of using self-supervised representations for task-agnostic continual learning was proposed in \cite{pham2021dualnet}, showing promising empirical performance. 

Though the emergence of clustering in episodic memory has been recently acknowledged in the child development literature \cite{horn2021development}, to the best of our knowledge, the proposed TA-A-GEM and TA-OGD are the first algorithms that use online clustering for dynamically updating the memory of continual learning methods. While we focus on A-GEM and OGD, the adopted strategy could be applied to other memory-based and task-dependent continual learning approaches for allowing them to operate in task-agnostic environments.
%

\subsection{Representation Learning}
Representation learning aims to find insightful data representations by exploiting their structure \cite{ma2022principles}. Recently,  learned representations have been at the heart of several continual learning algorithms.  In \cite{chaudhry2020continual}, the authors employed low-rank orthogonal subspace representations of the model parameters formulating continual learning as an optimization over the Stiefel manifold problem. The reported results showed promising performance and the ability of the approach to counteract forgetting. In \cite{guo2022online}, {\it holistic} representations learned via a mutual information maximization criterion were employed in the continual learning setting. The method can learn feature representations of the current task that are useful for the future tasks, hence leading to models that are more robust to forgetting. In \cite{Doan2020}, a variant of the projection-based OGD method was proposed. The main idea is to perform principal component analysis on the set of stored gradients of the model and keep only the most informative principal components. However, the work in \cite{Doan2020}, still assumes that task changes are provided to the algorithms and batch processing is utilized. Hence it is far from our proposed online clustering-based task-agnostic algorithms. 
\section{Proposed Approach}
We assume that the $n$ tasks $\{T_i\}_{i=1}^n$ arrive sequentially and that during task $T_k$ the data from tasks $T_i$ for $i<k$ are not presented to the learner. Each task consists of pairs of data points $(x,y)\in T_k$, where $x\in \mathbb{R}^d$ is the input and $y$ is a label. Here we assume that each task is a classification task and that all classification tasks share the same classes $j=1,\dots, c$, where $c$ is the number of classes. Therefore, we can represent $y\in \mathbb{R}^c$ as a one-hot class encoding vector, i.e., $y_j=1$ when $j$ is the class label and $y_j=0$ otherwise. We denote the network model as $f(x;w)\in \mathbb{R}^c$, where $w \in \mathbb{R}^p$ denotes the $p$-dimensional weights (parameters) of the network and $f_j(x;w)$ is the $j$-th logit corresponding to the $j$-th class. The model is trained to predict the class label for input $x$.

The proposed Task Agnostic Averaged Gradient Episodic Memory (TA-A-GEM) and Task Agnostic Orthogonal Gradient Descent (TA-OGD) methods rely on the forgetting counteracting mechanisms of Averaged Gradient Episodic Memory (A-GEM) \cite{chaudhry2018efficient} and Orthogonal Gradient Descent (OGD) \cite{pmlr-v108-farajtabar20a}, respectively.
Next, we briefly describe the main ideas behind A-GEM and OGD and refer the reader to the Appendix or \cite{chaudhry2018efficient} and \cite{pmlr-v108-farajtabar20a} for further details.
%

Both A-GEM and OGD assume the identity $k_t$ of the task $T_{k_t}$ at time step $t$ is known. The empirical loss, during time step $t$, with a batch size $|T_{k_t}|$, is given by,
\begin{equation}\label{eq:softmaxloss}
    L_t(w) = \frac{1}{|T_{k}|}\sum_{(x,y)\in T_{k}} L_{(x,y)}(w),
\end{equation}
where the per sample loss $L_{(x,y)}(w)$ is assumed to be the cross-entropy, which is
defined as
\begin{equation}\label{eq:softmax}
L_{(x,y)}(w) = - \sum_{j=1}^{c} y_j \log \left(\frac{\exp f_j(x;w)}{\sum^c_{m=1} \exp f_m(x;w)}\right).
\end{equation}

Both A-GEM and OGD use a pool of samples to counteract the catastrophic forgetting. The difference is that OGD stores network gradient, while A-GEM stores training data.

\subsection{Clustering-based Task Agnostic A-GEM (TA-A-GEM) and OGD (TA-OGD)}
\label{sec:method}

Figure \ref{fig:OGDvsTAOGDhighlevel} shows our strategy to convert a task-aware task-incremental projection algorithm to a task-agnostic domain-incremental algorithm. Task-incremental projection algorithms like A-GEM and OGD keep a pool of samples from either the training data or model gradients, respectively. This pool of samples is used to mitigate catastrophic forgetting of previous tasks through projection. When the algorithm is finished with training on one task, it stores samples from this task before it starts training on the new task. In this way, it ensures that the samples in the pool are relevant for previous tasks when addressing forgetting. However, this comes at the cost of {\it requiring to know the moment a task changes}.
In our approach, we make this process task-agnostic by updating the pool of samples during the process of training, i.e. {\it the pool of samples is updated every time the model is trained on a batch}. This removes the need to know the moment the task changes but introduces the problem that the size of the pool now grows more rapidly. However, {\it our goal is to keep the memory requirements constant in the number of tasks}. Hence, a strategy is necessary to decide which samples should be added to the pool and which ones should be removed during the updating process. Our strategy aims to select stored samples in a way that addresses forgetting all previous tasks in the most efficient way while being constrained by constant and finite pool size.
Because we aim for a true task-agnostic setting, all tasks are made to have the same label space, so the task identity can not be inferred from the labels.

Next, we detail the proposed online clustering-based approach that consists of the following four steps:

1) {\it Initialization:} We first set the number of clusters $Q$ and consider the first $Q$ samples becoming available as the centers $\boldsymbol{\mu}_i, q=1,2,\dots, Q$ of these clusters. 

2) {\it Cluster assignment:} A new sample $\mathbf{z}_p$ (corresponding to a training sample in the case of A-GEM or gradient logit in the case OGD) is assigned to the cluster $q^\ast$ that minimizes the $\ell_2$ norm i.e.,
\begin{equation}
    q^\ast = \mathrm{argmin}_{q\in\{1,2,\dots,Q\}}\|\mathbf{z}_p - \boldsymbol{\mu}_q\|^2_2
\end{equation}

3) {\it Memory update:}
The size of each cluster is predefined, and once the maximum size has been reached, for new samples to that assigned to that cluster  an equal number of older samples residing in the cluster should be removed. Note that the process of accepting/rejecting new samples and deciding which ``old'' samples to delete could be implemented using information-theoretic criteria or rejection sampling-based ideas. Here, in an effort to simplify the approach and make it computationally efficient, we follow a first-in-first-out (FIFO) approach. This dictates that samples that arrived first in the cluster are the first to be removed. Note that the strategy followed ensures that samples corresponding to a task with information distinct from other tasks will not be deleted from the pool. This will occur since these samples will ``live'' within clusters that will not be updated and thus remain unaffected by the memory updating process.\footnote{Empirical findings reported in the Appendix corroborate our hypothesis.}

3) {\it Update of cluster means:} Once samples are assigned to the clusters and the memory has been updated, the cluster means are re-computed i.e.,
\begin{equation}
    \boldsymbol{\mu}_q = \frac{1}{P}\sum^P_{p=1} {\mathbf{z}^q_p}, \;\;\;\;\;\;\;\;\;\; \forall i=1,2,\dots,N,
\end{equation}
where $P$ denotes the size of the clusters and $\mathbf{z}^q_p$ the $p_{th}$ element of cluster $q$. For the case of the task-agnostic version of A-GEM, i.e., TA-A-GEM, we have $\mathbf{z}_p\equiv \mathbf{x}_p \in \tilde{M}_t$ (where $t$ here denotes the batch index) whereas for the task-agnostic OGD algorithm (TA-OGD) $\mathbf{z}_p\equiv \nabla f_j(\mathbf{x}_p,w^*_{i})$. Our clustering-based strategy is depicted at Fig. \ref{fig:clustering}, while a pseudo-code of the algorithm is given in the Appendix.

{\it A single or a different pool for each class?}
A possible complication that can occur is that more similarity exists between samples of the same class that are of a different task than between different classes of the same task. If this happens, {\it class} information will be well represented in the pool, but {\it task} information can be easily lost. Since class labels of the samples are available, a way to get around that issue and disentangle the class from task information is to use a different pool for each class.
In that case, samples are first assigned to a pool based on their class label. Then, the procedure described above is independently followed for each pool. It is worth noting that this is critically important for the task-agnostic version of A-GEM (TA-A-GEM) since the pool contains training samples of different classes. Samples corresponding to the same class but different tasks, e.g., a digit and its rotated version might be close in the input space. As a result, if a single pool is used, those two samples will be assigned to the same cluster, and hence task information will be lost. This phenomenon is more likely not to be observed in the case of TA-OGD since clustering takes place in the space of model gradients, which are sufficiently separated for different tasks even for samples corresponding to same classes.



{\it The role of hyperparameters:} The choice of hyperparameters, such as the number of clusters $Q$ and their size, is important. A large number of clusters $Q$, allows more task and class diversity to be stored in different clusters in memory.
The size of the clusters should be large enough so it can capture the essence of a specific task. However, the size of $Q$ and the cluster size should be kept as small as possible to reduce the memory footprint. A trade-off can be made where $Q$ is large, and the cluster size is small versus using a small $Q$ with a large cluster size. In addition, we follow an adaptive strategy for the learning rate of the projected gradient step. Note that this is a form of task detection that our method does not necessarily need. Our focus is to create a truly task-agnostic method without any task detection.
Specifically, 
the learning rate $\eta^t$ at iteration $t$ decreases as follows:
\begin{equation}
    \eta^t = a \eta^{t-1},
\end{equation}
where $a<0$, when the loss function is {\it smoothly} increasing for a given number of iterations. This allows the algorithm to update the weights of the model following a non-increasing path for the loss function. Moreover, when a sudden increase is observed, then the learning rate is reset to its initial value (therefore increases), i.e., $\eta^t = \eta_{ini}$. The reasoning behind this rule is that spikes of the loss most likely imply task-change and therefore, a higher learning rate can help to move fast along decreasing directions of the loss corresponding to the new task. Empirical results on the effect of the sampling rate, the number, and the size of clusters on the performance of the proposed method, and more details on the adaptive updating process of learning rate, are provided in Section 4 and Appendix. 
%
%
%
%

\begin{figure*}
\centering
\includegraphics[width=0.9\textwidth]{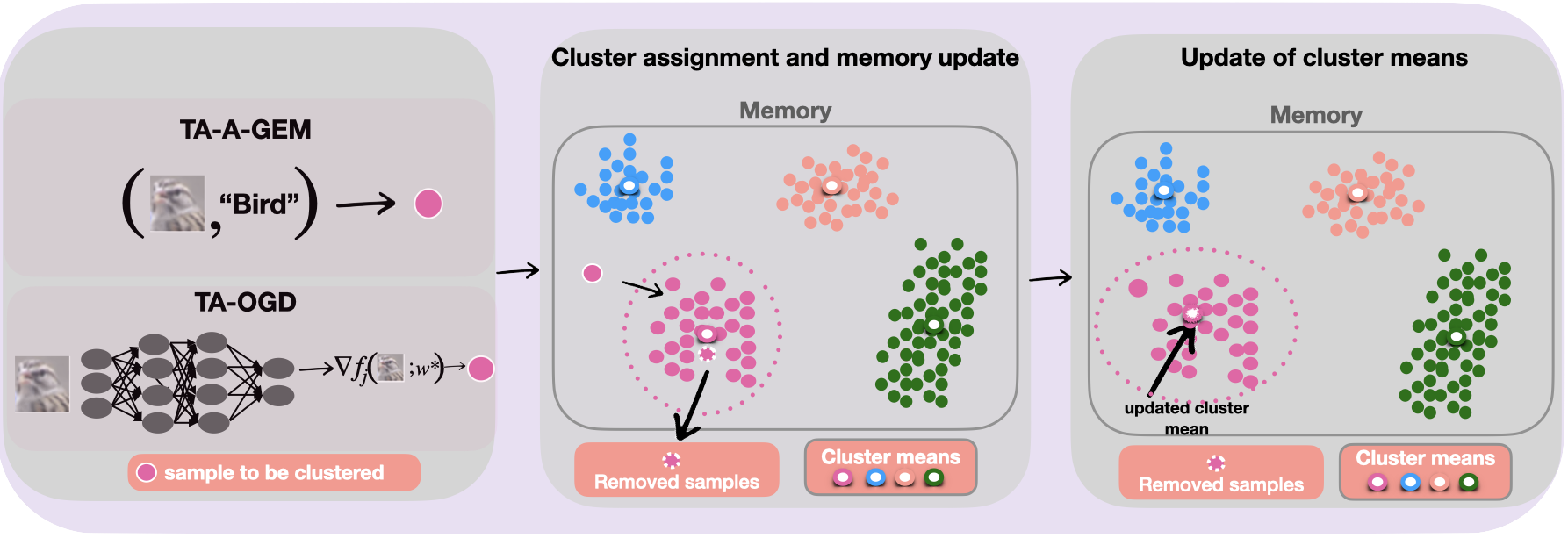 }
\caption{The clustering mechanism to add training set samples / model gradient samples  to the memory by matching it to the closest cluster (pink cluster), as used by TA-A-GEM / TA-OGD.}
\label{fig:clustering}
\end{figure*}

\section{Experiments}
\label{sec:experiments}
We divide the experiments into two main classes: a) the {\it disjoint tasks experiment} and b) {\it the continuous change experiments}. 
The task-aware methods are notified of the task change, while the task-agnostic methods do not get this information.
In the continuous change experiments, discrete tasks still exist, but task boundaries are no longer clearly defined. Details on the experimental setting can be found in the Appendix.
Since there is no clear point that a task-aware method can be notified, only task-agnostic methods are included in this experiment. 
For both methods, all tasks are made to have the same label space, since it should not be possible to infer the task identity from the labels. In cases where the label spaces are disjoint, the labels are cast to the same label space. Since no task identity is provided during training, the method is tested in a domain-incremental setting \cite{van2022three}.
Following empirical observations, we use the learning rate scheduler described in Section 3.3 for the case of OGD and the proposed task-agnostic version of it i.e., TA-OGD. 
%
The network used for training is a multi-layer perceptron (MLP) with two hidden layers of 200 nodes. To compare the performance of the tested methods, we use three metrics: a) The {\it validation accuracy}, b) The {\it average validation accuracy} over all tasks trained on thus far and c) The amount of {\it forgetting}. For an exact mathematical definition of these quantities , we refer to the Appendix.
To create separate tasks from existing datasets, three task generation mechanisms are implemented: a) task permutation, b) task rotation and c) class splitting. For the details of this task generation, we refer to the Appendix.


\subsection{Disjoint tasks experiment}
Table \ref{tab:disjoint_avg_acc} shows the results of the first class of experiments. It shows the average accuracy over all tasks trained thus far, thereby capturing both the ability to remember old tasks and the ability to learn new tasks. The average accuracy was then averaged over 20 epochs, then over five runs. Plots of these results can be found in the Appendix.
Our proposed TA-OGD and TA-A-GEM algorithms significantly outperform the state-of-the-art task-agnostic BGD algorithm, \cite{Zeno2018}, on the MNIST \cite{deng2012mnist}, Fashion MNIST \cite{xiao2017fashion} and NOT MNIST datasets. Moreover, their performance is comparable to BGD on CIFAR10 and SVHN.
Focusing on MNIST, Fashion MNIST and NOT MNIST, we observe that at the \textit{permutation experiments}, no remarkable differences can be seen among the methods. This can be explained by the fact that the baseline SGD method shows little signs of forgetting in the first place.
For the \textit{rotation experiments}, A-GEM is a clear winner, it is however not task-agnostic. On MNIST and NOT MNIST, TA-OGD and TA-A-GEM are moderately effective at mitigating forgetting. On Fashion MNIST however, TA-A-GEM is clearly the best method among all the tested task-agnostic methods. 
We attained the most remarkable results on the \textit{class split experiments}. On MNIST, both TA-OGD and TA-A-GEM clearly outperform the other task-agnostic methods. On Fashion MNIST, TA-A-GEM's performance is even on par with A-GEM, while on NOT MNIST, TA-OGD takes the crown by performing on par with A-GEM, which is a task-aware method.

\begin{table*}
\centering
\scalebox{0.75}{
\begin{tabular}{c|ccc|ccc|ccc|ccc|ccc}
    & \multicolumn{3}{c|}{MNIST} & \multicolumn{3}{c|}{Fashion MNIST} & \multicolumn{3}{c|}{NOT MNIST} & \multicolumn{3}{c|}{CIFAR10} & \multicolumn{3}{c}{SVHN}\\
    & perm & rot & class & perm & rot & class & perm & rot & class & perm & rot & class & perm & rot & class \\
    \hline
    SGD & 0.842 & 0.657 & 0.804 & 0.735 & 0.468 & 0.807 & \textbf{0.850} & 0.598 & 0.864 & 0.373 & 0.346 & 0.724 & \textbf{0.599} & 0.392 & 0.759\\
    SGD lr adapt & 0.842 & 0.663 & 0.811 & 0.736 & 0.469 & 0.820 & \textbf{0.851} & 0.598 & 0.888 & 0.376 & 0.347 & \textbf{0.727} & \textbf{0.596} & 0.390 & 0.762\\
    BGD & \textbf{0.883} & 0.682 & 0.790 & \textbf{0.765} & 0.507 & 0.802 & \textbf{0.856} & \textbf{0.633} & 0.875 & \textbf{0.385} & \textbf{0.357} & 0.718 & \textbf{0.591} & \textbf{0.417} & 0.763\\
    TA-OGD & \textbf{0.871} & \textbf{0.705} & 0.857 & 0.749 & 0.516 & 0.893 & 0.845 & 0.625 & \textbf{0.937} & 0.328 & 0.343 & \textbf{0.731} & 0.547 & 0.393 & \textbf{0.773}\\
    TA-A-GEM & \textbf{0.876} & 0.688 & \textbf{0.878} & 0.746 & \textbf{0.604} & \textbf{0.931} & \textbf{0.853} & 0.602 & 0.884 & 0.365 & 0.343 & \textbf{0.726} & \textbf{0.605} & 0.399 & \textbf{0.772}\\
    \hline
    \color{gray}{OGD} & \color{gray}0.865 & \color{gray}0.690 & \color{gray}0.822 & \color{gray}0.757 & \color{gray}0.512 & \color{gray}0.839 & \color{gray}0.846 & \color{gray}0.627 & \color{gray}0.925 & \color{gray}0.360 & \color{gray}0.348 & \color{gray}0.731 & \color{gray}0.587 & \color{gray}0.400 & \color{gray}0.768\\
    \color{gray}{A-GEM} & \color{gray}0.884 & \color{gray}0.806 & \color{gray}0.952 & \color{gray}0.761 & \color{gray}0.706 & \color{gray}0.934 & \color{gray}0.854 & \color{gray}0.740 & \color{gray}0.947 & \color{gray}0.360 & \color{gray}0.356 & \color{gray}0.741 & \color{gray}0.552 & \color{gray}0.451 & \color{gray}0.827\\
\end{tabular}
}
\caption{\textbf{Average validation accuracy}, averaged over all tasks trained thus far, then averaged over all epochs, then averaged over five runs, for the \textbf{disjoint tasks experiments} when using a MLP. Per column, the best result for the \textit{task-agnostic} methods are written in bold. In case a task-agnostic method's result is less optimal and not significantly different from the best result, with a confidence of 99\%, it is also written in bold. The results for the \textit{task-aware} methods OGD and A-GEM are given for context. Since these algorithms benefit from knowing task identities and changes, we just use them here as baselines for indicating the best performance we can achieve.}
\label{tab:disjoint_avg_acc}
\end{table*}
\subsection{Continuous task change experiment}
The results of the \textit{continuous change experiments} are extremely similar to the results in the \textit{disjoint tasks experiments}. They can be found in the Appendix. These experiments show that the proposed TA-OGD and TA-A-GEM fare just as well in the challenging setting where task boundaries are blurred.
\subsection{Effectiveness of the clustering-based procedure}
\label{subsec:clusteringanalysis}
In order to demonstrate the benefits obtained by the proposed clustering-based approach, we compared the performance of TA-A-GEM with and without clustering. To deactivate clustering we skipped the cluster assignment step and new samples were randomly to allocated clusters. Similarly to our approach, an equal number of old samples of update clusters is removed to keep the memory size constant.  For this experiment, a MLP was trained on Fashion MNIST, with the task split segmentation. All settings are the same as in the \textit{disjoint tasks experiments}.

\begin{figure}
\hspace{-1cm}
\includegraphics[width=.56\textwidth]{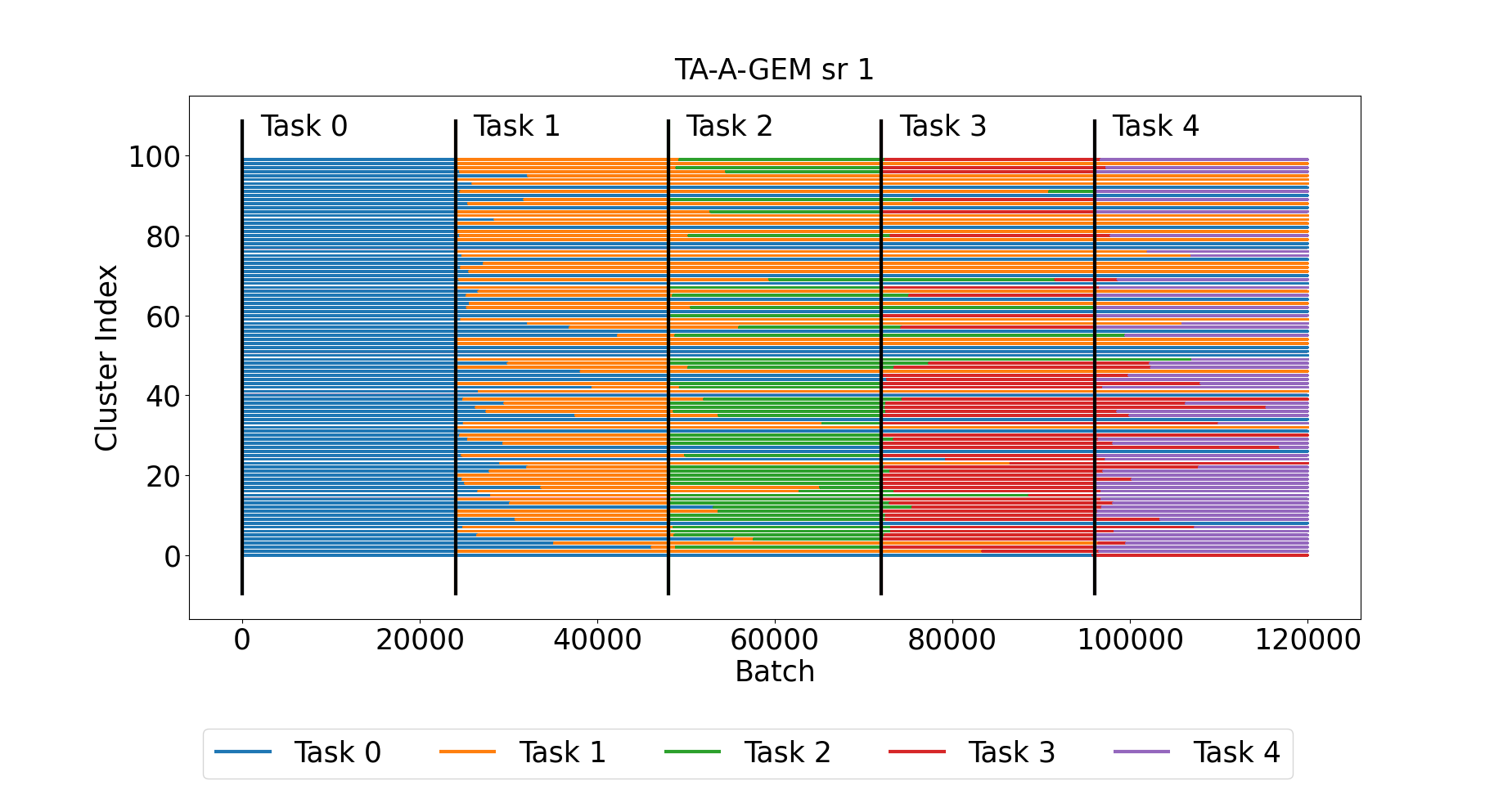} 
\caption{Clustering helps us address forgetting: samples from old tasks remain in the pool even after the of training on Task 4 ends.}\vspace{-0.65cm}
\label{fig:analyze_clusters_1_clustering}
\end{figure}

\begin{figure}
\hspace{-1cm}
\includegraphics[width=.56\textwidth]{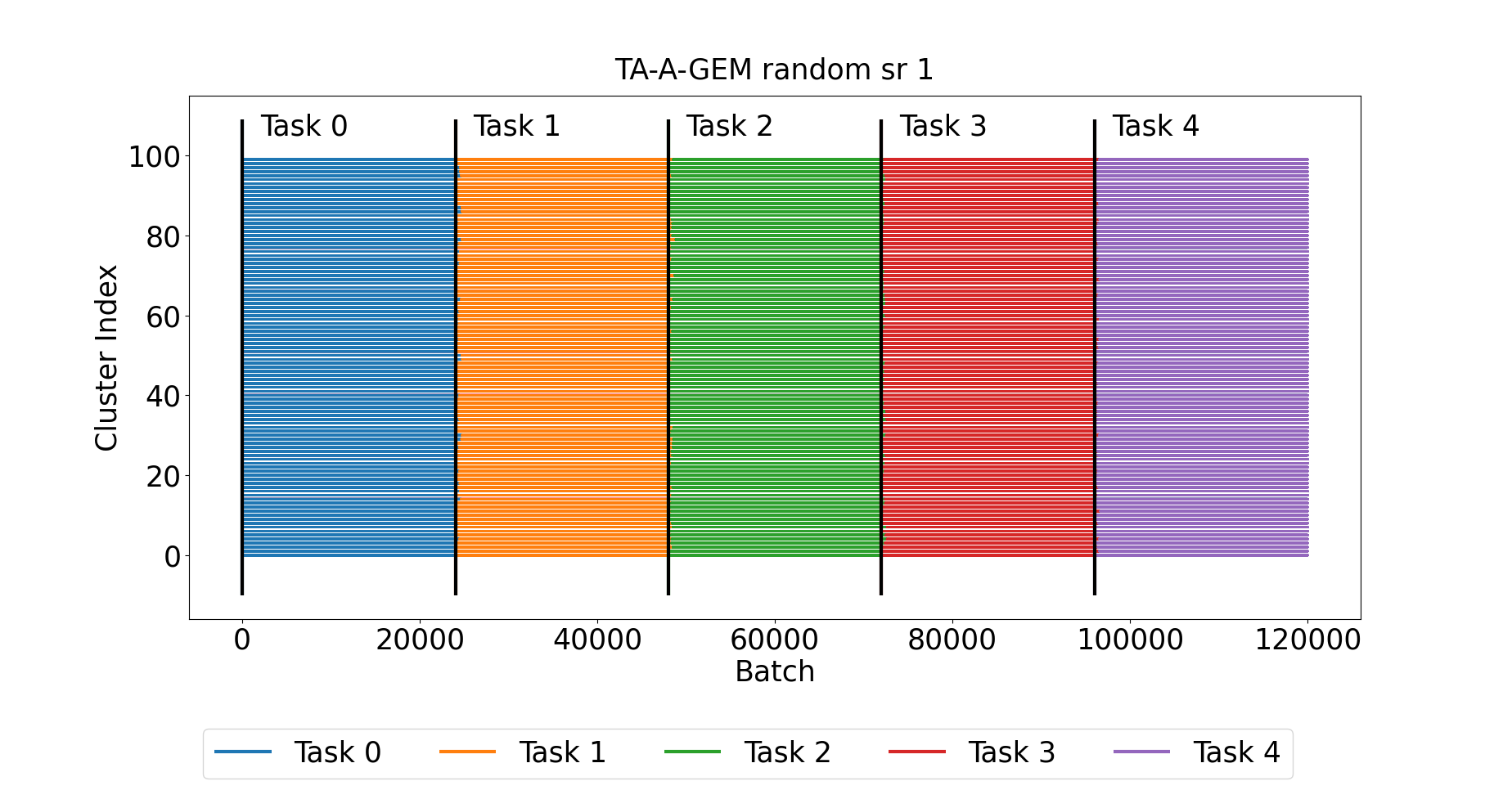}
\caption{When using random cluster assignment, the information of old tasks is almost immediately lost, once a new tasks starts.}
\label{fig:analyze_clusters_1_random}
\vspace{-0.65cm}
\end{figure}

Figure \ref{fig:analyze_clusters_1_clustering} and \ref{fig:analyze_clusters_1_random} show the content of each cluster during training time. Each horizontal line corresponds to a cluster. Each task is associated with a unique color, which represents the oldest task information that is present in the cluster. The horizontal line changes color the moment that the last information of the oldest task disappears from the cluster. Then, the second oldest task information becomes the new oldest task information. The moment that a new task starts -not available to the algorithms- is indicated by a black vertical line. As it can be observed in Figs \ref{fig:analyze_clusters_1_clustering} and  \ref{fig:analyze_clusters_1_random}, clustering helps in keeping a greater variety of task information in the gradient pool, with samples from Task 0 or Task 1 still being present in clusters even after the end of training on samples from Task 4. On the other hand, the use of random cluster assignment results in information of old task being almost immediately lost after a task change, thus illustrating the merits of our proposed clustering-based approach.

\section{Conclusions and future directions}
In an effort to counteract catastrophic forgetting in a task-agnostic setting, we proposed a clustering-based strategy to make task-aware projection methods task-agnostic with constant memory requirements. 
By leveraging the structure in the sampled data (in the case of TA-A-GEM) and model gradients (in the case of TA-OGD), we can effectively counteract catastrophic forgetting without providing knowledge of a task change and the need of a growing amount of memory. Extensive experimental results provided in section \ref{subsec:clusteringanalysis} and the Appendix show the benefits of our clustering-based method. 
As a future direction, we aspire to explore more sophisticated, yet computationally efficient, methods for the clustering and memory update step.
Our goal is also to illustrate the merits of our method on larger networks such as a ResNet \cite{he2016deep}, or more complicated datasets such as ImageNet \cite{deng2009imagenet}. It is worth noting that our proposed method is generic hence we also intend to inquire its application as an off-the-shelf tool to other projection-based methods. 


{
\bibliographystyle{unsrtnat}
\bibliography{egbib}
}

\clearpage

\appendix

\section{Averaged Gradient Episodic Memory}
A-GEM uses a projected gradient descent-type approach for updating the weights of the model. It first calculates the gradient of the loss function $\nabla L_t(w)\in \mathbb{R}^p$ and then compares this to a reference gradient $\nabla L_{ref}(w) \in \mathbb{R}^p$, of the loss function $L_{ref}(w)$ as defined by
\begin{equation}
    L_{ref}(w) = \frac{1}{|\tilde{M}|}\sum_{(x,y)\in \tilde{M}} L_{(x,y)}(w),
\end{equation}
where $\tilde{M}$ is a randomly selected batch of data sampled from the episodic memory $M$ which contains label training data from tasks seen before time $t$ i.e., $M = \cup_{k<t} M_k$, where $M_k$ is the episodic memory for task $k$. 
When the inner product between the gradient of the loss function $\nabla L_t(w)$ and the reference gradient $\nabla L_{ref}(w)$ is non-negative, the gradient update boils down to a vanilla gradient descent step without projections. 
%
However, when this inner product is less than zero, the weights of the model are updated using a projected gradient-type update, i.e.,  
\begin{equation}
    w \gets  w - \eta\tilde{g} \label{update:ta-a-gem}
\end{equation}
where $\eta$ denotes the step size and $\tilde{g}$ is defined as,
\begin{equation}
    \tilde{g} = \nabla L_t(w) - \frac{\nabla L_t(w)^\top \nabla L_{ref}(w)}{\nabla L_{ref}(w)^\top \nabla L_{ref}(w)}\nabla L_{ref}(w).
    \label{referencegradientprojection}
\end{equation}
Note that $\tilde{g}$ is now orthogonal to the reference gradient $\nabla L_{ref}(w)$; thus, if $\nabla L_{ref}(w)$ successfully represents a non-forgetting direction of previously sees tasks, then the update rule of the weights of the model in \eqref{update:ta-a-gem}, decreases the loss on the current task without incurring any forgetting of previous tasks \cite{chaudhry2018efficient}. 

Algorithm \ref{alg:AGEMedit} shows pseudo-code for A-GEM \cite{chaudhry2018efficient}. It uses both a loss gradient $\nabla L(w)$ and a reference gradient $\nabla L_{ref}(w)$ that is computed using a sub sample of the episodic memory, namely $\tilde{M}$. If the dot product between the loss gradient and the reference gradient is greater or equal to zero, gradient descent proceeds as normal. However, if it is smaller than zero, the loss gradient is projected onto the reference gradient and the resulting component is subtracted from the loss gradient. The weights $w$ are then updated using this newly obtained gradient and the step size $\eta$. After the training on task $k$ finishes, $s$ samples are taken from the data of task $T_k$ and added to the episodic memory $M$.

\begin{algorithm}
\caption{Averaged Gradient Episodic Memory}\label{alg:AGEMedit}
\hspace*{\algorithmicindent} \textbf{Input:} Task sequence $T_1, T_2, T_3, ...$ learning rate $\eta$\\
\hspace*{\algorithmicindent} \textbf{Output:} Model
\begin{algorithmic}[1]
\State \textbf{Initialize} $M \gets \{\}; w \gets w_0$
\For{Task ID $k = 1, 2, 3, ...$}
    \Repeat
        \State $L(w) \gets \frac{1}{|T_{k_t}|}\sum_{(x,y)\in T_{k_t}} L_{(x,y)}(w)$
        \State $L_{ref}(w) \gets \frac{1}{|\tilde{M}|}\sum_{(x,y)\in \tilde{M}} L_{(x,y)}(w)$
        \If{$\nabla L(w)^\top \nabla L_{ref}(w) \geq 0$}
            \State $\tilde{g} \gets \nabla L(w)$
        \Else
            \State $g_{proj} \gets \frac{\nabla L(w)^\top \nabla L_{ref}(w)}{\nabla L_{ref}(w)^\top \nabla L_{ref}(w)}\nabla L_{ref}(w)$
            \State $\tilde{g} \gets \nabla L(w) - g_{proj}$
        \EndIf
        \State $w \gets w - \eta \tilde{g}$
    \Until{converge}
    \For{$i = \{1, ..., s\}$}
        \State $(\mathbf{x}, y) \sim T_k$
        \State $M \gets M \cup (\mathbf{x}, y)$
    \EndFor
\EndFor
\end{algorithmic}
\end{algorithm}

\section{Orthogonal Gradient Descent}
Orthognal Gradient Descent (OGD) departs from A-GEM, which uses the gradient of the loss function, and performs projections leveraging gradients of the output $f(x;w)$ of the model denoted as $\nabla f(x;w)\in \mathbb{R}^{p\times c}, x\in T_{k_t}$.
We also denote as 
\begin{equation}
\nabla f_j(x;w) \in \mathbb{R}^p, (x, y) \in T_{k_t}
\label{modelgradient}
\end{equation}
the gradient associated with the output logit $j$:
\begin{equation}
    f_j(x;w) \in \mathbb{R}, (x,y) \in T_{k_t}, 
\end{equation}
i.e. the $j$-th output of the final layer, where $j$ again is the class label. For simplicity, from now on, we refer to \eqref{modelgradient} with the term \textit{model gradient}.

OGD uses projected gradient-type updates in order to update the model parameters for a new task in directions that minimally increase the loss for samples from previously seen tasks. Specifically, after training on task $k$ is finished, an update takes place that stores the relevant model gradients for task $k$ in the model gradient~pool. 

Assume our model has been trained on task $T_1$ using SGD and denote as $w^\ast_1$ the weights learned through the training process. The gradient pool after the training on task $T_1$ is the set $\mathcal{S}_1$,  defined as
%
\begin{equation}
\mathcal{S}_1 = \{\nabla f_j(x;w^\ast_1) \mid j: y_j=1, (x,y)\in T_1 \}.
\label{eq:Sfirst}
\end{equation}
When training on the next task $T_2$, rather than training using standard SGD, OGD first projects the gradient of the loss for each batch onto the orthogonal complement of the subspace spanned by all elements of the gradient pool $\mathcal{S}_1$. More specifically, the updates of the weights of OGD for given batches of task $T_2$ have the following form,
\begin{equation}
\label{eq:OGD}
    w \gets  w - \eta\hat{g}
\end{equation}
where $\eta$ is the stepsize and $\hat{g} = g - \sum_{u\in \mathcal{S}_1}\mathrm{proj}_u(g)$ with $g$ denoting the gradient $\nabla L_t(w)$ of the loss $L_t(w)$ corresponding to the batch $T_{k_t}$. When training on $T_2$ ends with weights $w_2^*$, OGD augments the gradient pool $\mathcal{S}_1$ with model gradients for $T_2$ and so on. After training on task $T_{k-1}$, the gradient pool is redefined as
\begin{equation}
\!\!\!\!\!
\mathcal{S}_{1:k-1} \!=\! \{\nabla f_j(x;w^\ast_i) \mid j: y_j=1,1\leq i < k , (x,y)\in T_i\}.
\!\!
\end{equation}
Hence, when training on task $T_k$ using the update in \eqref{eq:OGD}, we have that  $\forall _{u \in \mathcal{S}_{1:k-1}}, ~\hat{g}\perp u$.
%
%

The stored model gradients are interpreted as the directions of most change on the previous task. Thus, by projecting orthogonal to these, the resulting projected gradient steps move the weights towards the region of low error on the new task (Task B), while minimally changing the performance on previous tasks (Task A). 



%
%
%

Algorithm \ref{alg:OGDedit} shows pseudo-code for OGD \cite{pmlr-v108-farajtabar20a}. The loss gradient $\nabla L(w)$ is projected on an orthogonal basis $S$ and the resulting component is subtracted from the loss gradient $\nabla L(w)$. The weights $w$ are then updated using this newly obtained gradient and the step size $\eta$. After the training on task $T_k$ is finished, samples $(x,y)$ from the task $T_k$ are used to calculated the model gradient $f_j(x;w)$, which represents the gradient of the $j$th output of the network, where $j$ corresponds to the label $y$. This model gradient is then added to the basis $S$, but it is first made orthogonal to all elements in $S$, to assure that $S$ remains an orthogonal basis.

\begin{algorithm}
\caption{Orthogonal Gradient Descent}\label{alg:OGDedit}
\hspace*{\algorithmicindent} \textbf{Input:} Task sequence $T_1, T_2, T_3, ...$ learning rate $\eta$\\
\hspace*{\algorithmicindent} \textbf{Output:} Model
\begin{algorithmic}[1]
\State \textbf{Initialize} $S \gets \{\}; w \gets w_0$
\For{Task ID $k = 1, 2, 3, ...$}
    \Repeat
        \State $L(w) \gets \frac{1}{|T_{k_t}|}\sum_{(x,y)\in T_{k_t}} L_{(x,y)}(w)$
        \State $g_{proj} \gets \sum_{\mathbf{v} \in S} \frac{\nabla L(w)^\top \mathbf{v}}{\mathbf{v}^\top \mathbf{v}}\mathbf{v}$
        \State $\tilde{g} \gets \nabla L(w) - g_{proj}$
        \State $w \gets w - \eta \tilde{g}$
    \Until{converge}
    \For{$(x,y) \in T_k$ and $j \in [1, c]$ s.t. $y_j = 1$}
        \State $f_{proj} \gets \sum_{\mathbf{v} \in S} \frac {\nabla f_j(x;w)^\top \mathbf{v}}{\mathbf{v}^\top \mathbf{v}}\mathbf{v}$
        \State $u \gets \nabla f_j(x;w) - f_{proj}$
        \State $S \gets S \cup \{u\}$
    \EndFor
\EndFor
\end{algorithmic}
\end{algorithm}

\section{Clustering Mechanism}
Algorithm \ref{alg:CLM} shows the clustering mechanism used by TA-A-GEM and TA-OGD, when adding a sample $\mathbf{z}$ to the pool. First, a pool $C$ is initialized as an empty set. The maximum number of clusters is denoted with $Q$. The maximum cluster size is $P$. The fist $Q$ samples $\mathbf{z}$ initialize a cluster $q$, be being added to it and by setting the cluster's mean $\boldsymbol{\mu}_{q} $ to $\mathbf{z}$. After all clusters are initialized, the $\ell_2$ norm is used to determine the cluster $q^\ast$, that has its mean closest to $\mathbf{z}$. $\mathbf{z}$ is then added to this cluster. The size of $q^\ast$ is kept below the maximum cluster size $P$, by removing its oldest member $\mathbf{z}^{q^\ast}_\mathrm{oldest}$. The mean of the cluster $q^\ast$, namely $\boldsymbol{\mu}_{q^\ast}$, is recalculated by taking the average of all its members.

\begin{algorithm}
\caption{Clustering Mechanism}\label{alg:CLM}
\begin{algorithmic}[1]
\State $C \gets \{ \}$
\State $Q \gets 100$
\State $P \gets 3$
\Procedure{Add}{$\mathbf{z}$}
    \If{$|C| < Q$}
    \State $q \gets \{\mathbf{z}\}$
    \State $C \gets C \cup q $
    \State $\boldsymbol{\mu}_{q} \gets \mathbf{z}$
\Else
    \State $q^\ast \gets \mathrm{argmin}_{q\in\{1,2,\dots,Q\}}\|\mathbf{z} - \boldsymbol{\mu}_q\|^2_2$
    \State $q^\ast \gets q^\ast \cup \mathbf{z}$
    \If{$|q^\ast| > P$}
        \State $q^\ast \gets q^\ast \setminus \mathbf{z}^{q^\ast}_\mathrm{oldest}$
    \EndIf
    \State $\boldsymbol{\mu}_{q^\ast} \gets \frac{1}{|q^\ast|}\sum^{|q^\ast|}_{p=1} {\mathbf{z}^{q^\ast}_p}$
\EndIf
\EndProcedure
\end{algorithmic}
\end{algorithm}

\section{Metric definition}

The \textbf{accuracy} $A_l$ is defined as $A_l = \frac{M_{k_l}}{N_{k_l}}$, where $M_{k_l}$ is the number of correctly classified samples in dataset $T_{k_l}$ and $N_{k_l}$, is the total number of samples corresponding to task $T_{k_l}$. Note that the index $k_l$ is the index of samples from task $k$ at epoch $l$.
The \textbf{forgetting} $F_l$ on the validation accuracy is calculated from the validation accuracy $A_l$, at epoch $l$ using,

\begin{equation}\label{eq:forgetting}
    F_l = \max ( \{ A_i | i \leq l\}) - A_l
\end{equation}


\section{Learning Rate Scheduler}
Algorithm \ref{alg:LRS} shows the pseudo-code of the learning rate scheduler as used by the TA-OGD method. It \textit{lowers} the learning rate, by multiplying it with a \textbf{Factor} $< 1$, when the learning process \textit{stagnates}, i.e. when the loss value reaches a plateau. It \textit{increases} the learning rate, by resetting it to \textbf{LR init}, when a \textit{large spike} in the loss value overshoots the \textbf{Best} loss value by more than the \textbf{Reset Threshold} value. The \textbf{Patience} value, together with the counters \textbf{N\_Stagnant} and \textbf{N\_Spike}, make sure that the learning rate scheduler does not react to irrelevant outlier loss values. The \textbf{Min LR} determines a minimum learning rate, which the learning rate should never fall below. The values mentioned in the pseudo-code are used in the experiments.

\begin{algorithm}
\caption{Learning Rate Scheduler}\label{alg:LRS}
\begin{algorithmic}[1]
\State Factor : $0.9999$
\State Min LR : $10^{-5}$
\State LR init: $10^{-3}$
\State Patience : $5$
\State Best : $\infty$
\State N\_Stagnant : $0$
\State N\_Spike : $0$
\State Threshold: $10^{-4}$
\State Reset Threshold: $1$
\Procedure{Step}{Loss, LR}
\If{Loss $<$ Best $*$ ($1 -$ Threshold)}
    \State  Best $\gets$ Loss
    \State N\_Stagnant $\gets$ $0$
\Else
    \State N\_Stagnant $\gets$ N\_Stagnant $+ 1$
\EndIf
\If{Loss $>$ Best + Reset Threshold}
    \State N\_Spike $\gets$ N\_Spike $+ 1$
\Else
    \State N\_Spike $\gets$ $0$
\EndIf
\If {N\_Spike $>$ Patience}
    \State LR $\gets$ LR init
    \State Best $\gets$ Loss
    \State N\_Spike $\gets$ $0$
\EndIf
\If {N\_Stagnant $>$ Patience}
    \State LR $\gets$ max(LR $*$ Factor, Min LR)
    \State N\_Stagnant $\gets$ $0$
\EndIf
\State \Return LR
\EndProcedure
\end{algorithmic}
\end{algorithm}


\section{Experiments}
Two main classes of experiments are described in the main paper, the \textit{``Disjoint tasks experiments''} and the \textit{``Continuous change experiments''}. In this section, figures and results are given, for which no space existed in the main paper.

\subsection{Best practices}
In order to test the effectiveness of a continual learning method, different tasks need to be presented to it over time. These tasks can be synthesized from standard datasets, but it is crucial that the produced tasks give a good representation of a real world continual learning setting. In designing our experiment, we abide by all the desiderata described  in \cite{Farquhar2018}, i.e., 1) cross-task resemblances between tasks, 2) a shared output head, 3) no test-time assumed task labels, 4) no unconstrained retraining on old tasks, and 5) the use of more than just two tasks. By conforming to these, we ensure that the continual learning task is not trivially easy and thus a method does not seem to be more effective than it would truly be in real-case scenarios. Note that for the case of the experiments on permutation for task generation, we ``violate" the cross-task resemblances since such a requirement  is impossible to be enforced. 

\subsection{Task generation details}
Separate tasks were created from existing datasets by means of \textit{permutation}, \textit{rotation} and \textit{class splitting}.
The task permutation mechanism picks one permutation that shuffles all pixels at random per task and then applies this one permutation to all images in the dataset. The task rotation mechanism rotates every image in one task with a given number of degrees. We chose to increment the number of degrees by 20 for each new task. The class splitting mechanism splits the dataset on disjoint sets of labels. As subsets we chose the labels (0,1), (2,3), (4,5), (6,7) and (8,9).

To make the tasks more disjoint, for the task permutation and task rotation, the data was first split into disjoint subsets (one for each task). After this, the permutation or rotation was applied. For class splitting, the different tasks are already completely disjoint, so no extra separation step was required. Note that we do not use head swapping, meaning that we have one fixed output layer for all tasks. For the class split scenario, only two classes are given to the network during one task. Therefore we chose to give the network just two output nodes, instead of ten. To accommodate this, we changed the labels. All even labels were set to $0$ and all odd labels were set to $1$.

\subsection{Environment}
Each experiment instance ran on a single NVIDIA Tesla T4 card with 15 Gb of video memory. Up to fifteen NVIDIA Tesla T4 cards were available in parallel, together with 32 physical processor cores and 252 GB of RAM.

\subsection{Disjoint tasks experiments}
\begin{figure*}
\centering
\begin{subfigure}{1.0\textwidth}
    \includegraphics[width=.3\textwidth]{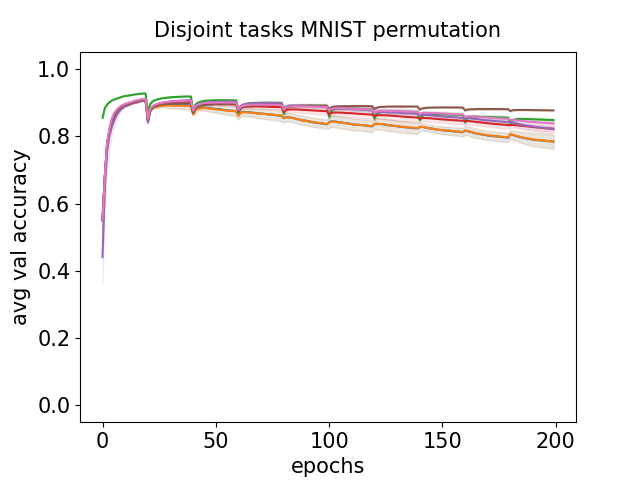}\hfill
    \includegraphics[width=.3\textwidth]{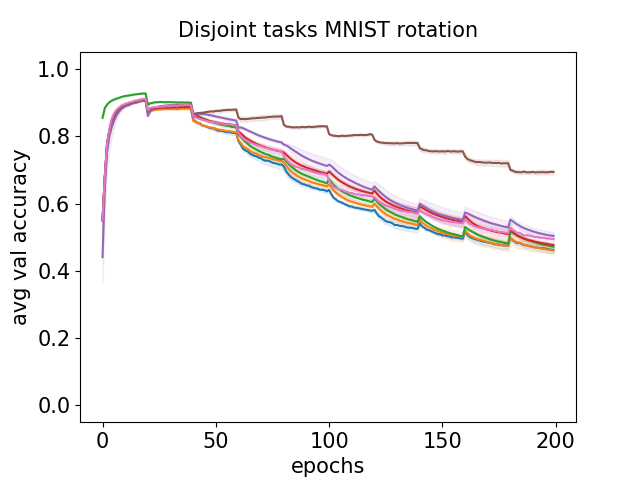}\hfill
    \includegraphics[width=.3\textwidth]{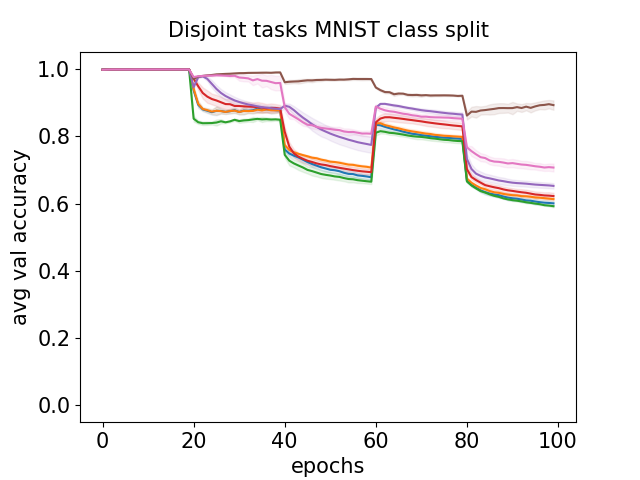}
    \caption{Average validation accuracy}
    \label{fig:disjoint_MNIST_s1}
\end{subfigure}\\
\begin{subfigure}{1.0\textwidth}
    \includegraphics[width=.3\textwidth]{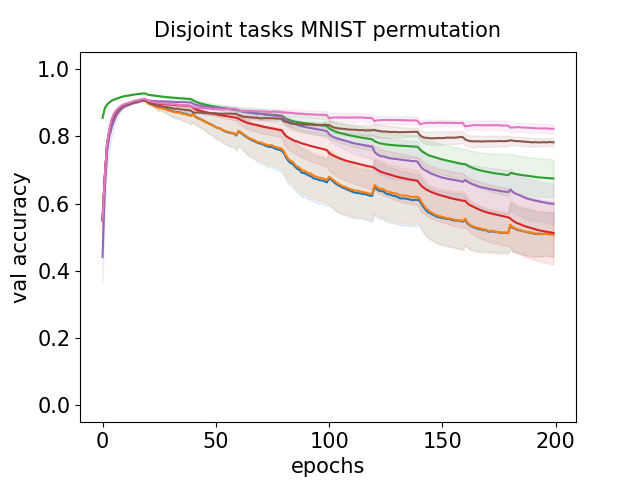}\hfill
    \includegraphics[width=.3\textwidth]{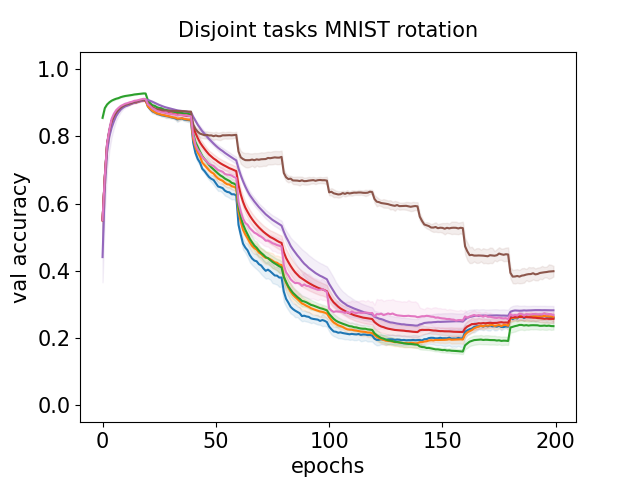}\hfill
    \includegraphics[width=.3\textwidth]{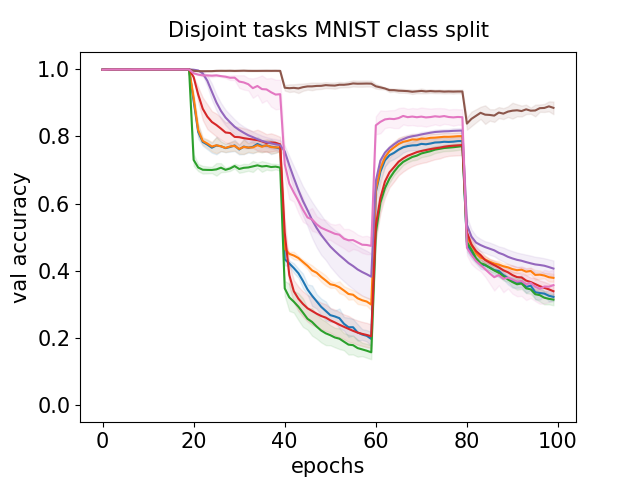}
    \caption{Validation accuracy of the first task}
    \label{fig:disjoint_MNIST_s2}
\end{subfigure}\\
\begin{subfigure}{1.0\textwidth}
    \includegraphics[width=.3\textwidth]{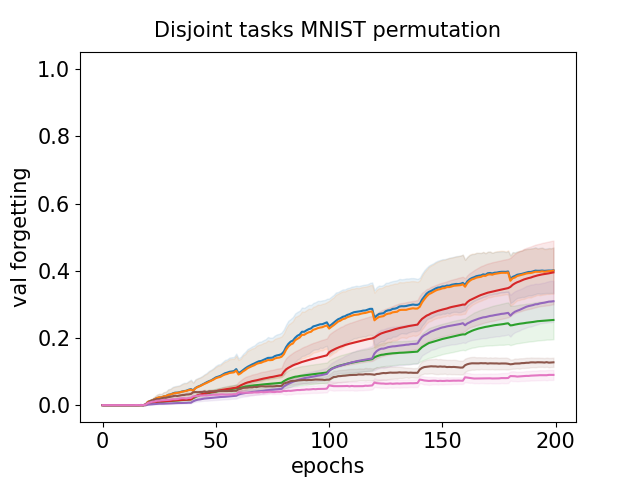}\hfill
    \includegraphics[width=.3\textwidth]{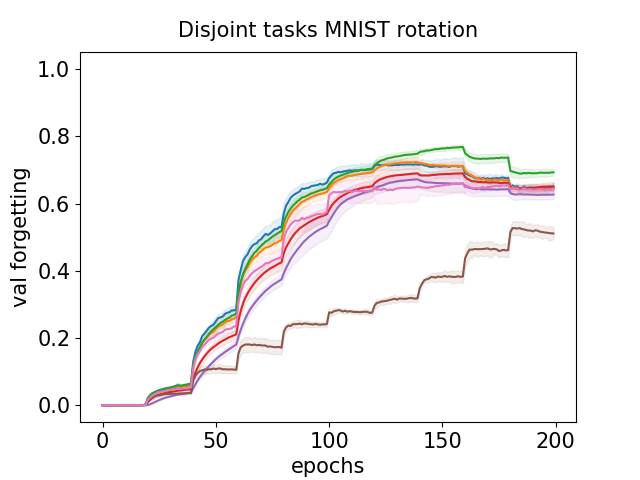}\hfill
    \includegraphics[width=.3\textwidth]{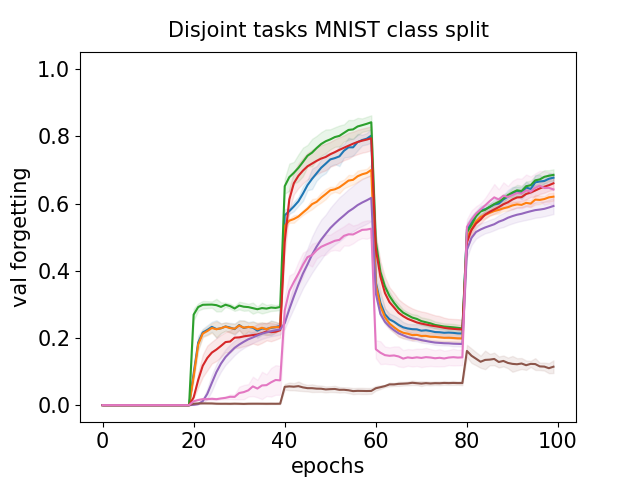}
    \caption{Forgetting on validation accuracy of the first task}
    \label{fig:disjoint_MNIST_s3}
\end{subfigure}\\
\begin{subfigure}{1.0\textwidth}
    \centering
    \includegraphics[width=1.0\textwidth]{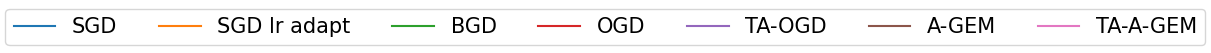}
    \label{fig:disjoint_MNIST_legend}
\end{subfigure}
\caption{Results of the ``\textbf{Disjoint tasks experiments}'' on the \textbf{MNIST} dataset. (a) Average validation accuracy, (b) Validation accuracy of the first task, (c) Forgetting on validation accuracy of the first task, averaged over all tasks trained thus far, then averaged over five runs, depicted by a solid line plot with $\pm$ one standard deviation as a shaded area. From left to right: task separation by permutation, task separation by rotation and task separation by class split.}
\label{fig:disjoint_MNIST}
\end{figure*}

\begin{figure*}
\centering
\begin{subfigure}{1.0\textwidth}
    \includegraphics[width=.3\textwidth]{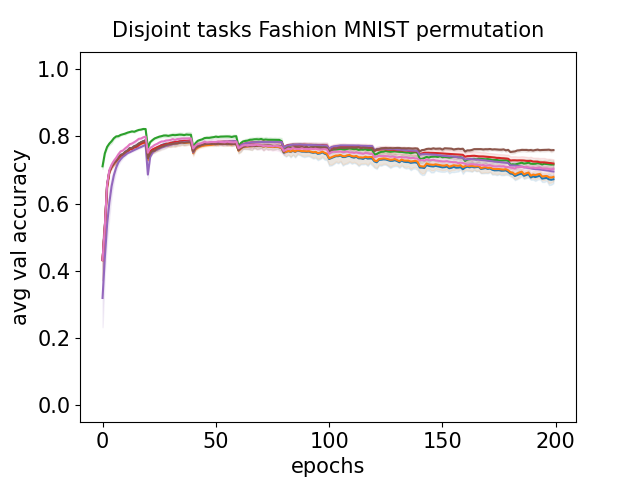}\hfill
    \includegraphics[width=.3\textwidth]{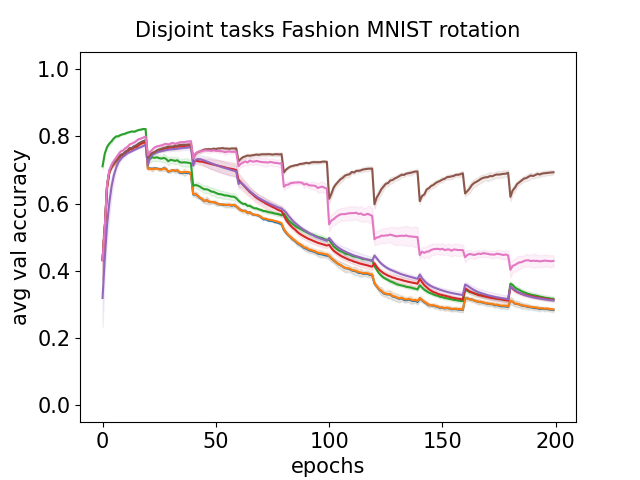}\hfill
    \includegraphics[width=.3\textwidth]{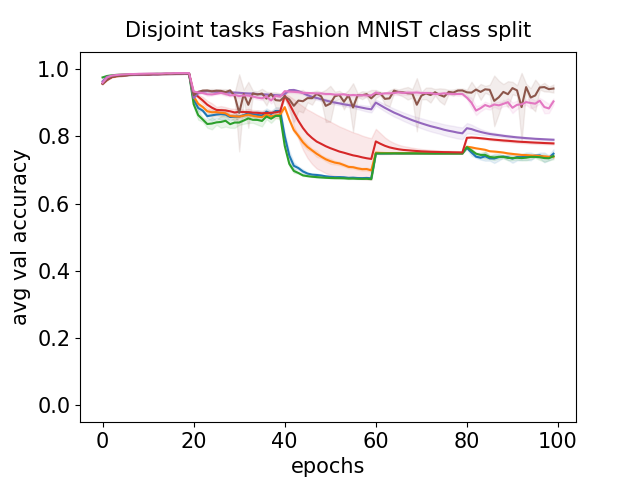}
    \caption{Average validation accuracy}
    \label{fig:disjoint_Fashion_MNIST_s1}
\end{subfigure}\\
\begin{subfigure}{1.0\textwidth}
    \includegraphics[width=.3\textwidth]{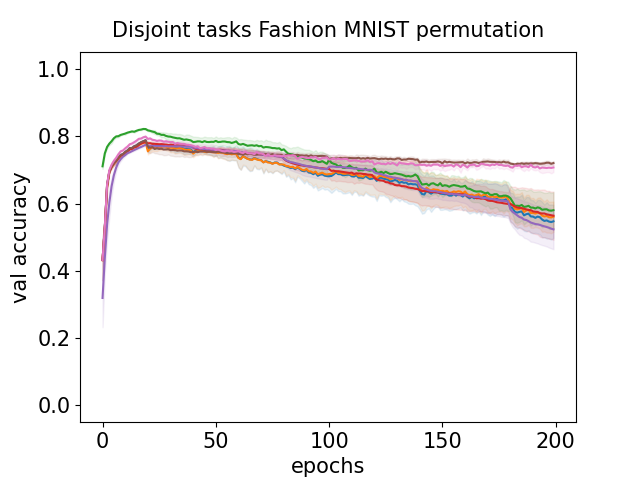}\hfill
    \includegraphics[width=.3\textwidth]{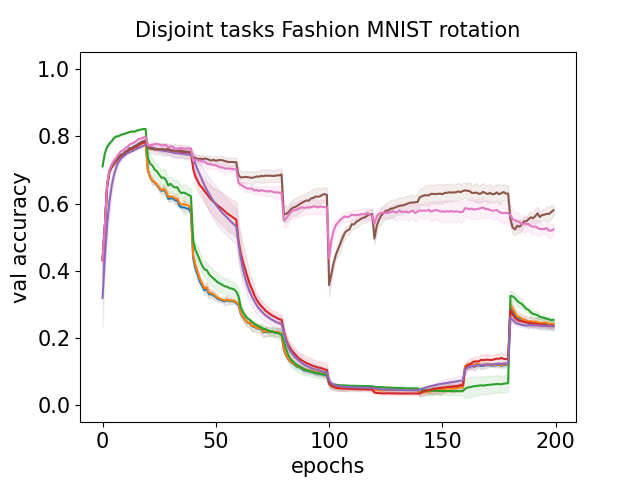}\hfill
    \includegraphics[width=.3\textwidth]{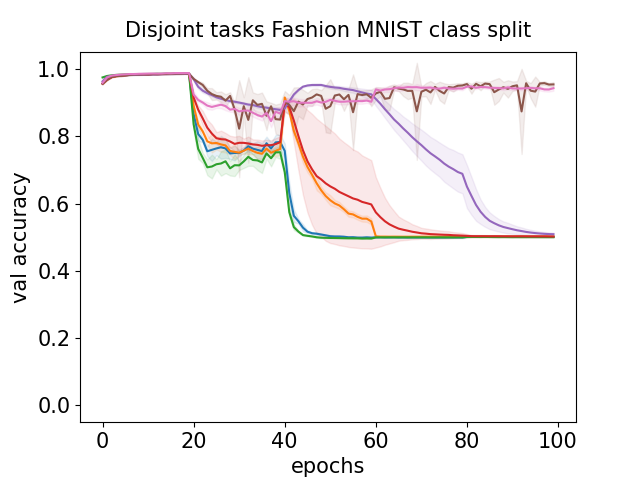}
    \caption{Validation accuracy of the first task}
    \label{fig:disjoint_Fashion_MNIST_s2}
\end{subfigure}\\
\begin{subfigure}{1.0\textwidth}
    \includegraphics[width=.3\textwidth]{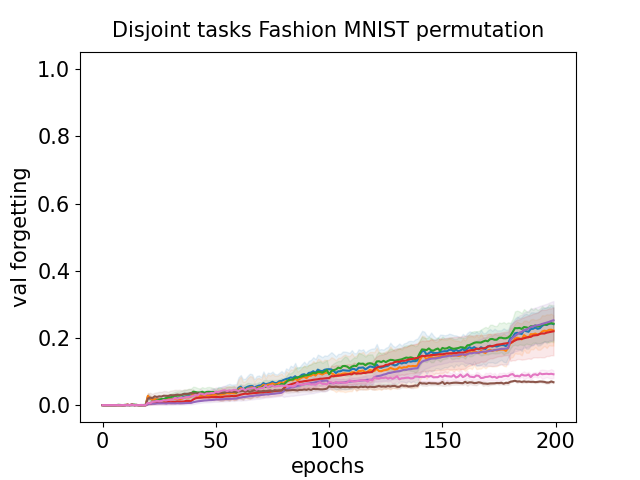}\hfill
    \includegraphics[width=.3\textwidth]{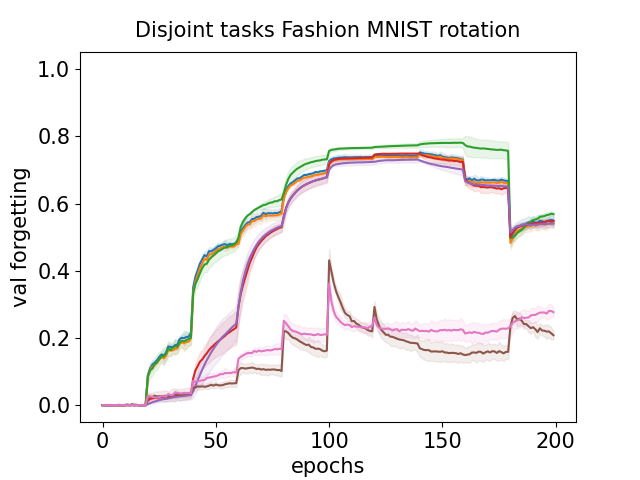}\hfill
    \includegraphics[width=.3\textwidth]{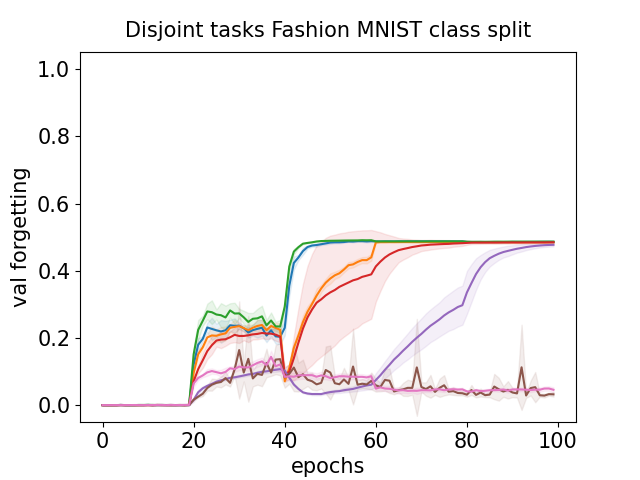}
    \caption{Forgetting on validation accuracy of the first task}
    \label{fig:disjoint_Fashion_MNIST_s3}
\end{subfigure}\\
\begin{subfigure}{1.0\textwidth}
    \centering
    \includegraphics[width=1.0\textwidth]{figures/disjoint_tasks_experiments/disjoint_legend.png}
    \label{fig:disjoint_Fashion_MNIST_legend}
\end{subfigure}
\caption{Results of the ``\textbf{Disjoint tasks experiments}'' on the \textbf{Fashion MNIST} dataset. (a) Average validation accuracy, (b) Validation accuracy of the first task, (c) Forgetting on validation accuracy of the first task, averaged over all tasks trained thus far, then averaged over five runs, depicted by a solid line plot with $\pm$ one standard deviation as a shaded area. From left to right: task separation by permutation, task separation by rotation and task separation by class split.}
\label{fig:disjoint_Fashion_MNIST}
\end{figure*}

\begin{figure*}
\centering
\begin{subfigure}{1.0\textwidth}
    \includegraphics[width=.3\textwidth]{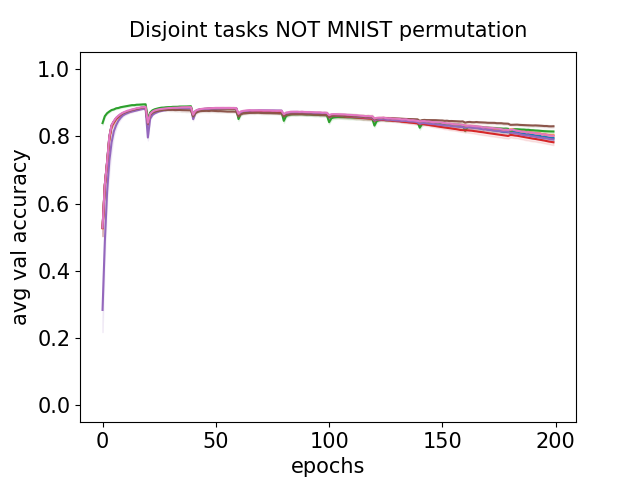}\hfill
    \includegraphics[width=.3\textwidth]{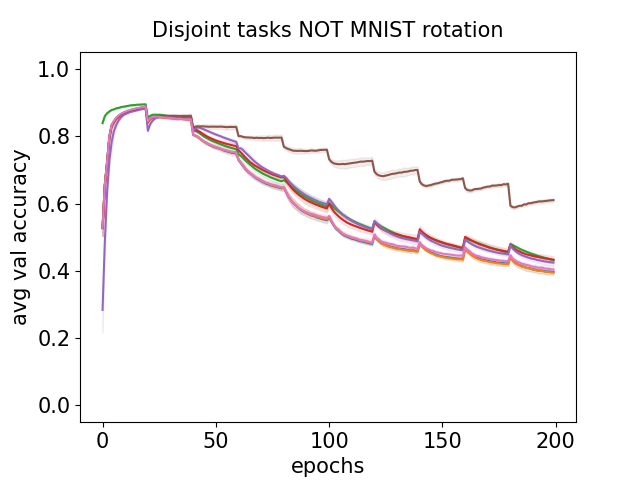}\hfill
    \includegraphics[width=.3\textwidth]{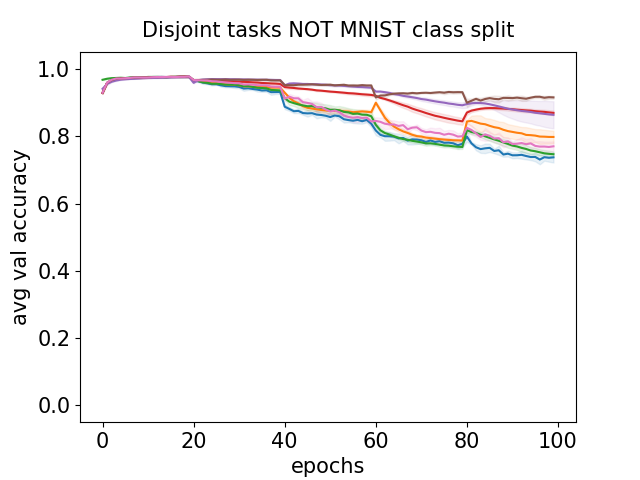}
    \caption{Average validation accuracy}
    \label{fig:disjoint_NOT_MNIST_s1}
\end{subfigure}\\
\begin{subfigure}{1.0\textwidth}
    \includegraphics[width=.3\textwidth]{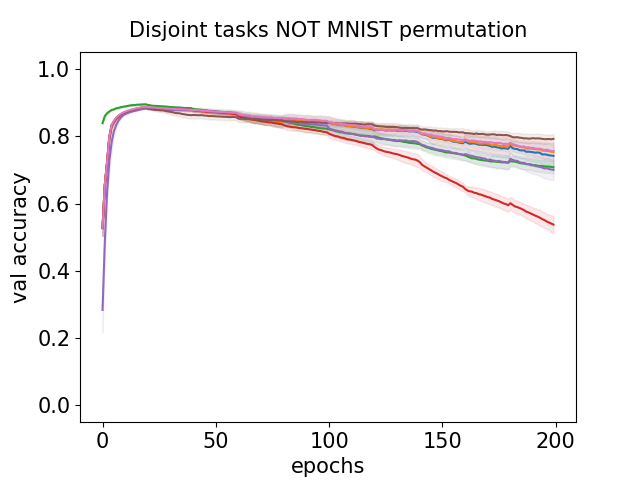}\hfill
    \includegraphics[width=.3\textwidth]{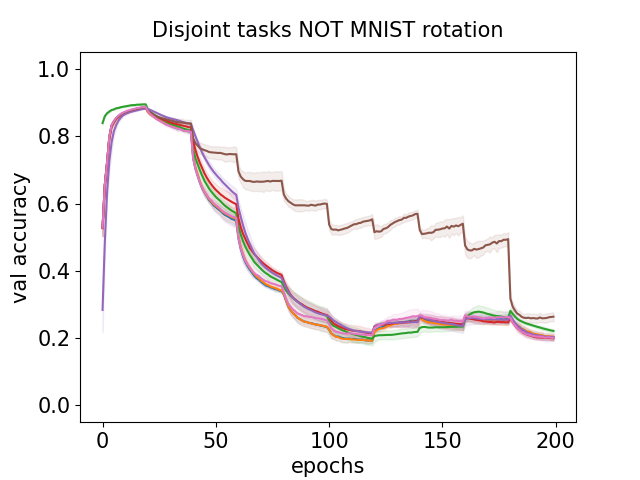}\hfill
    \includegraphics[width=.3\textwidth]{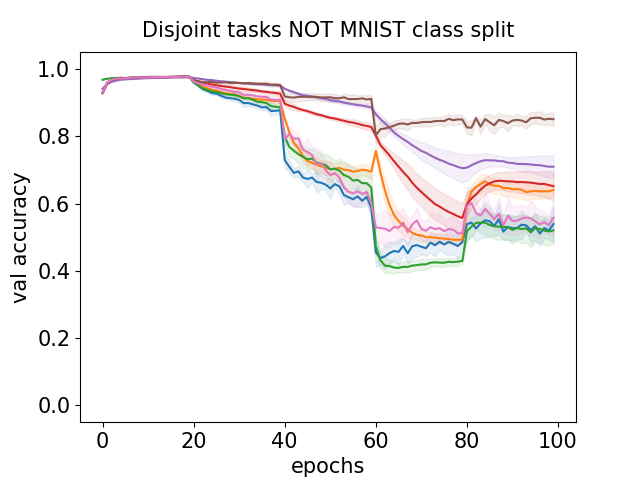}
    \caption{Validation accuracy of the first task}
    \label{fig:disjoint_NOT_MNIST_s2}
\end{subfigure}\\
\begin{subfigure}{1.0\textwidth}
    \includegraphics[width=.3\textwidth]{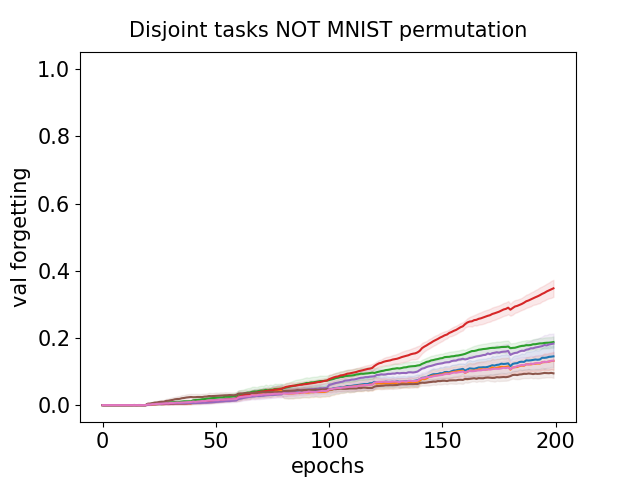}\hfill
    \includegraphics[width=.3\textwidth]{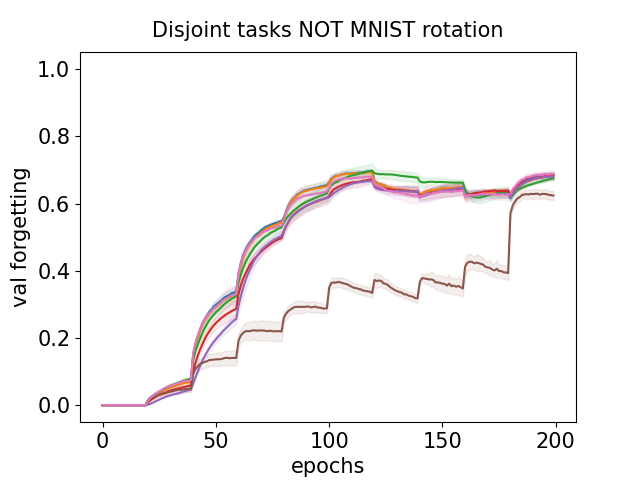}\hfill
    \includegraphics[width=.3\textwidth]{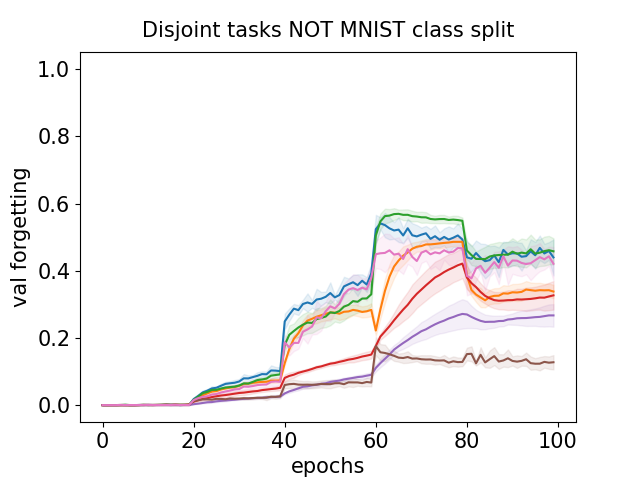}
    \caption{Forgetting on validation accuracy of the first task}
    \label{fig:disjoint_NOT_MNIST_s3}
\end{subfigure}\\
\begin{subfigure}{1.0\textwidth}
    \centering
    \includegraphics[width=1.0\textwidth]{figures/disjoint_tasks_experiments/disjoint_legend.png}
    \label{fig:disjoint_NOT_MNIST_legend}
\end{subfigure}
\caption{Results of the ``\textbf{Disjoint tasks experiments}'' on the \textbf{NOT MNIST} dataset. (a) Average validation accuracy, (b) Validation accuracy of the first task, (c) Forgetting on validation accuracy of the first task, averaged over all tasks trained thus far, then averaged over five runs, depicted by a solid line plot with $\pm$ one standard deviation as a shaded area. From left to right: task separation by permutation, task separation by rotation and task separation by class split.}
\label{fig:disjoint_NOT_MNIST}
\end{figure*}

\begin{figure*}
\centering
\begin{subfigure}{1.0\textwidth}
    \includegraphics[width=.3\textwidth]{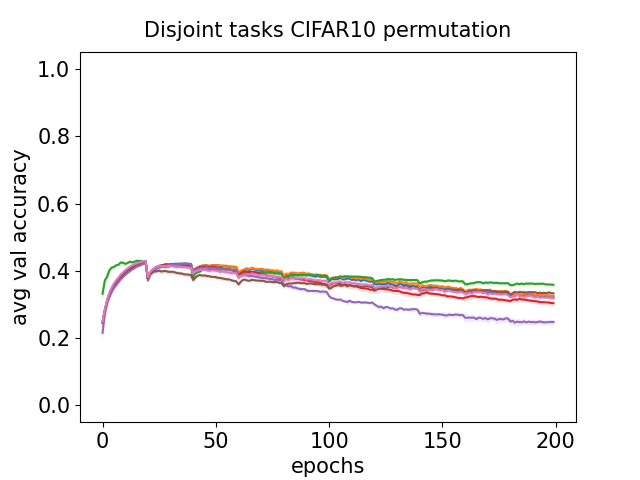}\hfill
    \includegraphics[width=.3\textwidth]{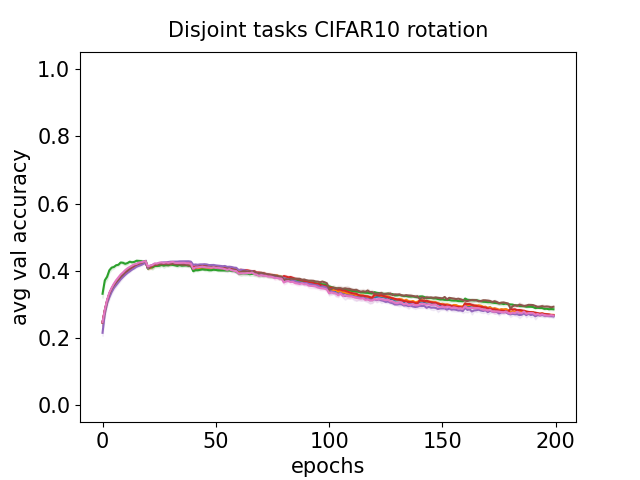}\hfill
    \includegraphics[width=.3\textwidth]{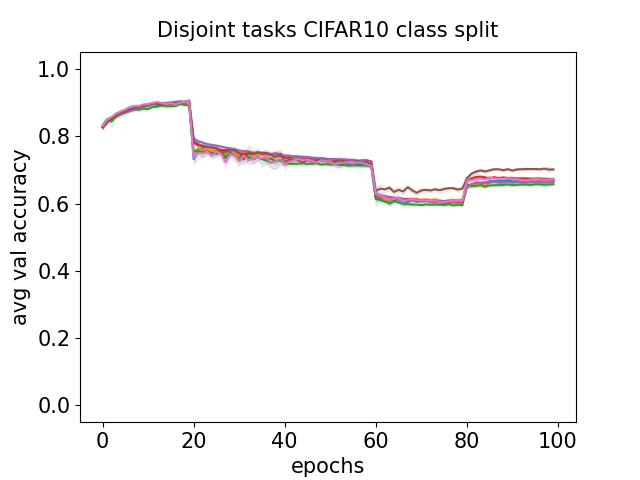}
    \caption{Average validation accuracy}
    \label{fig:disjoint_CIFAR10_s1}
\end{subfigure}\\
\begin{subfigure}{1.0\textwidth}
    \includegraphics[width=.3\textwidth]{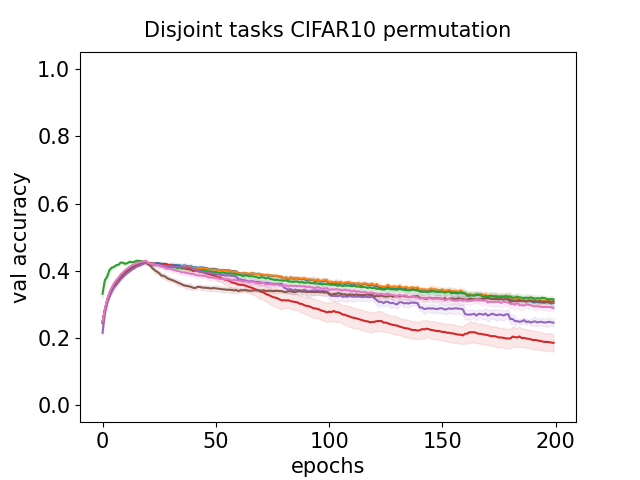}\hfill
    \includegraphics[width=.3\textwidth]{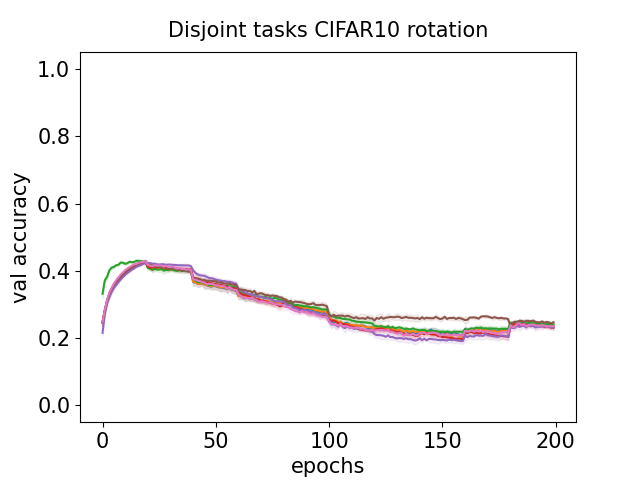}\hfill
    \includegraphics[width=.3\textwidth]{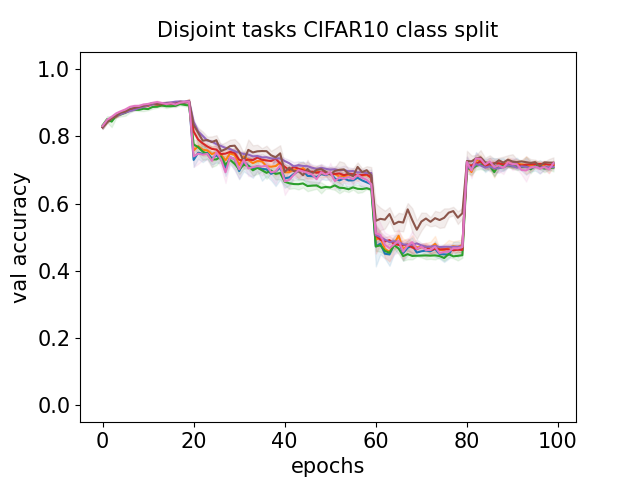}
    \caption{Validation accuracy of the first task}
    \label{fig:disjoint_CIFAR10_s2}
\end{subfigure}\\
\begin{subfigure}{1.0\textwidth}
    \includegraphics[width=.3\textwidth]{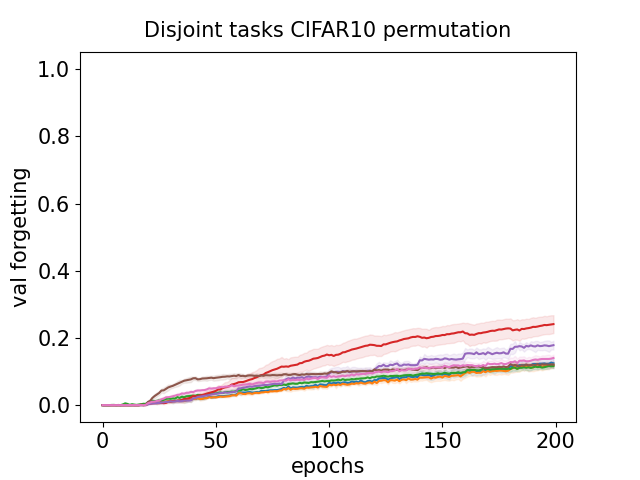}\hfill
    \includegraphics[width=.3\textwidth]{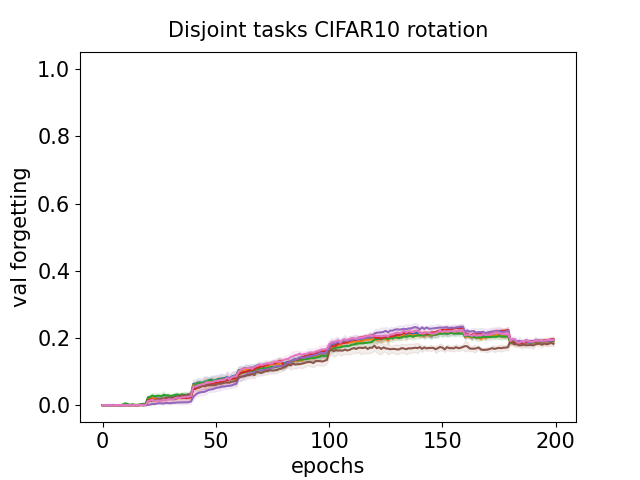}\hfill
    \includegraphics[width=.3\textwidth]{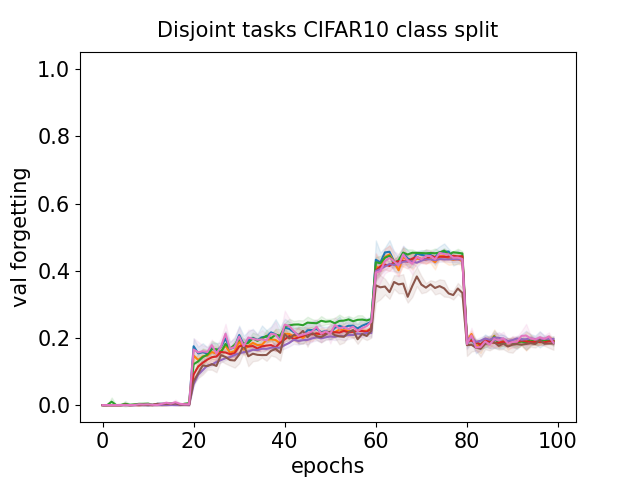}
    \caption{Forgetting on validation accuracy of the first task}
    \label{fig:disjoint_CIFAR10_s3}
\end{subfigure}\\
\begin{subfigure}{1.0\textwidth}
    \centering
    \includegraphics[width=1.0\textwidth]{figures/disjoint_tasks_experiments/disjoint_legend.png}
    \label{fig:disjoint_CIFAR10_legend}
\end{subfigure}
\caption{Results of the ``\textbf{Disjoint tasks experiments}'' on the \textbf{CIFAR10} dataset. (a) Average validation accuracy, (b) Validation accuracy of the first task, (c) Forgetting on validation accuracy of the first task, averaged over all tasks trained thus far, then averaged over five runs, depicted by a solid line plot with $\pm$ one standard deviation as a shaded area. From left to right: task separation by permutation, task separation by rotation and task separation by class split.}
\label{fig:disjoint_CIFAR10}
\end{figure*}

\begin{figure*}
\centering
\begin{subfigure}{1.0\textwidth}
    \includegraphics[width=.3\textwidth]{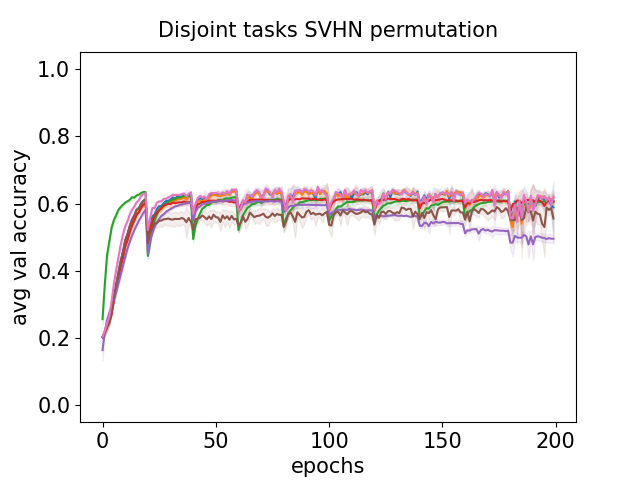}\hfill
    \includegraphics[width=.3\textwidth]{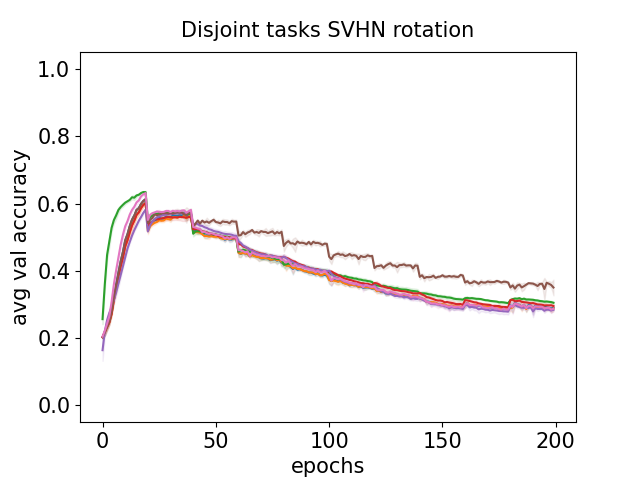}\hfill
    \includegraphics[width=.3\textwidth]{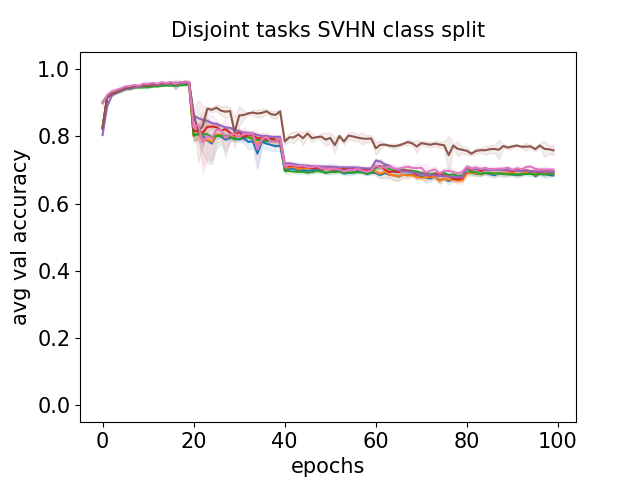}
    \caption{Average validation accuracy}
    \label{fig:disjoint_SVHN_s1}
\end{subfigure}\\
\begin{subfigure}{1.0\textwidth}
    \includegraphics[width=.3\textwidth]{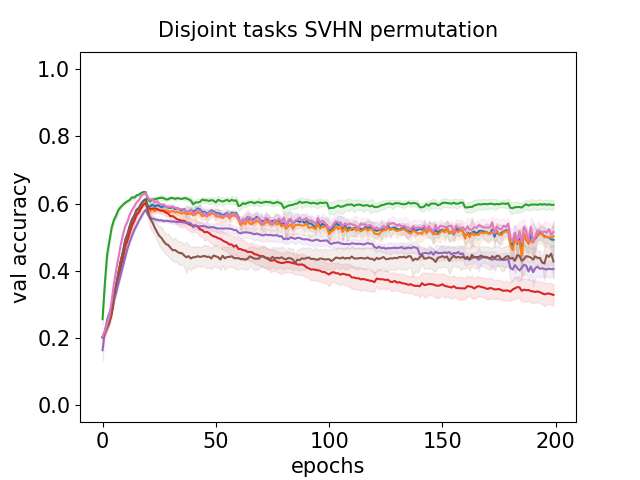}\hfill
    \includegraphics[width=.3\textwidth]{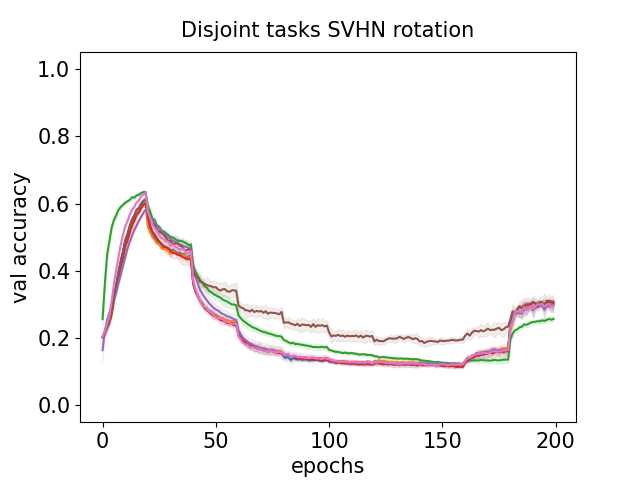}\hfill
    \includegraphics[width=.3\textwidth]{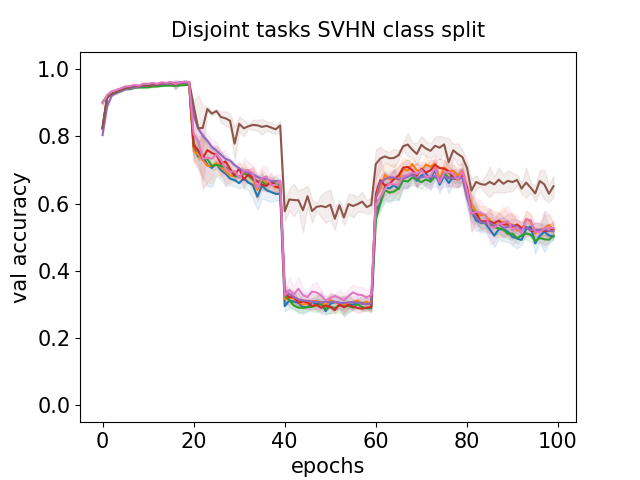}
    \caption{Validation accuracy of the first task}
    \label{fig:disjoint_SVHN_s2}
\end{subfigure}\\
\begin{subfigure}{1.0\textwidth}
    \includegraphics[width=.3\textwidth]{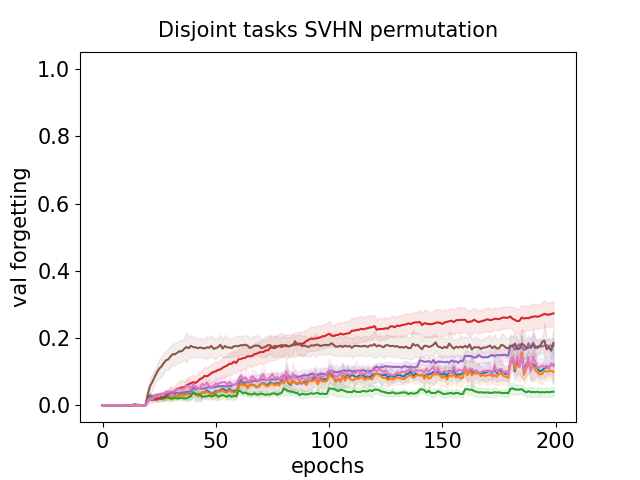}\hfill
    \includegraphics[width=.3\textwidth]{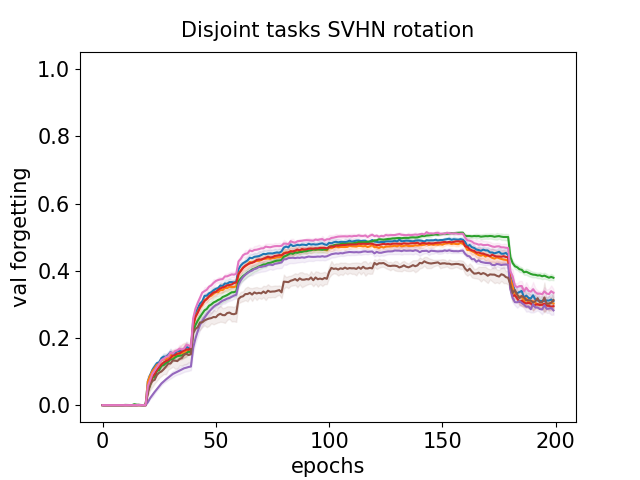}\hfill
    \includegraphics[width=.3\textwidth]{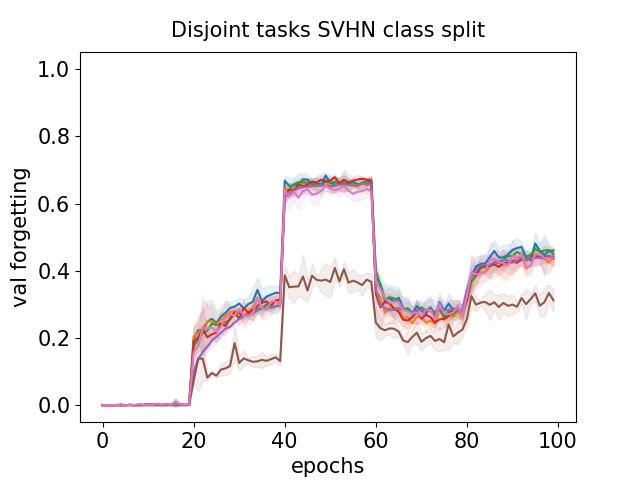}
    \caption{Forgetting on validation accuracy of the first task}
    \label{fig:disjoint_SVHN_s3}
\end{subfigure}\\
\begin{subfigure}{1.0\textwidth}
    \centering
    \includegraphics[width=1.0\textwidth]{figures/disjoint_tasks_experiments/disjoint_legend.png}
    \label{fig:disjoint_SVHN_legend}
\end{subfigure}
\caption{Results of the ``\textbf{Disjoint tasks experiments}'' on the \textbf{SVHN} dataset. (a) Average validation accuracy, (b) Validation accuracy of the first task, (c) Forgetting on validation accuracy of the first task, averaged over all tasks trained thus far, then averaged over five runs, depicted by a solid line plot with $\pm$ one standard deviation as a shaded area. From left to right: task separation by permutation, task separation by rotation and task separation by class split.}
\label{fig:disjoint_SVHN}
\end{figure*}

Figures \ref{fig:disjoint_MNIST}, \ref{fig:disjoint_Fashion_MNIST}, \ref{fig:disjoint_NOT_MNIST}, \ref{fig:disjoint_CIFAR10} and  \ref{fig:disjoint_SVHN} show the average validation accuracy, the validation accuracy on the first task and the forgetting on the validation accuracy during training for different task split methods, using the MNIST, Fashion MNIST, NOT MNIST, CIFAR10 and SVHN datasets. Table \ref{tab:disjoint_acc} and \ref{tab:disjoint_forg} summarize the results on the validation accuracy on the first task and the forgetting on the validation accuracy on the first task, respectively. A summary for the average validation accuracy is included in the main paper.

Specific settings for the task splitting can be seen in table \ref{tab:tasksplitsettings}. The settings specific to TA-A-GEM and TA-OGD can be seen in  table \ref{tab:taagemsettings} and \ref{tab:taogdsettings}, respectively. All results depicted are the average of five independent runs.

\begin{table*}
\centering
\scalebox{0.8}{
\begin{tabular}{c|ccc|ccc|ccc|ccc|ccc}
    & \multicolumn{3}{c|}{MNIST} & \multicolumn{3}{c|}{Fashion MNIST} & \multicolumn{3}{c|}{NOT MNIST} & \multicolumn{3}{c|}{CIFAR10} & \multicolumn{3}{c}{SVHN}\\
    & perm & rot & class & perm & rot & class & perm & rot & class& perm & rot & class & perm & rot & class \\
    \hline
    SGD & 0.691 & 0.419 & 0.643 & 0.682 & 0.254 & 0.657 & \textbf{0.825} & 0.408 & 0.707 & \textbf{0.364} & 0.288 & 0.692 & 0.527 & 0.231 & 0.622\\
    SGD lr adapt & 0.693 & 0.429 & 0.669 & 0.686 & 0.255 & 0.684 & \textbf{0.827} & 0.408 & 0.761 & \textbf{0.365} & 0.288 & \textbf{0.699} & 0.520 & 0.231 & \textbf{0.635}\\
    BGD & 0.812 & 0.431 & 0.606 & \textbf{0.716} & 0.270 & 0.646 & \textbf{0.813} & \textbf{0.425} & 0.711 & \textbf{0.365} & \textbf{0.297} & 0.682 & \textbf{0.596} & \textbf{0.261} & 0.623\\
    TA-OGD & 0.779 & \textbf{0.487} & \textbf{0.721} & 0.686 & 0.302 & 0.831 & 0.804 & \textbf{0.429} & \textbf{0.866} & 0.332 & 0.283 & \textbf{0.708} & 0.473 & 0.235 & \textbf{0.637}\\
    TA-A-GEM & \textbf{0.858} & \textbf{0.465} & \textbf{0.748} & \textbf{0.734} & \textbf{0.625} & \textbf{0.931} & \textbf{0.830} & 0.415 & 0.738 & 0.347 & 0.285 & 0.695 & 0.536 & 0.236 & \textbf{0.640}\\
    \hline
    \color{gray}OGD & \color{gray}0.737 & \color{gray}0.461 & \color{gray}0.645 & \color{gray}0.688 & \color{gray}0.306 & \color{gray}0.699 & \color{gray}0.766 & \color{gray}0.426 & \color{gray}0.817 & \color{gray}\color{gray}0.293 & \color{gray}0.285 & \color{gray}0.703 & \color{gray}0.414 & \color{gray}0.229 & \color{gray}0.634\\
    \color{gray}A-GEM & \color{gray}0.830 & \color{gray}0.654 & \color{gray}0.95 & \color{gray}0.735 & \color{gray}0.640 & \color{gray}0.934 & \color{gray}0.834 & \color{gray}0.605 & \color{gray}0.906 & \color{gray}0.338 & \color{gray}0.308 & \color{gray}0.725 & \color{gray}0.443 & \color{gray}0.292 & \color{gray}0.757\\
\end{tabular}
}
\caption{\textbf{Validation accuracy} of the first task, averaged over all epochs, then averaged over five runs, for the \textbf{disjoint tasks experiments} when using a MLP. Per column, the best result for the \textit{task-agnostic} methods are written in bold. In case a task-agnostic method's result is less optimal and not significantly different from the best result, with a confidence of 99\%, it is also written in bold. The results for the \textit{task-aware} methods OGD and A-GEM are given for context. Since these algorithms benefit from knowing task identities and changes, we just use them here as baselines for indicating the best performance we can achieve.}
\label{tab:disjoint_acc}
\end{table*}

\begin{table*}
\centering
\scalebox{0.8}{
\begin{tabular}{c|ccc|ccc|ccc|ccc|ccc}
    & \multicolumn{3}{c|}{MNIST} & \multicolumn{3}{c|}{Fashion MNIST} & \multicolumn{3}{c|}{NOT MNIST} & \multicolumn{3}{c|}{CIFAR10} & \multicolumn{3}{c}{SVHN}\\
    & perm & rot & class & perm & rot & class & perm & rot & class& perm & rot & class & perm & rot & class \\
    \hline
    SGD & 0.213 & 0.485 & 0.356 & 0.099 & 0.527 & 0.329 & \textbf{0.056} & 0.473 & 0.270 & \textbf{0.060} & \textbf{0.136} & 0.209 & 0.068 & 0.364 & 0.337\\
    SGD lr adapt & 0.209 & 0.473 & 0.330 & \textbf{0.091} & 0.521 & 0.302 & \textbf{0.053} & 0.471 & 0.215 & \textbf{0.056} & \textbf{0.134} & 0.201 & 0.064 & 0.353 & \textbf{0.322}\\
    BGD & 0.115 & 0.496 & 0.393 & 0.104 & 0.550 & 0.340 & 0.081 & 0.469 & 0.266 & \textbf{0.065} & \textbf{0.133} & 0.212 & \textbf{0.032} & 0.367 & \textbf{0.328}\\
    TA-OGD & 0.123 & \textbf{0.415} & \textbf{0.278} & \textbf{0.081} & 0.464 & 0.154 & \textbf{0.070} & \textbf{0.444} & \textbf{0.109} & 0.086 & \textbf{0.135} & \textbf{0.193} & 0.091 & \textbf{0.330} & \textbf{0.319}\\
    TA-A-GEM & \textbf{0.049} & 0.442 & \textbf{0.251} & \textbf{0.058} & \textbf{0.167} & \textbf{0.057} & \textbf{0.053} & 0.468 & 0.239 & 0.078 & \textbf{0.139} & 0.207 & 0.078 & 0.378 & \textbf{0.322}\\
    \hline
    \color{gray}OGD & \color{gray}0.165 & \color{gray}0.441 & \color{gray}0.354 & \color{gray}0.089 & \color{gray}0.470 & \color{gray}0.286 & \color{gray}0.113 & \color{gray}0.453 & \color{gray}0.160 & \color{gray}0.128 & \color{gray}0.137 & \color{gray}0.198 & \color{gray}0.170 & \color{gray}0.354 & \color{gray}0.323\\
    \color{gray}A-GEM & \color{gray}0.074 & \color{gray}0.250 & \color{gray}0.049 & \color{gray}0.046 & \color{gray}0.142 & \color{gray}0.052 & \color{gray}0.048 & \color{gray}0.276 & \color{gray}0.072 & \color{gray}0.086 & \color{gray}0.116 & \color{gray}0.176 & \color{gray}0.153 & \color{gray}0.303 & \color{gray}0.202\\
\end{tabular}
}
\caption{\textbf{Forgetting} on validation accuracy of the first task, averaged over five runs, then averaged over all epochs, for the \textbf{disjoint tasks experiments} when using a MLP. Per column, the best result for the \textit{task-agnostic} methods are written in bold. In case a task-agnostic method's result is less optimal and not significantly different from the best result, with a confidence of 99\%, it is also written in bold. The results for the \textit{task-aware} methods OGD and A-GEM are given for context. Since these algorithms benefit from knowing task identities and changes, we just use them here as baselines for indicating the best performance we can achieve.}
\label{tab:disjoint_forg}
\end{table*}
\subsection{Continuous change experiments}
In the \textit{Continuous change experiments}, tasks slowly change into the next task halfway through the training process. For the first 10 epochs (0-9), the network is trained on purely the current task. For epoch 10 to 19, for every batch, slowly more and more samples of the next task are mixed in. So the first batch of epoch 10 will contain 100\% samples from the current tasks and 0\% of the next task, while the last batch of epoch 19 will contain 0\% samples of the current task and 100\% samples of the next task. These percentages gradually change by linear interpolation.
Table \ref{tab:continuous_change_avg_acc} shows the result of the \textit{continuous change experiments}.
In these experiments, tasks gradually change into the next task. This is accomplished by first loading two datasets $A$ and $B$, where dataset $A$ represent the current task and dataset $B$ represents the next task. For each epoch, $A$ is randomly shuffled and $B$ is randomly shuffled. A fraction $q \in (0,1)$ determines how many samples of $A$ and how many samples of $B$ should be mixed into one batch. A batch of size $z$ is taken from both $A$ and $B$. From the batch of $A$, the first  $\lfloor q*z \rfloor$ samples are selected. From the batch of $B$, we select the last $\lceil (1-q)*z \rceil$ samples. Since datasets $A$ and $B$ are shuffled every epoch, this is equivalent to random sampling without replacement. The selected samples are joined in a mixed batch. This mixed batch is then shuffled again, so no distinction between the tasks can be made. This mixed batch is then used for training.

For this experiment, training commences for each pair of datasets for 20 epochs. For the first 10 epochs (0-9), $q$ is set to $1.0$. This means that for these epochs, only the first of the datasets is presented to the learner. For the next 10 tasks (10-19), $q$ gradually changes for each batch from $1.0$ to $0.0$. This means that for the first batch of epoch 10, $q$ equals $1.0$ and for the last batch of epoch 19, $q$ equals $0.0$. Between these batches, $q$ changes linearly. 

\begin{table*}
\centering
\scalebox{0.75}{
\begin{tabular}{c|ccc|ccc|ccc|ccc|ccc}
    & \multicolumn{3}{c|}{MNIST} & \multicolumn{3}{c|}{Fashion MNIST} & \multicolumn{3}{c|}{NOT MNIST} & \multicolumn{3}{c|}{CIFAR10} & \multicolumn{3}{c}{SVHN}\\
    & perm & rot & class & perm & rot & class & perm & rot & class & perm & rot & class & perm & rot & class \\
    \hline
    SGD & 0.841 & 0.647 & 0.816 & 0.738 & 0.471 & 0.814 & 0.849 & 0.588 & 0.866 & 0.378 & 0.348 & 0.728 & 0.597 & 0.396 & 0.755\\
    SGD lr adapt & 0.859 & 0.654 & 0.819 & 0.748 & 0.487 & 0.821 & \textbf{0.854} & 0.592 & 0.893 & 0.380 & 0.349 & 0.731 & 0.592 & 0.394 & 0.772\\
    BGD & \textbf{0.884} & 0.678 & 0.808 & \textbf{0.765} & 0.515 & 0.812 & \textbf{0.860} & \textbf{0.621} & 0.865 & \textbf{0.390} & \textbf{0.358} & 0.719 & \textbf{0.605} & \textbf{0.419} & 0.760\\
    TA-OGD & 0.875 & \textbf{0.706} & \textbf{0.912} & 0.748 & 0.522 & 0.902 & 0.846& \textbf{0.622} & \textbf{0.937} & 0.335 & 0.343 & \textbf{0.735} & 0.567 & 0.392 & \textbf{0.792}\\
    TA-A-GEM & 0.876 & 0.670 & 0.87 & 0.745 & \textbf{0.594} & \textbf{0.929} & \textbf{0.850} & 0.592 & 0.876 & 0.369 & 0.346 & 0.715 & \textbf{0.606} & 0.403 & 0.765\\
\end{tabular}
}
\caption{\textbf{Average validation accuracy}, averaged over all tasks trained thus far, then averaged over all epochs, then averaged over five runs, for the \textbf{continuous change experiments} when using a MLP. Per column, the best result is written in bold. In case a result is less optimal and not significantly different from the best result, with a confidence of 99\%, it is also written in bold.}
\label{tab:continuous_change_avg_acc}
\end{table*}

\begin{figure*}
\centering
\begin{subfigure}{1.0\textwidth}
    \includegraphics[width=.3\textwidth]{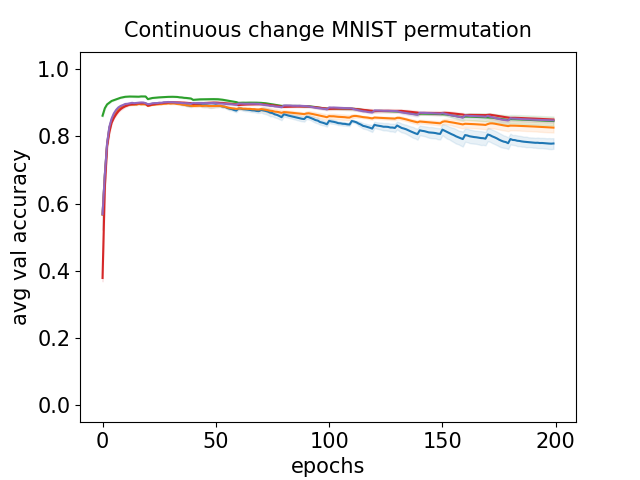}\hfill
    \includegraphics[width=.3\textwidth]{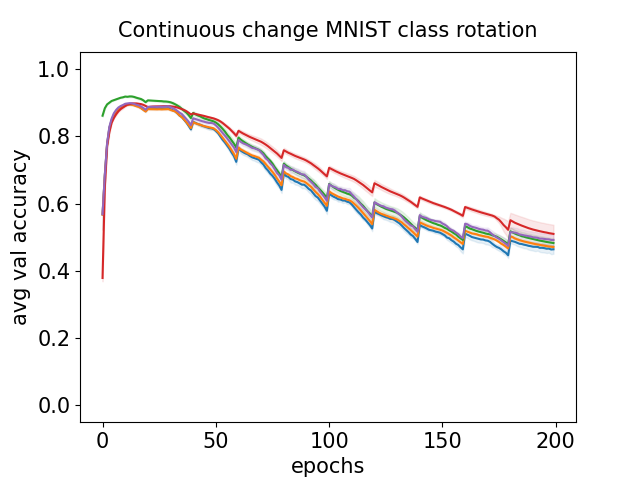}\hfill
    \includegraphics[width=.3\textwidth]{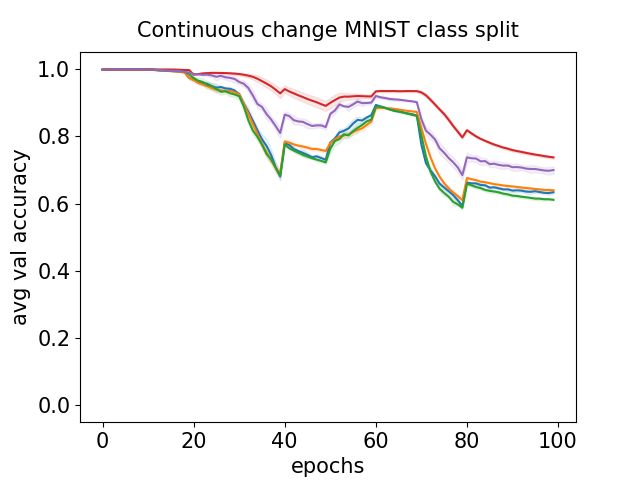}
    \caption{Average validation accuracy}
    \label{fig:continuous_change_MNIST_s1}
\end{subfigure}\\
\begin{subfigure}{1.0\textwidth}
    \includegraphics[width=.3\textwidth]{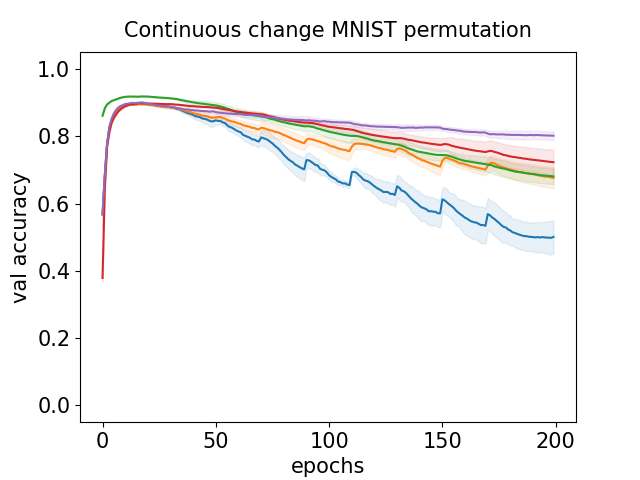}\hfill
    \includegraphics[width=.3\textwidth]{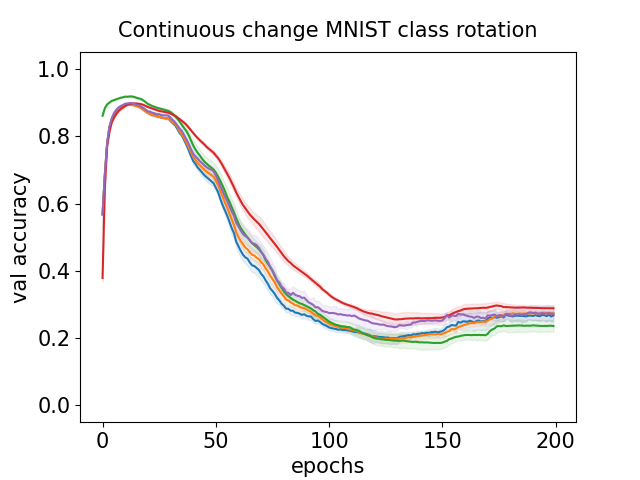}\hfill
    \includegraphics[width=.3\textwidth]{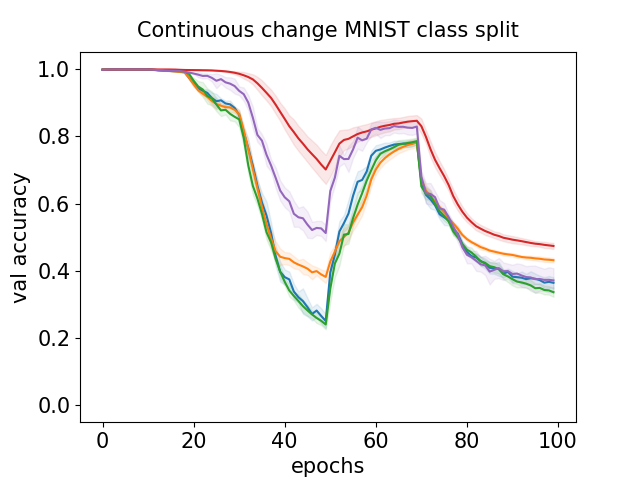}
    \caption{Validation accuracy of the first task}
    \label{fig:continuous_change_MNIST_s2}
\end{subfigure}\\
\begin{subfigure}{1.0\textwidth}
    \includegraphics[width=.3\textwidth]{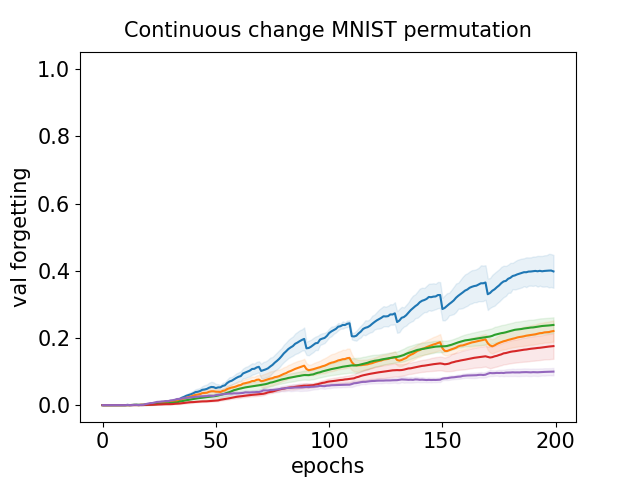}\hfill
    \includegraphics[width=.3\textwidth]{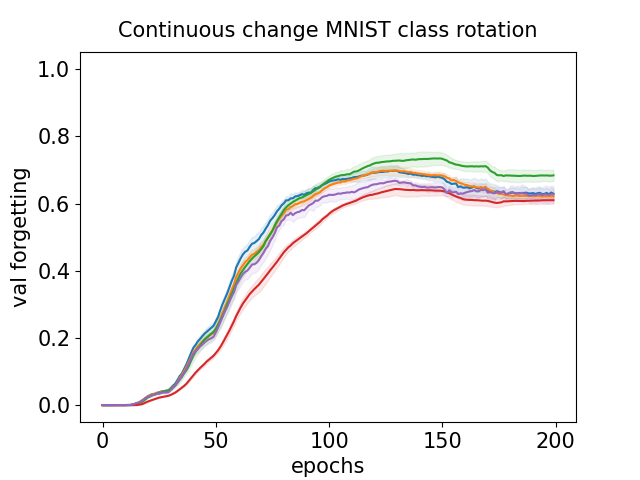}\hfill
    \includegraphics[width=.3\textwidth]{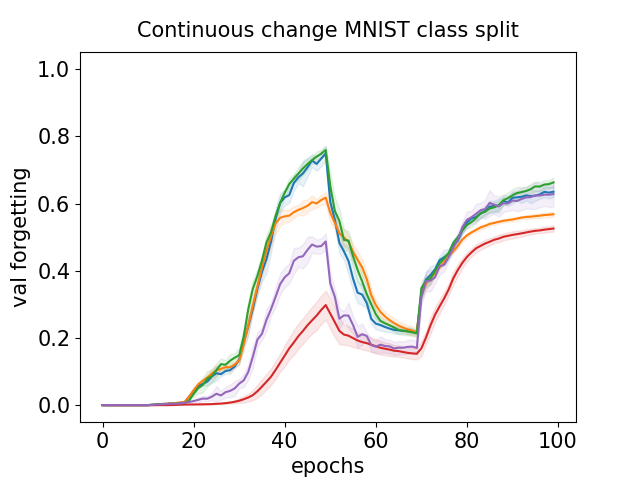}
    \caption{Forgetting on validation accuracy of the first task}
    \label{fig:continuous_change_MNIST_s3}
\end{subfigure}\\
\begin{subfigure}{1.0\textwidth}
    \centering
    \includegraphics[width=1.0\textwidth]{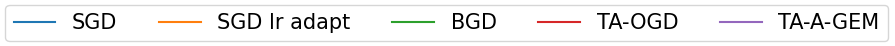}
    \label{fig:continuous_change_MNIST_legend}
\end{subfigure}
\caption{Results of the ``\textbf{Continuous change experiments}'' on the \textbf{MNIST} dataset. (a) Average validation accuracy, (b) Validation accuracy of the first task, (c) Forgetting on validation accuracy of the first task, averaged over all tasks trained thus far, then averaged over five runs, depicted by a solid line plot with $\pm$ one standard deviation as a shaded area. From left to right: task separation by permutation, task separation by rotation and task separation by class split.}
\label{fig:continuous_change_MNIST}
\end{figure*}

\begin{figure*}
\centering
\begin{subfigure}{1.0\textwidth}
    \includegraphics[width=.3\textwidth]{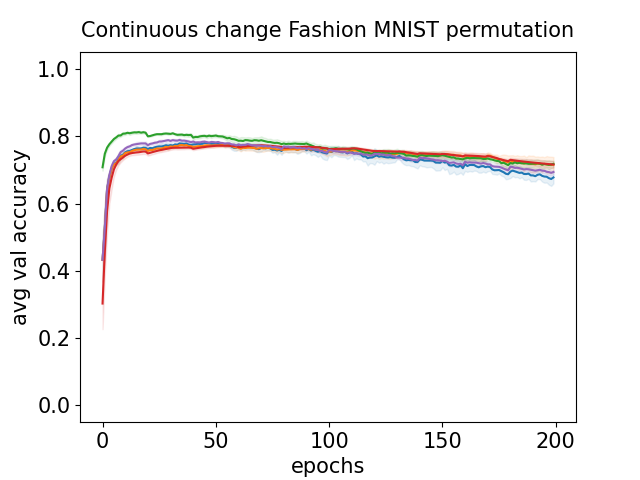}\hfill
    \includegraphics[width=.3\textwidth]{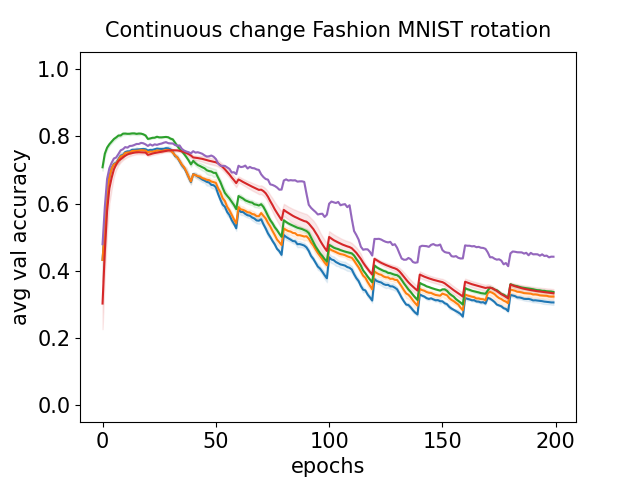}\hfill
    \includegraphics[width=.3\textwidth]{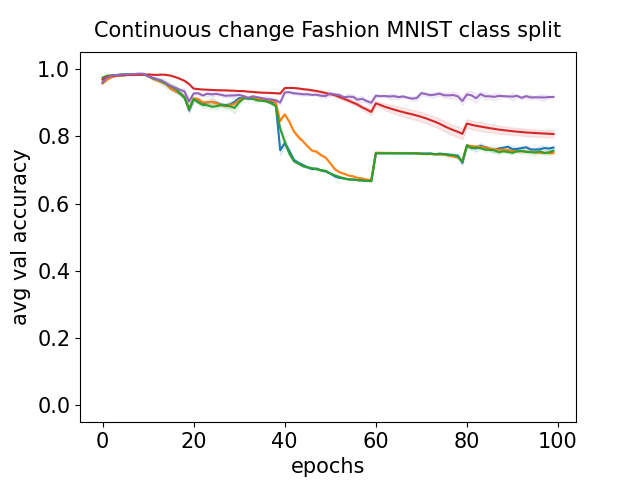}
    \caption{Average validation accuracy}
    \label{fig:continuous_change_Fashion_MNIST_s1}
\end{subfigure}\\
\begin{subfigure}{1.0\textwidth}
    \includegraphics[width=.3\textwidth]{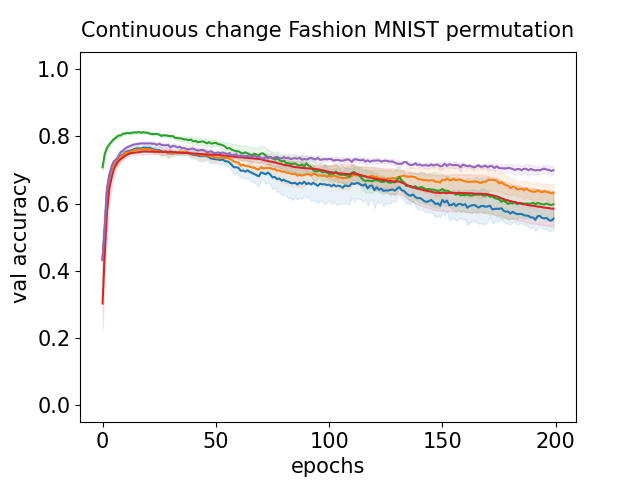}\hfill
    \includegraphics[width=.3\textwidth]{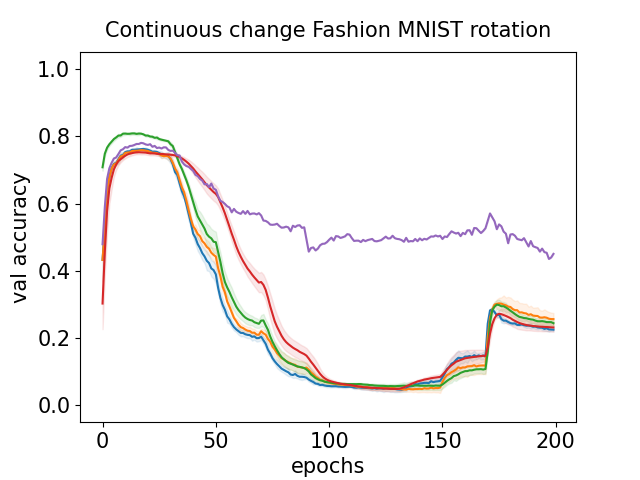}\hfill
    \includegraphics[width=.3\textwidth]{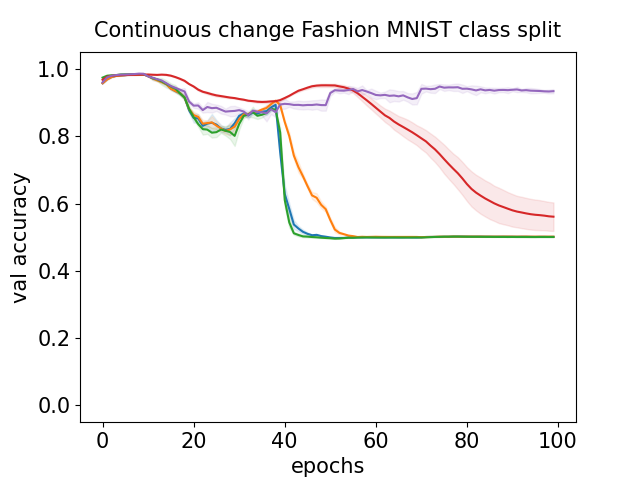}
    \caption{Validation accuracy of the first task}
    \label{fig:continuous_change_Fashion_MNIST_s2}
\end{subfigure}\\
\begin{subfigure}{1.0\textwidth}
    \includegraphics[width=.3\textwidth]{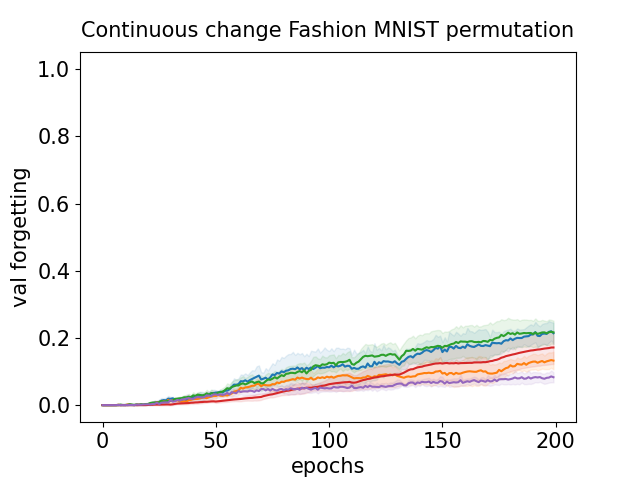}\hfill
    \includegraphics[width=.3\textwidth]{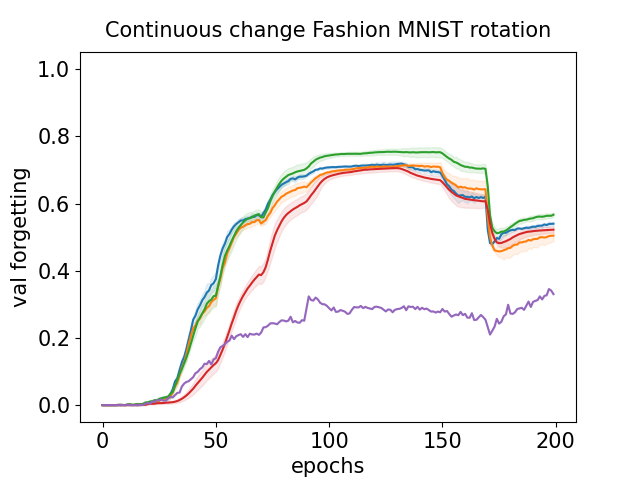}\hfill
    \includegraphics[width=.3\textwidth]{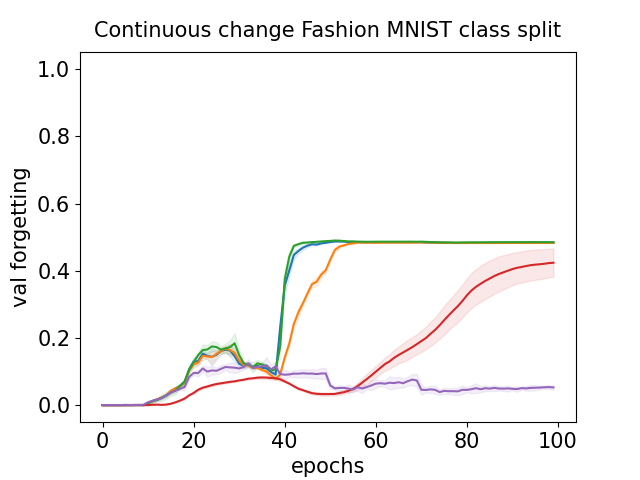}
    \caption{Forgetting on validation accuracy of the first task}
    \label{fig:continuous_change_Fashion_MNIST_s3}
\end{subfigure}\\
\begin{subfigure}{1.0\textwidth}
    \centering
    \includegraphics[width=1.0\textwidth]{figures/continuous_change_experiments/continuous_change_legend.png}
    \label{fig:continuous_change_Fashion_MNIST_legend}
\end{subfigure}
\caption{Results of the ``\textbf{Continuous change experiments}'' on the \textbf{Fashion MNIST} dataset. (a) Average validation accuracy, (b) Validation accuracy of the first task, (c) Forgetting on validation accuracy of the first task, averaged over all tasks trained thus far, then averaged over five runs, depicted by a solid line plot with $\pm$ one standard deviation as a shaded area. From left to right: task separation by permutation, task separation by rotation and task separation by class split.}
\label{fig:continuous_change_Fashion_MNIST}
\end{figure*}

\begin{figure*}
\centering
\begin{subfigure}{1.0\textwidth}
    \includegraphics[width=.3\textwidth]{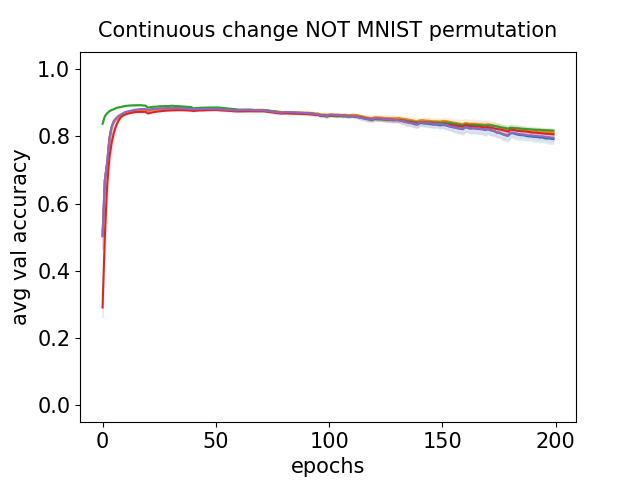}\hfill
    \includegraphics[width=.3\textwidth]{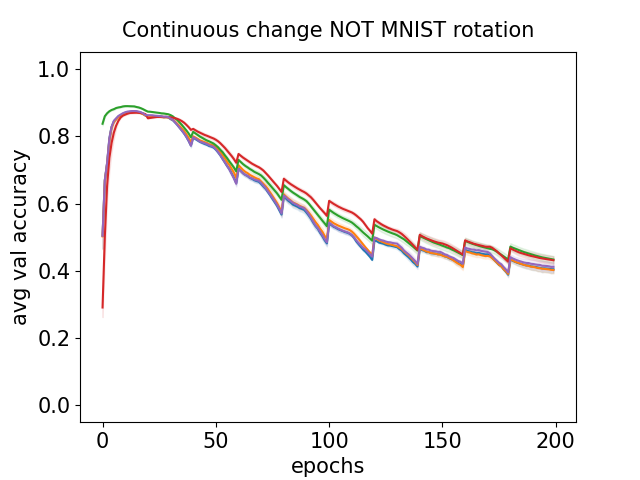}\hfill
    \includegraphics[width=.3\textwidth]{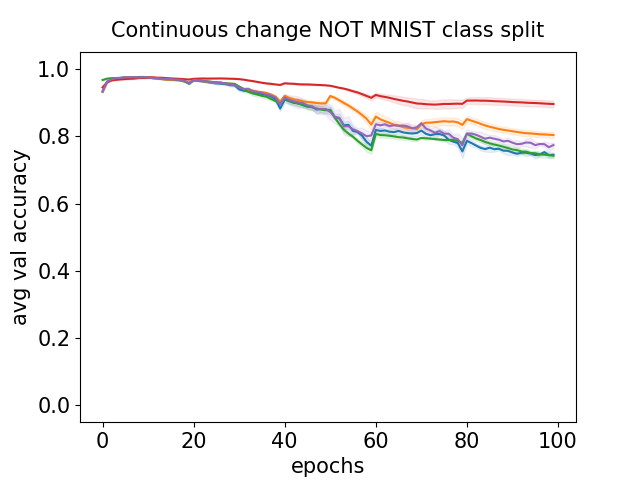}
    \caption{Average validation accuracy}
    \label{fig:continuous_change_NOT_MNIST_s1}
\end{subfigure}\\
\begin{subfigure}{1.0\textwidth}
    \includegraphics[width=.3\textwidth]{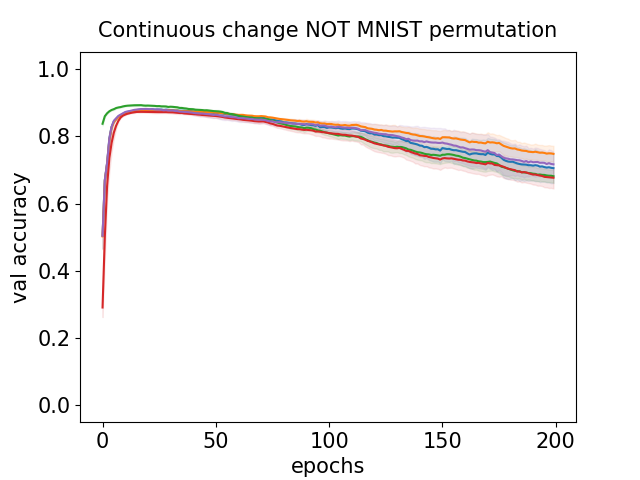}\hfill
    \includegraphics[width=.3\textwidth]{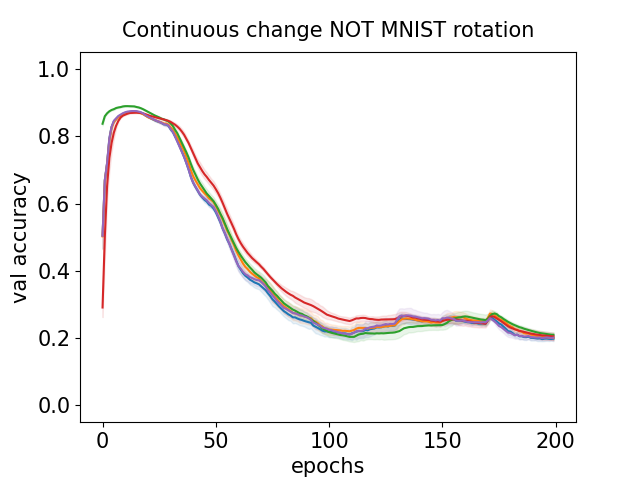}\hfill
    \includegraphics[width=.3\textwidth]{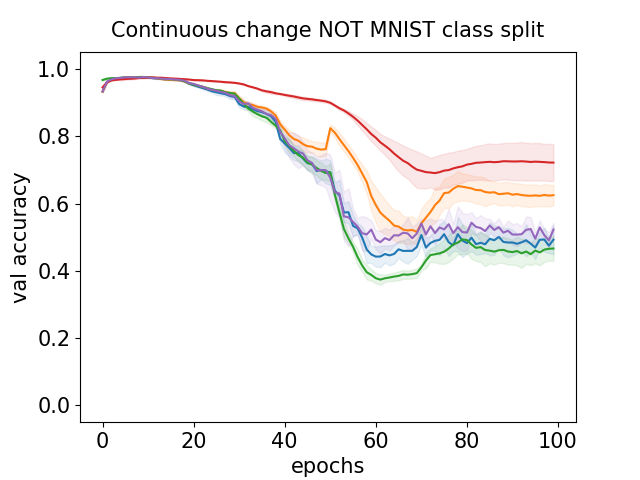}
    \caption{Validation accuracy of the first task}
    \label{fig:continuous_change_NOT_MNIST_s2}
\end{subfigure}\\
\begin{subfigure}{1.0\textwidth}
    \includegraphics[width=.3\textwidth]{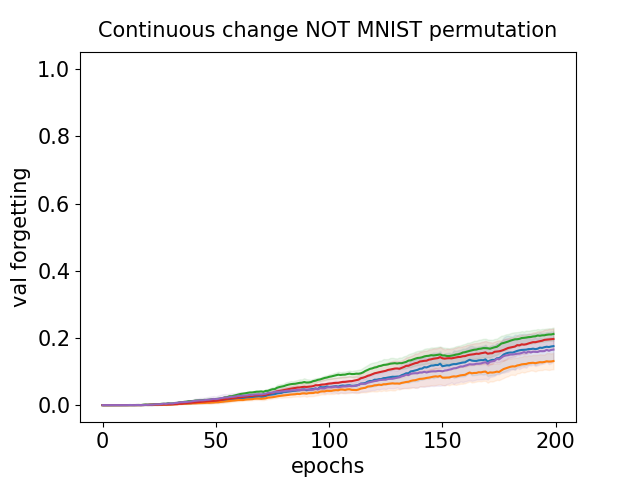}\hfill
    \includegraphics[width=.3\textwidth]{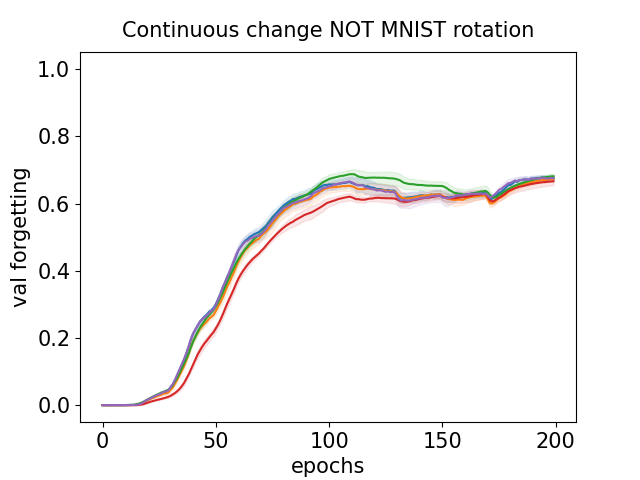}\hfill
    \includegraphics[width=.3\textwidth]{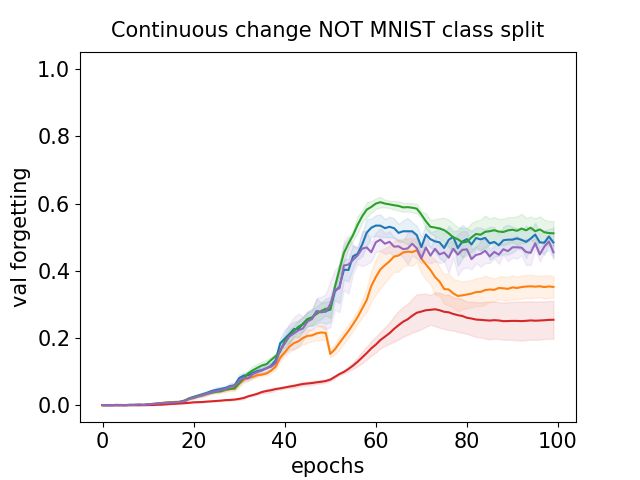}
    \caption{Forgetting on validation accuracy of the first task}
    \label{fig:continuous_change_NOT_MNIST_s3}
\end{subfigure}\\
\begin{subfigure}{1.0\textwidth}
    \centering
    \includegraphics[width=1.0\textwidth]{figures/continuous_change_experiments/continuous_change_legend.png}
    \label{fig:continuous_change_NOT_MNIST_legend}
\end{subfigure}
\caption{Results of the ``\textbf{Continuous change experiments}'' on the \textbf{NOT MNIST} dataset. (a) Average validation accuracy, (b) Validation accuracy of the first task, (c) Forgetting on validation accuracy of the first task, averaged over all tasks trained thus far, then averaged over five runs, depicted by a solid line plot with $\pm$ one standard deviation as a shaded area. From left to right: task separation by permutation, task separation by rotation and task separation by class split.}
\label{fig:continuous_change_NOT_MNIST}
\end{figure*}

\begin{figure*}
\centering
\begin{subfigure}{1.0\textwidth}
    \includegraphics[width=.3\textwidth]{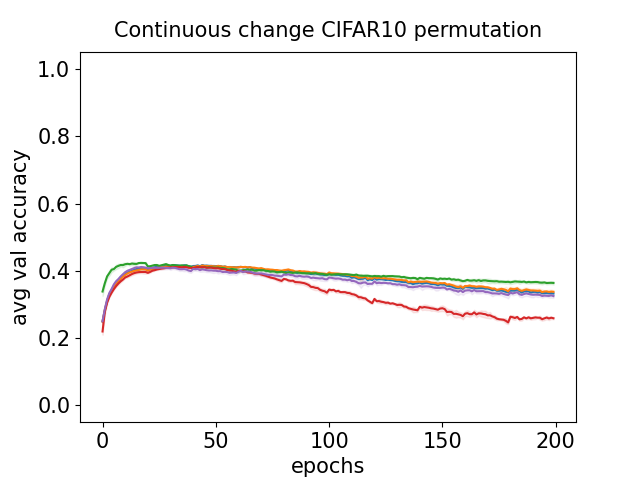}\hfill
    \includegraphics[width=.3\textwidth]{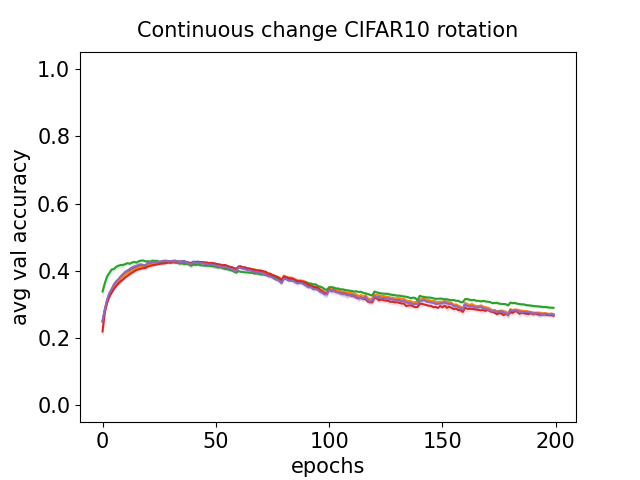}\hfill
    \includegraphics[width=.3\textwidth]{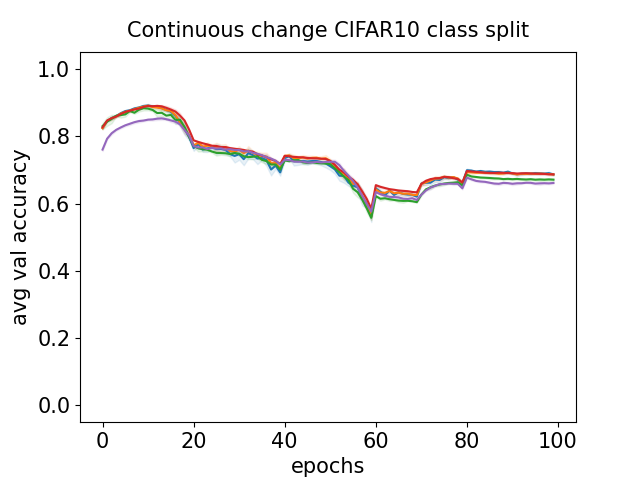}
    \caption{Average validation accuracy}
    \label{fig:continuous_change_CIFAR10_s1}
\end{subfigure}\\
\begin{subfigure}{1.0\textwidth}
    \includegraphics[width=.3\textwidth]{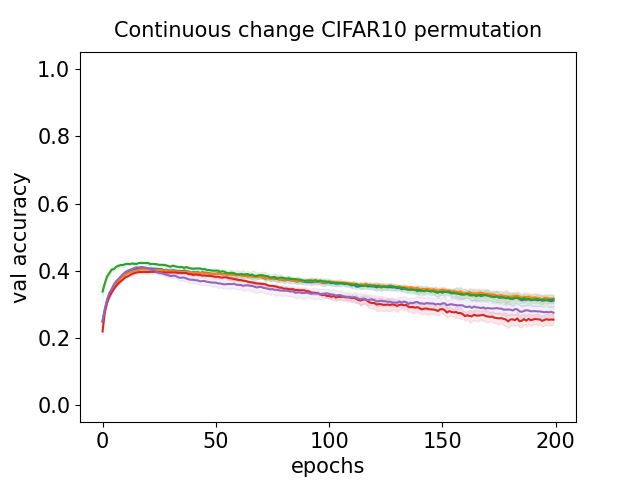}\hfill
    \includegraphics[width=.3\textwidth]{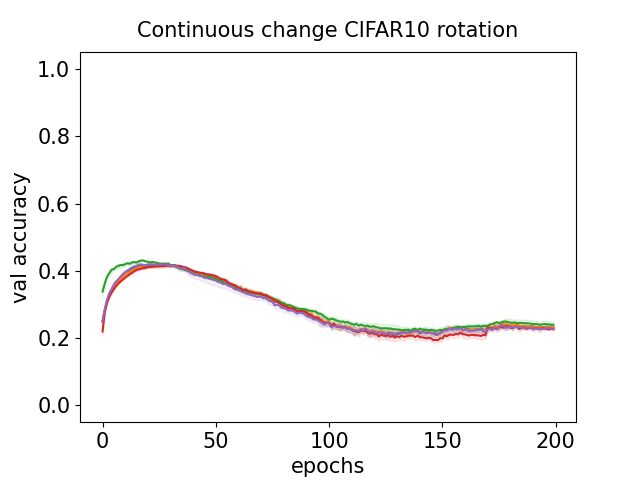}\hfill
    \includegraphics[width=.3\textwidth]{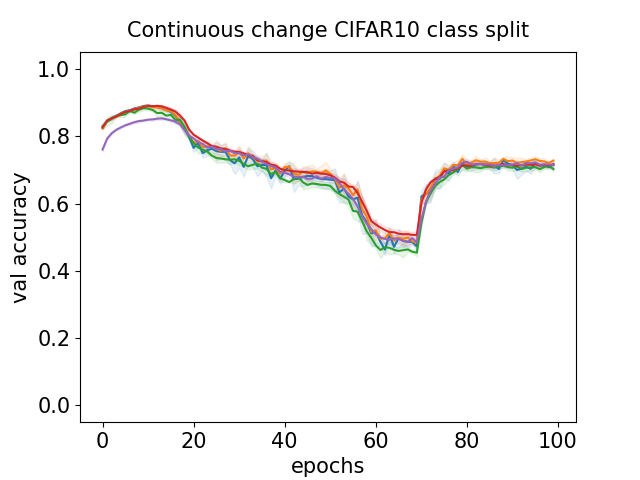}
    \caption{Validation accuracy of the first task}
    \label{fig:continuous_change_CIFAR10_s2}
\end{subfigure}\\
\begin{subfigure}{1.0\textwidth}
    \includegraphics[width=.3\textwidth]{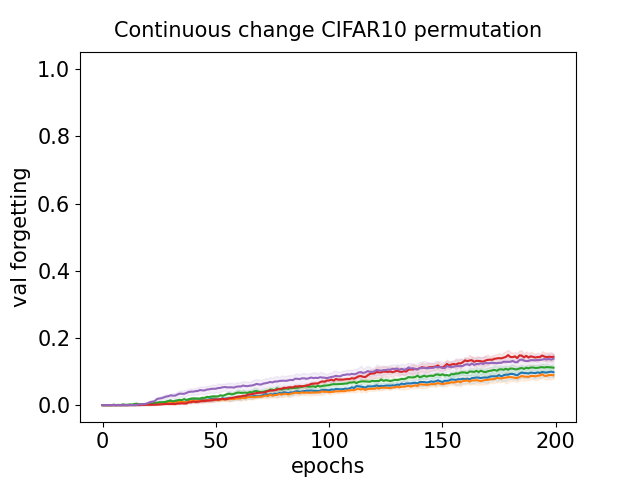}\hfill
    \includegraphics[width=.3\textwidth]{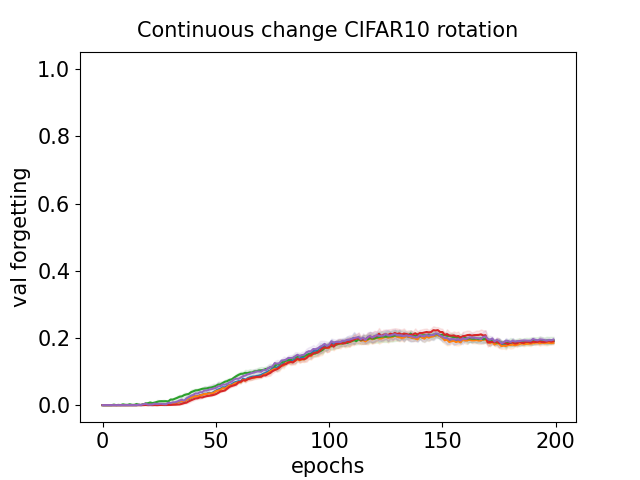}\hfill
    \includegraphics[width=.3\textwidth]{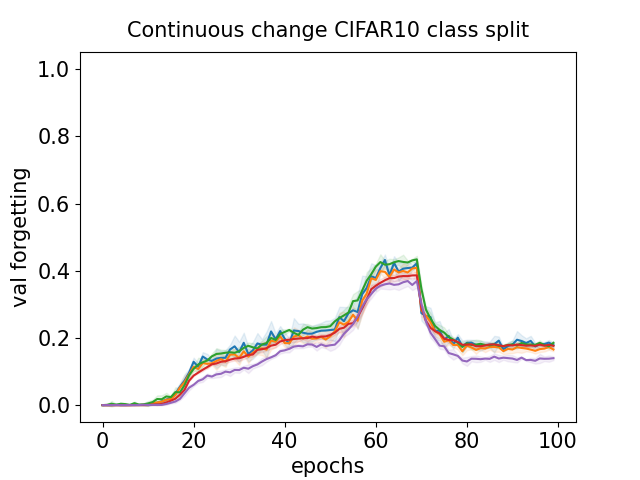}
    \caption{Forgetting on validation accuracy of the first task}
    \label{fig:continuous_change_CIFAR10_s3}
\end{subfigure}\\
\begin{subfigure}{1.0\textwidth}
    \centering
    \includegraphics[width=1.0\textwidth]{figures/continuous_change_experiments/continuous_change_legend.png}
    \label{fig:continuous_change_CIFAR10_legend}
\end{subfigure}
\caption{Results of the ``\textbf{Continuous change experiments}'' on the \textbf{CIFAR10} dataset. (a) Average validation accuracy, (b) Validation accuracy of the first task, (c) Forgetting on validation accuracy of the first task, averaged over all tasks trained thus far, then averaged over five runs, depicted by a solid line plot with $\pm$ one standard deviation as a shaded area. From left to right: task separation by permutation, task separation by rotation and task separation by class split.}
\label{fig:continuous_change_CIFAR10}
\end{figure*}

\begin{figure*}
\centering
\begin{subfigure}{1.0\textwidth}
    \includegraphics[width=.3\textwidth]{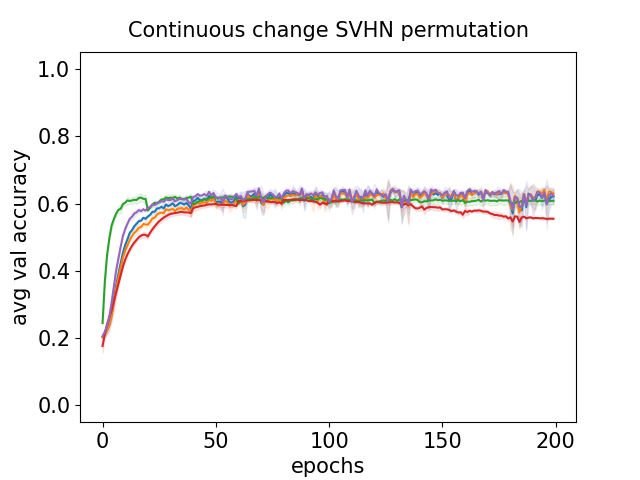}\hfill
    \includegraphics[width=.3\textwidth]{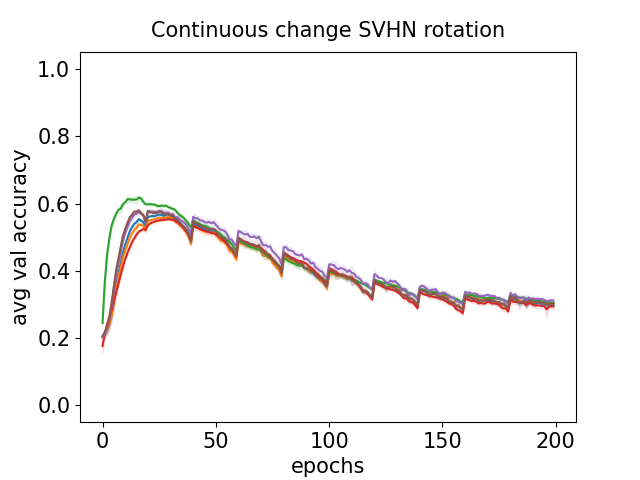}\hfill
    \includegraphics[width=.3\textwidth]{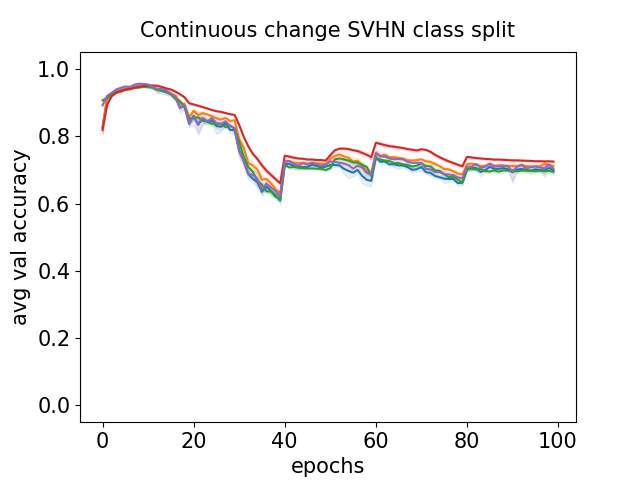}
    \caption{Average validation accuracy}
    \label{fig:continuous_change_SVHN_s1}
\end{subfigure}\\
\begin{subfigure}{1.0\textwidth}
    \includegraphics[width=.3\textwidth]{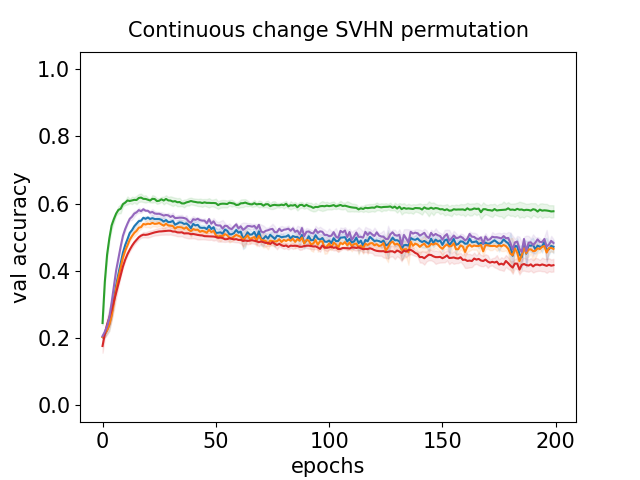}\hfill
    \includegraphics[width=.3\textwidth]{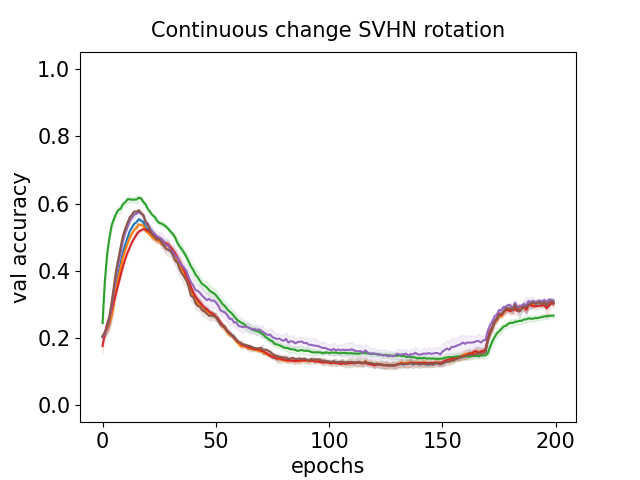}\hfill
    \includegraphics[width=.3\textwidth]{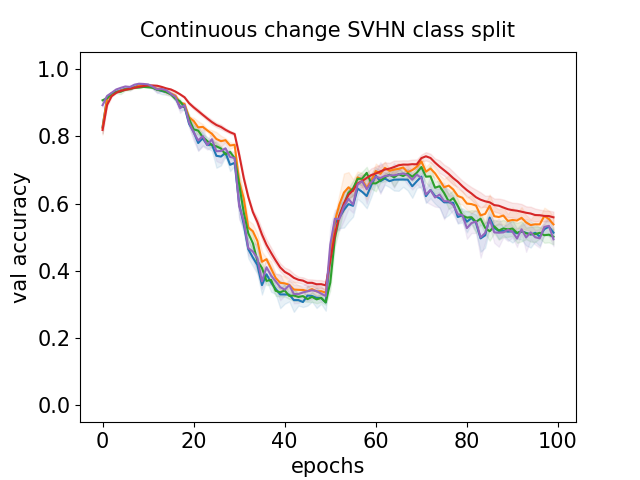}
    \caption{Validation accuracy of the first task}
    \label{fig:continuous_change_SVHN_s2}
\end{subfigure}\\
\begin{subfigure}{1.0\textwidth}
    \includegraphics[width=.3\textwidth]{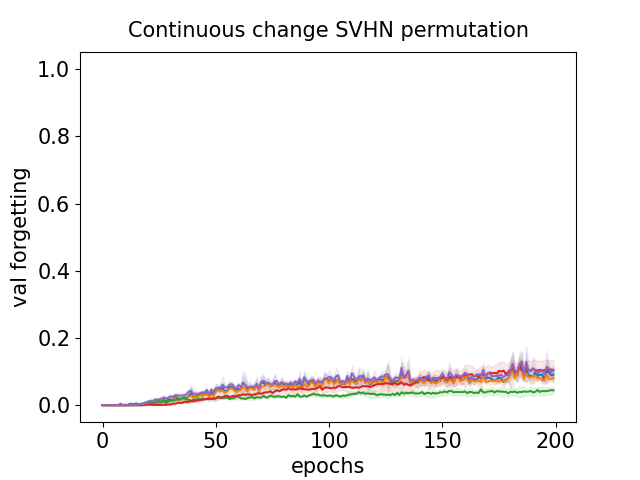}\hfill
    \includegraphics[width=.3\textwidth]{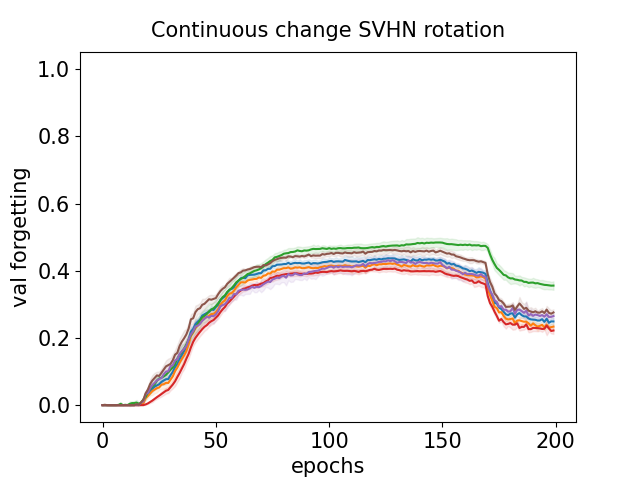}\hfill
    \includegraphics[width=.3\textwidth]{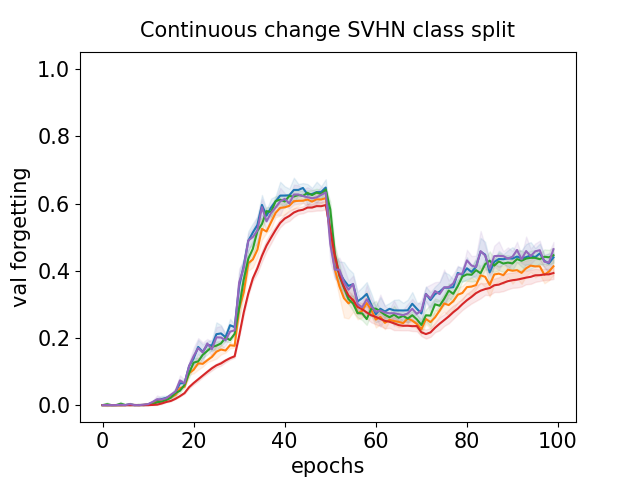}
    \caption{Forgetting on validation accuracy of the first task}
    \label{fig:continuous_change_SVHN_s3}
\end{subfigure}\\
\begin{subfigure}{1.0\textwidth}
    \centering
    \includegraphics[width=1.0\textwidth]{figures/continuous_change_experiments/continuous_change_legend.png}
    \label{fig:continuous_change_SVHN_legend}
\end{subfigure}
\caption{Results of the ``\textbf{Continuous change experiments}'' on the \textbf{SVHN} dataset. (a) Average validation accuracy, (b) Validation accuracy of the first task, (c) Forgetting on validation accuracy of the first task, averaged over all tasks trained thus far, then averaged over five runs, depicted by a solid line plot with $\pm$ one standard deviation as a shaded area. From left to right: task separation by permutation, task separation by rotation and task separation by class split.}
\label{fig:continuous_change_SVHN}
\end{figure*}

Figures \ref{fig:continuous_change_MNIST},  \ref{fig:continuous_change_Fashion_MNIST}, \ref{fig:continuous_change_NOT_MNIST}, \ref{fig:continuous_change_CIFAR10} and \ref{fig:continuous_change_SVHN} show the average validation accuracy during training for different task split methods, using the MNIST, Fashion MNIST, NOT MNIST, CIFAR10 and SVHN datasets, for the ``Continuous change'' experiments. Table \ref{tab:continuous_change_acc} and \ref{tab:continuous_change_forg} summarize the results on the validation accuracy on the first task and the forgetting on the validation accuracy on the first task, respectively. Again, all results are the average of five independent runs.

\begin{table*}
\centering
\scalebox{0.8}{
\begin{tabular}{c|ccc|ccc|ccc|ccc|ccc}
    & \multicolumn{3}{c|}{MNIST} & \multicolumn{3}{c|}{Fashion MNIST} & \multicolumn{3}{c|}{NOT MNIST} & \multicolumn{3}{c|}{CIFAR10} & \multicolumn{3}{c}{SVHN}\\
    & perm & rot & class & perm & rot & class & perm & rot & class& perm & rot & class & perm & rot & class \\
    \hline
    SGD & 0.702 & 0.414 & 0.657 & 0.659 & 0.267 & 0.665 & \textbf{0.810} & 0.395 & 0.694 & \textbf{0.359} & 0.289 & 0.707 & 0.490 & 0.230 & 0.626\\
    SGD lr adapt & 0.785 & 0.420 & 0.672 & 0.693 & 0.278 & 0.684 & \textbf{0.825} & 0.404 & 0.769 & \textbf{0.361} & 0.289 & \textbf{0.717} & 0.479 & 0.228 & 0.658\\
    BGD & 0.812 & 0.426 & 0.645 & \textbf{0.702} & 0.297 & 0.663 & \textbf{0.804} & 0.409 & 0.675 & \textbf{0.366} & \textbf{0.299} & 0.695 & \textbf{0.588} & \textbf{0.264} & 0.636\\
    TA-OGD & 0.819 & \textbf{0.476} & \textbf{0.803} & 0.682 & 0.317 & 0.842 & 0.789 & \textbf{0.422} & \textbf{0.851} & 0.325 & 0.284 & \textbf{0.719} & 0.457 & 0.228 & \textbf{0.681}\\
    TA-A-GEM & \textbf{0.842} & 0.442 & 0.726 & \textbf{0.729} & \textbf{0.600} & \textbf{0.926} & \textbf{0.813} & 0.398 & 0.712 & 0.331 & 0.286 & 0.704 & 0.509 & 0.233 & 0.634\\
\end{tabular}
}
\caption{\textbf{Validation accuracy} of the first task, averaged over all epochs, then averaged over five runs, for the \textbf{continuous change experiments} when using a MLP. Per column, the best result is written in bold. In case a result is less optimal and not significantly different from the best result, with a confidence of 99\%, it is also written in bold.}
\label{tab:continuous_change_acc}
\end{table*}

\begin{table*}
\centering
\scalebox{0.8}{
\begin{tabular}{c|ccc|ccc|ccc|ccc|ccc}
    & \multicolumn{3}{c|}{MNIST} & \multicolumn{3}{c|}{Fashion MNIST} & \multicolumn{3}{c|}{NOT MNIST} & \multicolumn{3}{c|}{CIFAR10} & \multicolumn{3}{c}{SVHN}\\
    & perm & rot & class & perm & rot & class & perm & rot & class& perm & rot & class & perm & rot & class \\
    \hline
    SGD & 0.192 & 0.478 & 0.342 & 0.104 & 0.492 & 0.319 & \textbf{0.066} & 0.475 & 0.281 & \textbf{0.046} & \textbf{0.127} & 0.185 & 0.057 & 0.312 & 0.325\\
    SGD lr adapt & 0.107 & 0.470 & 0.327 & \textbf{0.067} & 0.478 & 0.300 & \textbf{0.049} & 0.464 & 0.207 & \textbf{0.041} & \textbf{0.124} & 0.174 & 0.053 & 0.299 & 0.290\\
    BGD & 0.106 & 0.492 & 0.355 & 0.110 & 0.512 & 0.323 & 0.088 & 0.480 & 0.302 & 0.058 & \textbf{0.132} & 0.190 & \textbf{0.027} & 0.351 & 0.310\\
    TA-OGD & \textbf{0.073} & \textbf{0.416} & \textbf{0.196} & \textbf{0.067} & 0.429 & 0.142 & 0.077 & \textbf{0.441} & \textbf{0.124} & 0.069 & \textbf{0.127} & 0.171 & 0.050 & \textbf{0.283} & \textbf{0.268}\\
    TA-A-GEM & \textbf{0.055} & 0.454 & 0.273 & \textbf{0.047} & \textbf{0.172} & \textbf{0.061} & \textbf{0.064} & 0.473 & 0.265 & 0.078 & \textbf{0.131} & \textbf{0.147} & 0.064 & 0.335 & 0.322\\
\end{tabular}
}
\caption{\textbf{Forgetting} on validation accuracy of the first task, averaged over five runs, then averaged over all epochs, for the \textbf{continuous change experiments} when using a MLP. Per column, the best result is written in bold. In case a result is less optimal and not significantly different from the best result, with a confidence of 99\%, it is also written in bold.}
\label{tab:continuous_change_forg}
\end{table*}

\begin{table*}
\centering
\scalebox{0.8}{
\begin{tabular}{l|l|l}
    Task generation & Mechanism & Parameters\\
    \hline
    Permutation & Number of tasks & 10\\
    Rotation & Rotation angles & [0, 20, 40, 60, 80, 100, 120, 140, 160, 180] \\
    Class split & Subset labels & [[0, 1], [2, 3], [4, 5], [6, 7], [8, 9]]\\
\end{tabular}
}
\caption{Summary of the settings used for the task split for the ``Disjoint tasks'' experiment and the ``Continuous change'' experiment}
\label{tab:tasksplitsettings}
\end{table*}

\begin{table*}
\centering
\scalebox{0.75}{
\begin{tabular}{l|l|l}
    Parameter name & Value & Explanation \\
    \hline
    Learning rate & $10^{-3}$ & The same as for TA-OGD\\
    Batch size & 10 & The same batch size as TA-OGD is chosen.\\
    Sampling rate & $1$ & The number of labeled data points sampled per batch, so in this case, one labeled data point is sampled every batch.\\
    Number of pools & 10 or 2 & The number of separate pools; one for each class, so 10 classes for permutation and rotation and 2 for class split.\\
    Number of clusters & 10 or 50 & Each pool contains 10 clusters in case there are 10 classes, or 50 clusters in case there are 2 classes.\\
    Cluster size & 3 & The maximum number of sampled data points per cluster, if this number is exceeded, the oldest sample is removed.\\
\end{tabular}
}
\caption{TA-A-GEM uses these parameters. Their values were hand picked to maximize the average accuracy.}
\label{tab:taagemsettings}
\end{table*}

\begin{table*}
\centering
\scalebox{0.75}{
\begin{tabular}{l|l|l}
    Parameter name & Value & Explanation \\
    \hline
    Learning rate & $10^{-3}$ & The same as in the OGD paper of Farajtabar et al. \cite{pmlr-v108-farajtabar20a}\\
    Batch size & 10 & TA-OGD needs small batch sizes. Larger batch sizes will not yield good results.\\
    Sampling rate & $1$ & The number of gradients sampled per batch, so in this case, one gradient is sampled every batch.\\
    Number of pools & 1 & TA-OGD uses one pool for all sampled model gradients.\\
    Number of clusters & 99 & The number of distinct model gradient clusters.\\
    Cluster size & 3 & The maximum number of model gradients per cluster, if this number is exceeded, the oldest model gradient is removed.\\
\end{tabular}
}
\caption{TA-OGD uses these parameters. Their values were hand picked to maximize the average accuracy.}
\label{tab:taogdsettings}
\end{table*}

\section{Sampling Rate}
For simplicity, we simply always add one sample per batch and set a maximum pool size. Once the pool is full, for every sample that is added, one is removed. If we remove one sample at random however, samples that are relevant to previous tasks quickly disappear from the pool. In this case, the pool itself suffers from catastrophic forgetting. 
The easiest way to fix this is to lower the sampling rate, i.e. the frequency at which samples are added to the pool. In pre-experiments, we found out that setting the sampling rate to $0.01$ gave optimal results in negating the forgetting. This effectively means that one sample was added to the pool every $100$ batches. As it turned out, this gave the algorithm enough time to take a decent amount of samples for a task, while it did not take so many samples as to completely eliminate samples from older tasks from the pool. We then realized however, that this ``optimal'' sampling rate was completely dependent on the fact that every task was presented to the model for $20$ epochs. It thus completely depends on the knowledge that a task will change after $20$ epochs. In a truly task-agnostic setting, this information is not known. We therefore consider this method that solely relies on the ``optimal'' sampling rate to be a trivial case that is useless in a task-agnostic setting.


\section{Clustering versus Random sampling}
\begin{figure}
\centering
\includegraphics[width=.6\textwidth]{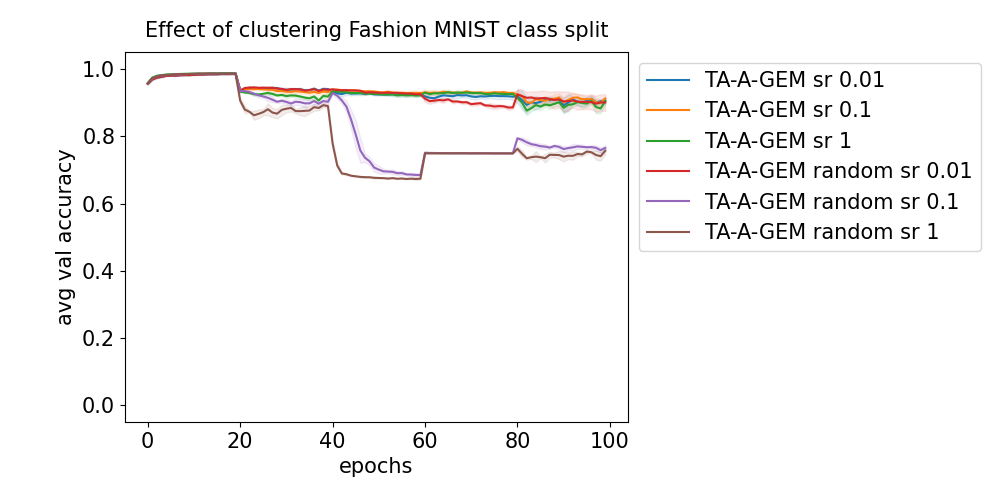}
\caption{TA-A-GEM without its clustering mechanism (TA-A-GEM random) can only successfully negate forgetting with an ``optimal'' sampling rate (sr) of 0.01, that can only be known if the task change frequency is known, thus undoing its task-agnostic nature. On the other hand, when clustering is introduced to TA-A-GEM (TA-A-GEM), a wide variety of sampling rates (sr) (0.01, 0.1 and 1) can effectively reduce the amount of forgetting.}
\label{fig:analyze_clusters_plotted}
\end{figure}

Figure \ref{fig:analyze_clusters_plotted} demonstrates that using clustering is a necessity. It shows that without the use of clustering (TA-A-GEM random), TA-A-GEM can only counteract the forgetting with an ``optimal'' sampling rate (sr) of $0.01$, that is completely dependent on the frequency of task change; once every $20$ epochs in this case. When clustering is introduced (TA-A-GEM), a wide variety of sampling rates, $0.01$, $0.1$ and $1$ can effectively negate forgetting.

\begin{figure*}
\centering
\includegraphics[width=.5\textwidth]{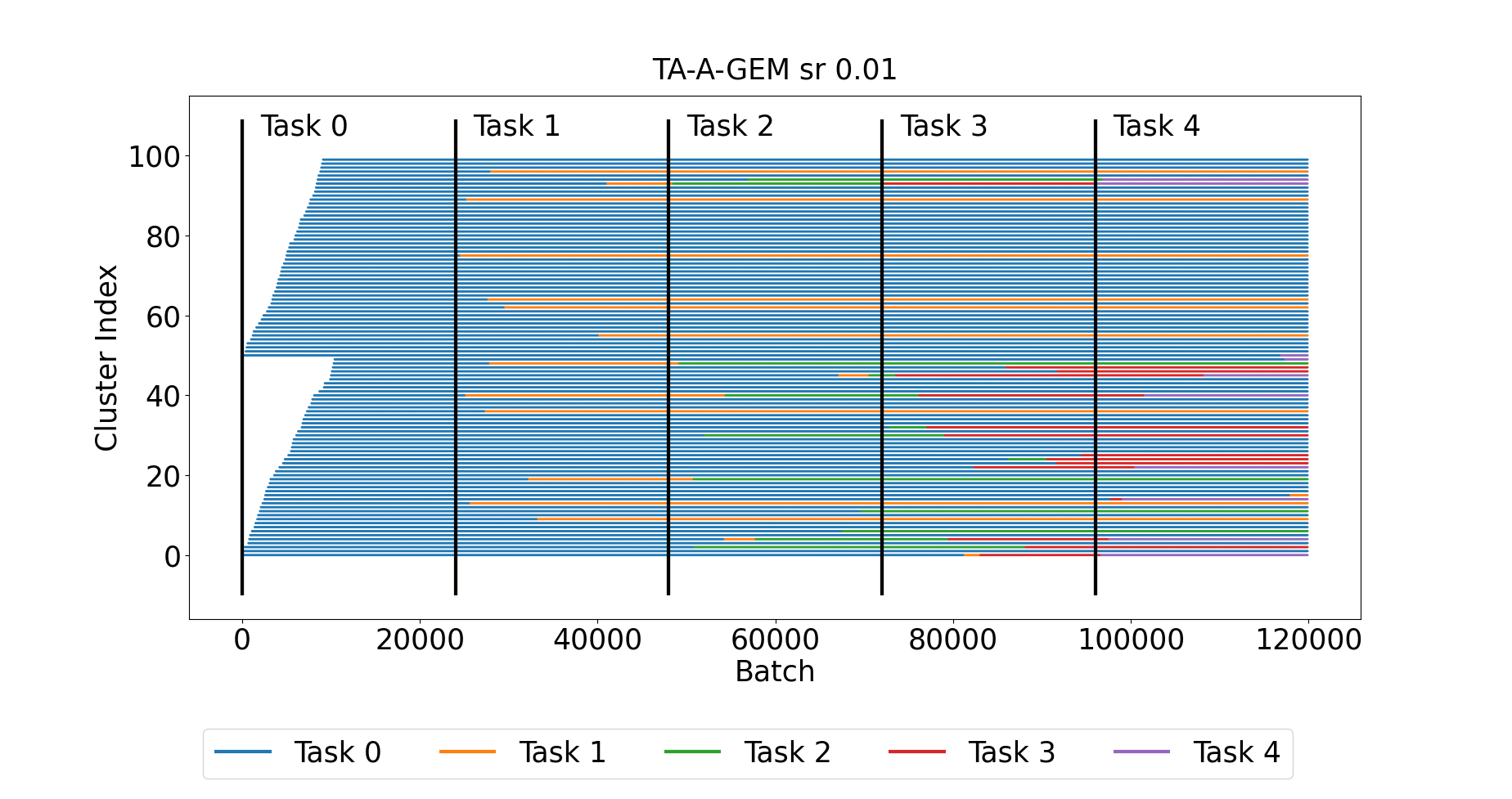}\hfill
\includegraphics[width=.5\textwidth]{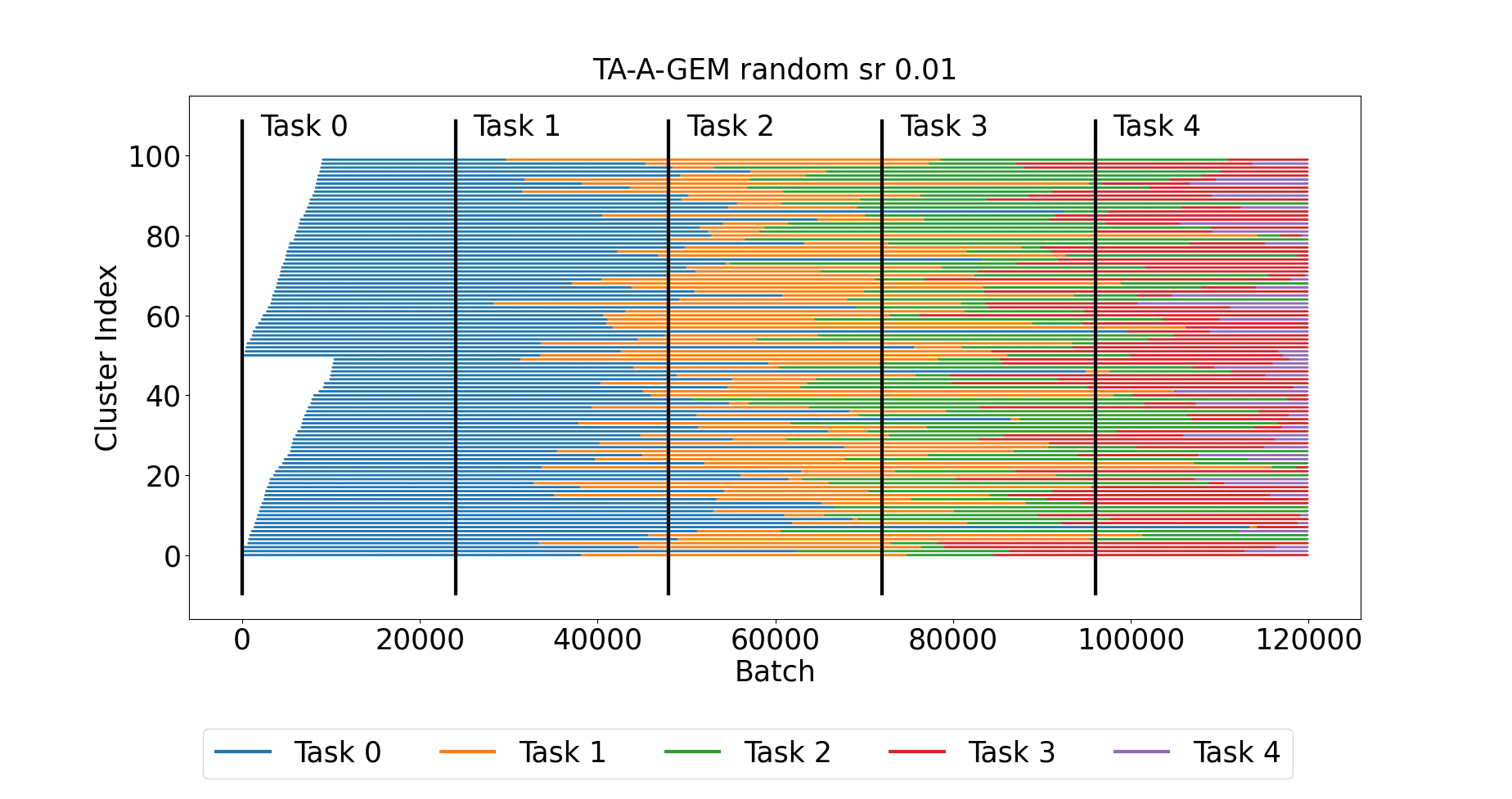}
\caption{The horizontal lines depict the content of clusters when using class split as a task segmentation method on Fashion MNIST with a sampling rate of 0.01. Each task is associated with a unique color. The color of the vertical lines represents the oldest task information that is present in the cluster. The moment that a new task starts is indicated by a black vertical line. The clusters with index 0 to 49 correspond to the pool of class label 0. The clusters with index 50 to 99 correspond to the pool of class label 1. Left: Clustering helps in keeping a greater variety of task information in the gradient pool, but this effect is hardly noticable due to the optimal sampling rate being used. Right: Using random cluster assignment can result in a good amount of variety, but it is completely dependent on the optimal sampling rate of 0.01. but it is still just as effective at counteracting forgetting, as compared to clustering. This is only ideal when every task presents itself for exactly 20 epochs each time. For this, the frequency of task switching need to be known, thereby destroying the task-agnostic nature of the method.}
\label{fig:analyze_clusters_0_01}
\end{figure*}

\begin{figure*}
\centering
\includegraphics[width=.5\textwidth]{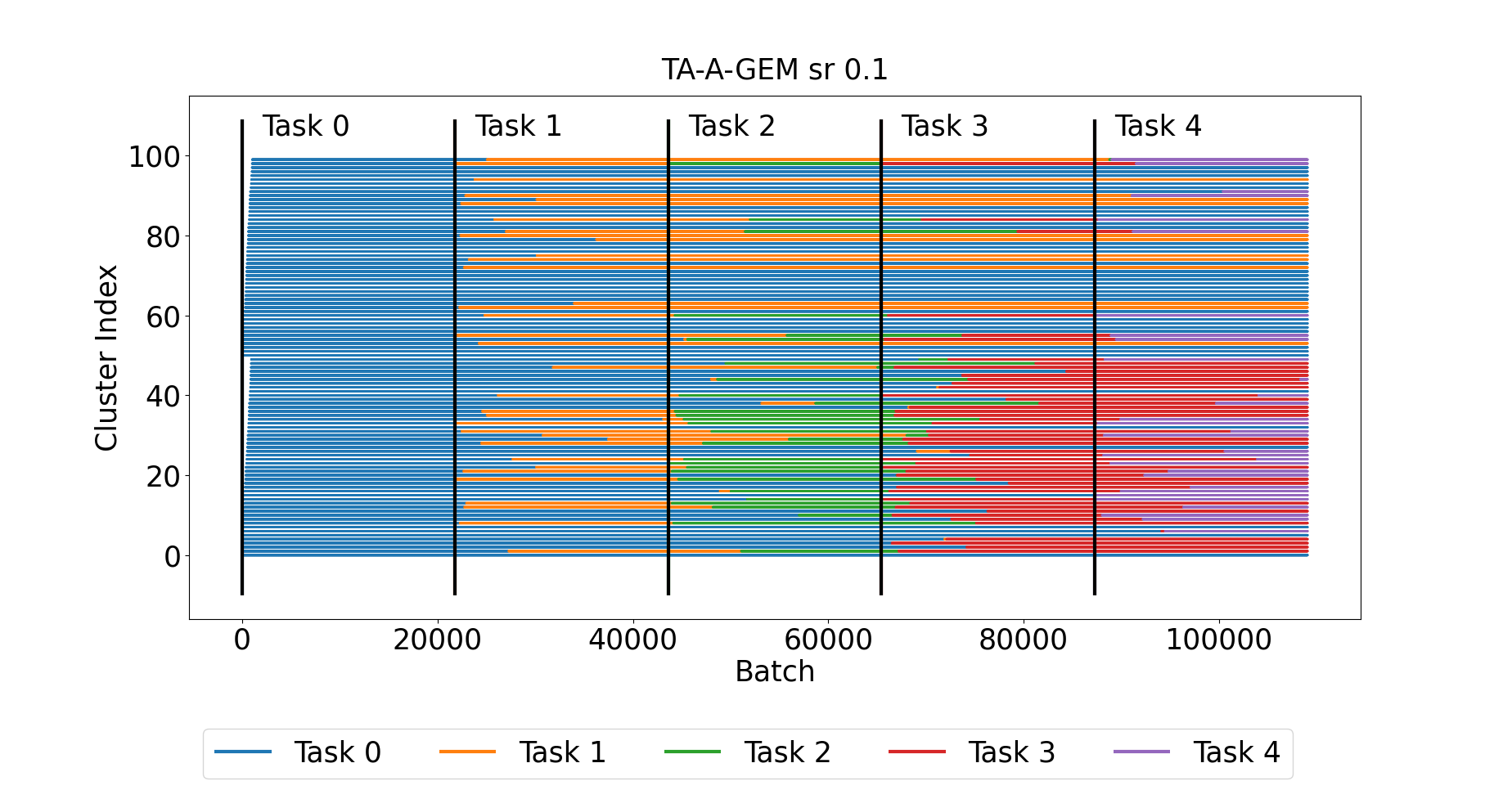}\hfill
\includegraphics[width=.5\textwidth]{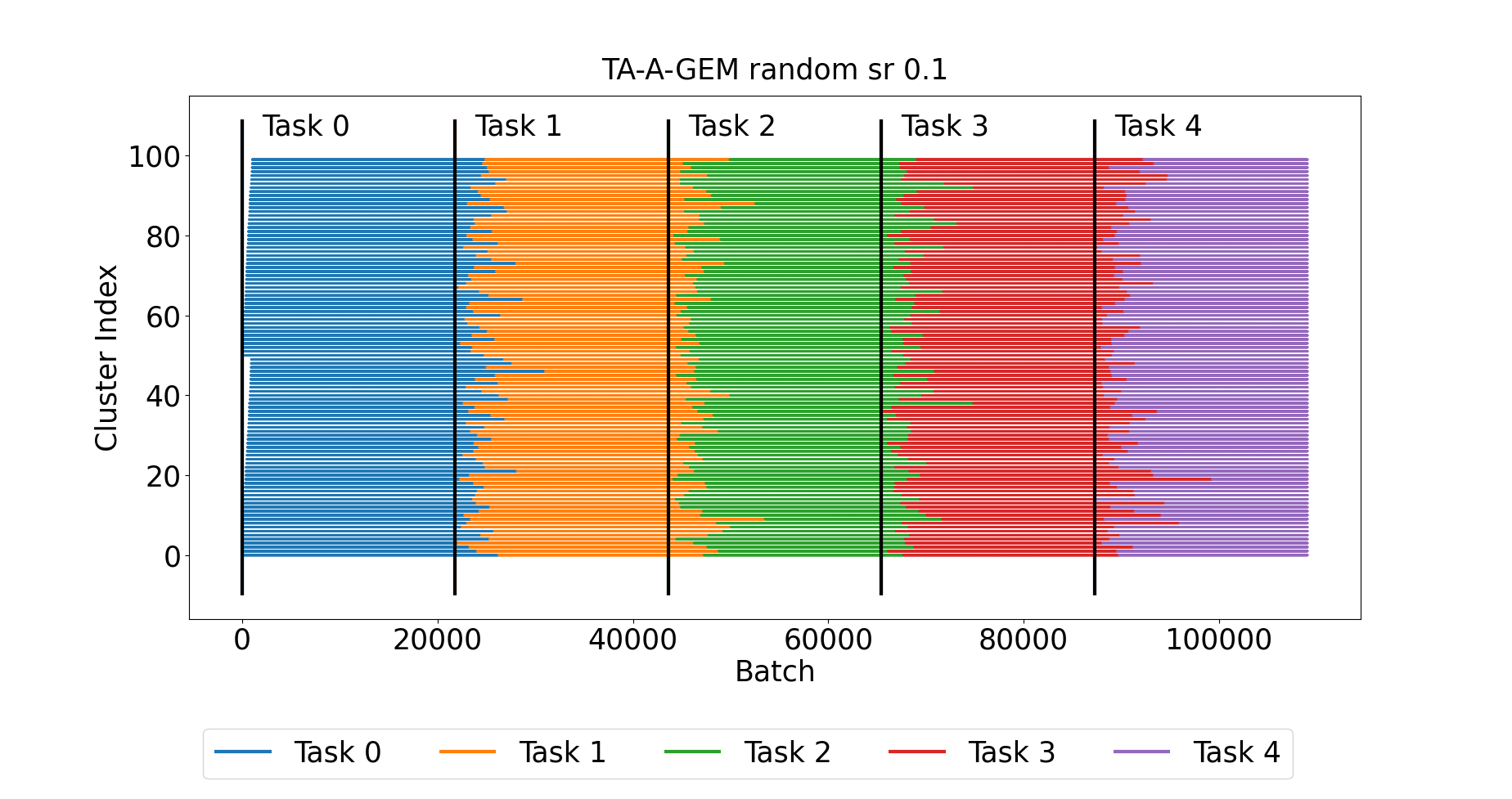}
\caption{The content of clusters using when using class split as a task segmentation method on Fashion MNIST with a sampling rate of 0.1. Left: The use of clustering assures a good variety of task information is kept in the pool. Right: Information from previous tasks is quickly lost with this sampling rate setting.}
\label{fig:analyze_clusters_0_1}
\end{figure*}

\begin{figure*}
\centering
\includegraphics[width=.5\textwidth]{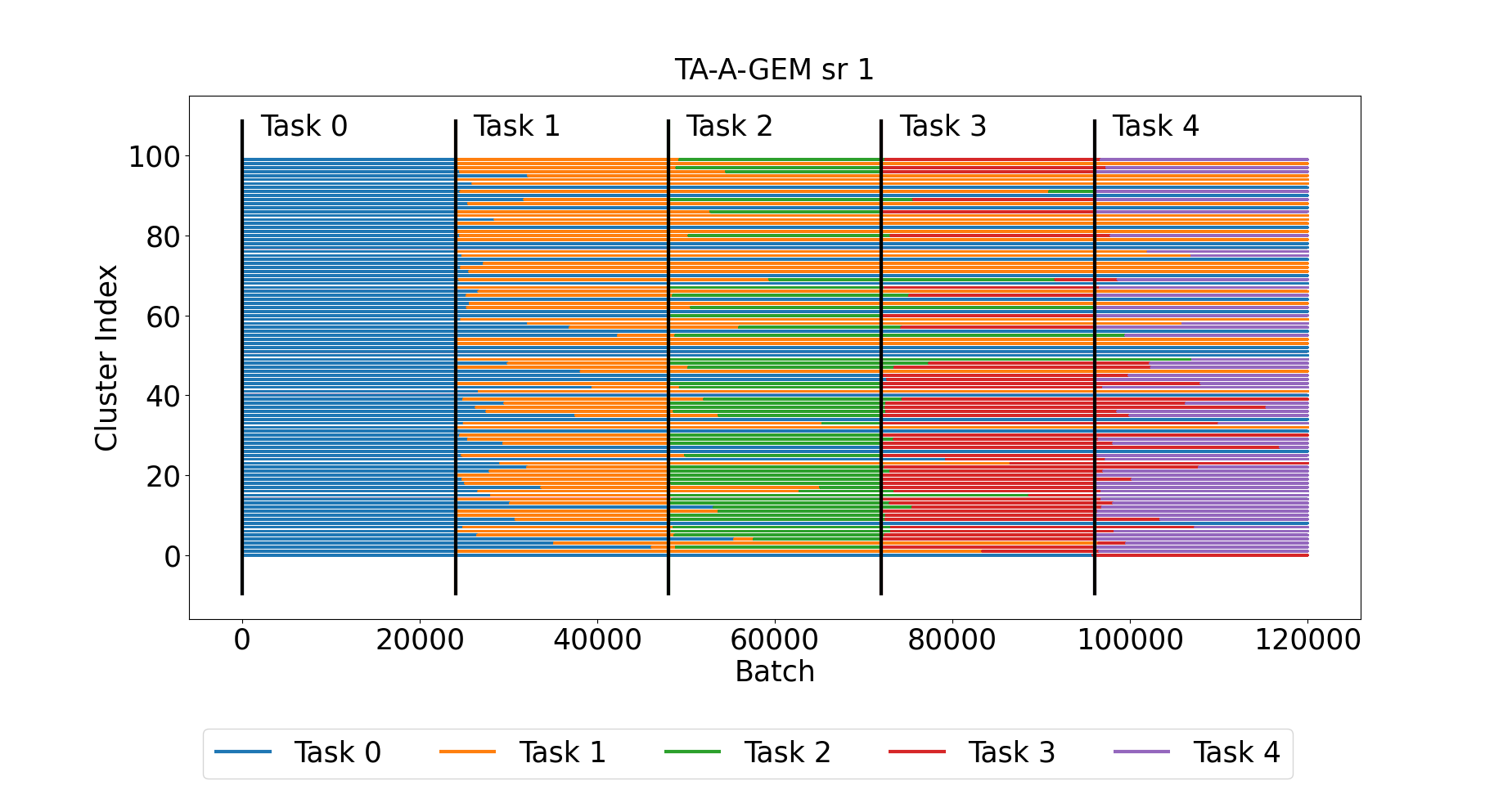}\hfill
\includegraphics[width=.5\textwidth]{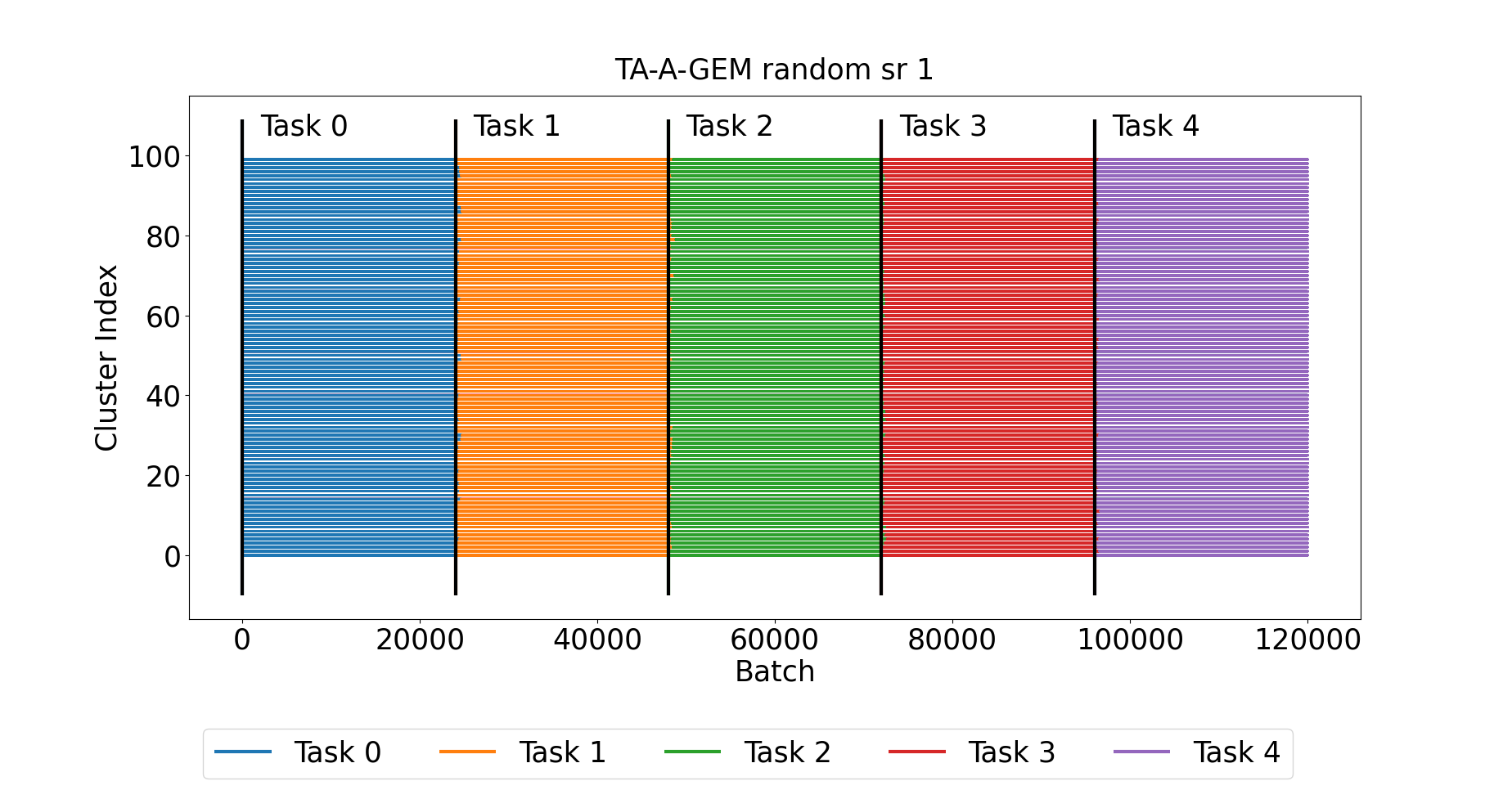}
\caption{The content of clusters when using class split as a task segmentation method on Fashion MNIST with a sampling rate of 1. Left: Even with a high sampling rate, the clustering assures a good variety of task information in the pool. Right: Information from previous tasks is immediately lost after the moment of task change.}
\label{fig:analyze_clusters_1}
\end{figure*}

To further investigate the effect of clustering on the model gradient pool, the content of the pool is plotted over time for TA-A-GEM, that uses clustering and TA-A-GEM with random cluster assignment. In the latter, a newly sampled data point is added to a randomly selected cluster instead of being added to the cluster that has its mean closest to it (as is the case in the standard clustering-based TA-A-GEM approach). We test the performance of the two methods for three  different values of the sampling rate i.e., 0.01, 0.1 and 1.
Figures \ref{fig:analyze_clusters_0_01}, \ref{fig:analyze_clusters_0_1} and \ref{fig:analyze_clusters_1} show the content of the gradient pool for TA-A-GEM (left) and TA-A-GEM with random cluster assignment (right), for a sampling rate of 0.01, 0.1 and 1. Looking at the cluster contents of TA-A-GEM, it can be seen that the amount of task variety is highly affected by the sampling rate. The sampling rate can therefore be tweaked to an optimal value, in this case 0.01, given that a new task arrives every 20 epochs. This is a trivial case that we consider not to be truly task-agnostic, since it is tweaked on 20 epochs. By using clustering, TA-A-GEM manages to keep a wide variety of task information in the pool for all sampling rates. It therefore improves on the trivial case, which arises by fine-tuning the sampling rate.

\end{document}